%% file: 26_rubing_effective_kernel.tex
\documentclass[twoside,11pt, letterpaper]{article}
\usepackage{blindtext}

\usepackage[preprint]{jmlr2e}
\usepackage{pratik}

\ShortHeadings{The Dynamics of Generalization in Deep Learning}{Yang and Chaudhari}
\firstpageno{1}

\usepackage[color=gray!30,textwidth=2.5cm,textsize=tiny,disable]{todonotes}
\reversemarginpar

\graphicspath{{../},{../fig/}}

\def \bs {-}

\newcommand{\R}{\mathbb R}
\newcommand{\one}{\bm 1}
\newcommand{\softmax}{\operatorname{softmax}}
\newcommand{\Span}{\operatorname{span}}
\newcommand{\epsbar}{\bar\epsilon}

\begin{document}

{\footnotesize \listoftodos}
\clearpage

\title{The Dynamics of Generalization in Deep Learning
\xspace}

\author{\name Rubing Yang \email rubingy@sas.upenn.edu \\
       \addr Department of Applied Mathematics and Computational Sciences,\\
       University of Pennsylvania\\
       Philadelphia, PA, 19104, USA
       \AND
       \name Pratik Chaudhari \email pratikac@seas.upenn.edu\\
       \addr Department of Electrical and Systems Engineering,\\
       University of Pennsylvania\\
       Philadelphia, PA, 19104, USA}

\maketitle
\thispagestyle{empty} 

\begin{abstract}%
We derive a differential equation that governs the evolution of the generalization gap when a model is trained by gradient descent-based methods. This differential equation is driven by two key quantities, a contraction factor that brings together trajectories corresponding to slightly different datasets, and a perturbation factor that accounts for them training on different datasets. The coupled decay of contraction and perturbation guarantees a controlled accumulation of generalization gap during training. We analyze this differential equation to show that the generalization gap is given by a quadratic form that consists of an ``effective Gram matrix'' that depends upon the training trajectory and a certain residual of the predictor at initialization. Our framework is applicable to general deep networks and smooth loss functions. In numerical experiments on different neural network architectures, datasets and sample sizes, we show that this quadratic form accurately captures the actual generalization gap. We also show how to instantiate our framework in a number of examples via analytical calculations. For example, for high-dimensional linear regression, our framework matches existing calculations of generalization gap in the literature exactly in under-parameterized, over-parameterized and critical regimes.\\[1ex]
\end{abstract}
\begin{keywords}
generalization gap, contraction theory, algorithmic stability, gradient flow, general deep networks
\end{keywords}

\input{introduction}
\input{preliminaries}
\input{methods}
\input{discussions}

\acks{
This work was funded by grants from the National Science Foundation (IIS-2145164, CCF-2212519) and the Office of Naval Research (N00014-22-1-2255).
}

\input{appendix}

\bibliography{bib/ref,bib/pratik}

\end{document}

%% file: introduction.tex

\section{Introduction}
\label{s:intro}

Consider a training dataset that has $n$ data points sampled independently from the uniform distribution on a two-point domain $\{(x_1,y_1), (x_2, y_2)\}$ where $x_1, x_2 \in \reals^d$ are orthogonal vectors with unit norm. There are many ways to understand the generalization loss of a parameterized predictor fitted to such data. For example, standard weight-norm based techniques, such as those of \citet{AroraFineGrained2019} give a bound of $\OO(n^{-1/2})$ on the generalization loss. Mutual information-based techniques, such as those by \citet{pensia2018generalization} and \citet{NegreaInformationSGLD2019}, lead to a better bound, $\OO(n^{-1})$. As we will show in \cref{sec: eg two pt}, the true generalization gap of this problem is only $\Theta(2^{-n})$. Therefore, even in this simple toy setting, these techniques drastically over-estimate the generalization loss.

The central idea behind both approaches above is to characterize the ``volume'' of the hypothesis class explored during training, though they do so in distinct ways. Weight-norm based approaches compute the Rademacher complexity of weight configurations reachable from initialization, say, using gradient descent. These bounds can be quite loose for neural networks. Rather than directly dealing with the hypothesis space, mutual information-based approaches characterize how the generalization gap accumulates along the training trajectory in terms of gradient covariance. This leads to tighter bounds but these techniques apply only to randomized algorithms, such as stochastic gradient descent with additive isotropic noise. And the quality of the bounds is sensitive to the magnitude of this added noise.

\subsection{Contributions of this paper}
Our work introduces two key technical advances to address the limitations of the above frameworks. First, we analyze the generalization gap directly in terms of the loss rather than the model parameterization. This effectively handles the non-uniformity of the Lipschitz constant of neural loss landscape and scenarios where multiple distinct regions of the weight space have the same loss. Second, we derive an estimate of the generalization gap for deterministic training algorithms that mirrors the structure of mutual information-based bounds. By tracking the training trajectory directly, our analysis models the learning dynamics more faithfully and yields significantly more accurate estimates of the generalization gap. In the simple example above, our technique accurately recovers the $\Theta(2^{-n})$ rate.

We investigate the dynamics of a quantity called the ``averaged loss difference'' $\bar \D_n(t)$ which is akin to a leave-one-out estimate of the generalization gap. We show that
\[
    \dv{\bar\D_n(t)}{t} = -\bar c_n(t) \bar\D_n(t) + \bar\e_n(t), \quad t > 0, \quad \bar\D_n(0) = 0.
\]
Here, $\bar c_n$ is a ``contraction factor'' that describes how quickly two training trajectories, one trained on all $n$ samples in the dataset and another trained on $(n-1)$ samples, come together. The quantity $\bar \e_n$ is a ``perturbation factor'' that keeps these two trajectories apart because of the one distinct sample. Both these quantities are positive and typically decrease in magnitude during training. This explains why generalization gap is controlled in neural networks. We develop detailed mathematical interpretations of the contraction and perturbation factors. The former can be approximated by a generalized Rayleigh quotient between the per-sample gradient and difference between weights along the two training trajectories weighted by the Hessian. The perturbation factor is directly related to the trace of the covariance of the gradients. Our technique for analyzing $\bar \D_n$ in terms of the contraction and perturbation factors is a generalization of contraction theory for time-varying dynamical systems that evolve on non-convex energy landscapes.


The key result of this paper is to show that, for general deep networks and smooth loss functions, after gradient flow runs for time $t$, the averaged loss difference
\[
\bar \D_n(t) = \vec r_n(0)^\top K_n(0, t)\ \vec r_n(0),
\]
where $K_n$ is a quantity that we call the ``effective Gram matrix'' and $\vec r_n(0)$ is the gradient of the loss function with respect to the predictor at initialization. The effective Gram matrix is positive semi-definite, it scales as $\OO(n^{-1})$ and depends upon a certain covariance of the gradients integrated along the entire training trajectory. The generalization gap is small if the initial residual projects on the subspace of the effective Gram matrix with small eigenvalues. The quadratic form for $\bar \D_n(t)$ provides a new perspective into similarities between deep networks with kernel machines, not in terms of their training dynamics (where the equivalence holds only under certain modeling assumptions such as infinite width), but in terms of the generalization gap. Across numerous experiments on different neural network architectures, datasets and sample sizes, we show that the quadratic form accurately captures the actual generalization gap. For high-dimensional linear regression, we analytically calculate the contraction and perturbation factors to show that the asymptotic averaged loss difference $\lim_{t \to \infty} \bar \D_\infty(t)$ exactly matches existing calculations in the literature of the generalization gap in under-parameterized, over-parameterized and critical regimes. We  provide an example where these quantities can be calculated analytically for a simplified one-layer self-attention-based network. Finally, we also show how our framework can be adapted to stochastic gradient descent.


%% file: preliminaries.tex

\section{Preliminaries}
\label{s: prelim}

Let $[n]$ denote the set of integers $\{1,...,n\}$.
The notation $a\cdot b$ denotes the inner product of vectors $a$ and $b$. For a function $h$, we write $h(w)|^b_a \equiv h(b)-h(a)$ and $h(w)|_a \equiv h(a)$. We use $\abs{\cdot}$ for the absolute value of a scalar, $, \norm{\cdot}_2$ for the $\ell_2$-norm of a vector or a matrix, and $\norm{\cdot}_F$ for the Frobenius norm of a matrix.

Let $\XX$ and $\YY$ be input and output spaces respectively, and let $\ZZ=\XX \times \YY$ be the sample space. If $\WW$ denotes the parameter (weight) space, consider the predictor $f:\WW \times \XX \to \YY$ and a loss function $\ell:\YY \times \YY \to \reals^+$. As an example, for the cross-entropy loss on a $C$-class classification problem, we could set $\YY = \reals^C$ and $\ell(f(w,x), y)=-\sum_{j=1}^{C} y^{(j)} \log p^{(j)}$, where $p^{(j)} \propto \text{exp}(f^{(j)}(w,x))$ (normalized appropriately) is the softmax probability of class $C$. We use the shorthand $\ell(w,z) \equiv \ell(f(w,x),y)$.

Consider a dataset $S_{n} = \{z_i = (x_i,y_i)\}_{i\in [n]}$ of size $n$ with each $z_i \in \ZZ$ drawn i.i.d.\@ from a distribution $D$. Let $D^n$ denote the distribution of such datasets, i.e., $S_n \sim D^n$. Let $S_n^{\bs i} = \{z_1,...,z_{i-1},z_{i+1},...,z_n\}$ denote a dataset obtained by removing the $i$-th datum. The quantity $\bar \ell(w,S_n) = n^{-1} \sum_{i=1}^n \ell(w,z_i)$ is the average loss over the dataset $S_n$. Let $w_n(t)$ and $w_n^{\bs i}(t)$ be the  solutions of gradient flows
\aeq{
\label{eq: gf}
\dv{w}{t}=-\nabla \ell(w,S_n),\quad \dv{w}{t}=-\nabla \ell(w,S_n^{\bs i}),
}
respectively. Our calculations will often use quantities that depend upon $w_n(t)$ and $w_n^{\bs i}(t)$, i.e., two points along weight trajectories that are trained on slightly different datasets. The numerical experiments that validate our calculations will be conducted by omitting $m$ samples. All expressions will therefore replace $z_i \equiv S_{(m)}$ and $S_n^{\bs i} \equiv S_n^{-(m)}$.

To reduce clutter, we denote $\delta w_i(t) = w_n^{-i}(t) - w_n(t)$. For the first-order derivatives, we denote the single-sample gradient by $g_i = \nabla \ell(w_n, z_i)$ and the average gradient over the leave-one-out dataset as $g^{\bs i} = \nabla \bar{\ell}(w_n, S_n^{\bs i})$. Similarly, for the second-order derivatives, we distinguish the single-sample Hessian $H_i = \nabla^2 \ell(w_n, z_i)$ from the leave-one-out dataset Hessian $H^{\bs i} = \nabla^2 \overline{\ell}(w_n, S_n^{-i})$. In such expressions, we will typically omit the subscript $n$ indicating the size of dataset and $t$ that indicates time; we will clarify in places where there could be ambiguity.

Given a predictor $f$, a loss function $\ell$, and an initialization $w_n(0)$,
\aeqs{
R(S_n,t) =\mathbb{E}_{z}[\ell(w_n(t),z)], \quad
R_{\text{train}}(S_n,t) = \bar \ell(w_n(t),S_n),
}
are the \textbf{generalization loss} and the \textbf{train loss} of gradient flow trained on $S_n$ at time $t$, respectively.
Our main quantity of interest is their difference, \textbf{the generalization gap}
\aeq{
    \d R(S_n,t)=R(S_n,t) - R_{\text{train}}(S_n,t).
    \label{eq:gap}
}
We will typically use the averaged versions of the quantities, e.g., $\E_{S_n}\sbr{R(S_n,t)}$ which is the average generalization loss over the training samples. We will often omit the subscript $S_n$ in the expectation when there is no ambiguity.

%% file: methods.tex

\section{Results}
\label{s: methods}

The \textbf{pointwise loss difference}
\beqs{
\D_n^{\bs i}(t) = \ell(w_n^{\bs i}(t),z_i)-\ell(w_n(t),z_i),
}
is the difference of the loss along two weight trajectories on slightly different datasets. The \textbf{averaged loss difference}
\beq{
\label{eq: delta}
\bar \D_n(t) = \frac{1}{n}\sum_{i=1}^n \D_n^{\bs i}(t)
}
is similar to the Leave-One-Out-Cross-Validation (LOOCV) loss. The following lemma shows how the expected generalization gap can be approximated using the averaged loss difference.

\begin{lemma}
\label{lem: gap rep}
Assume that the expected generalization loss $\E \sbr{R(S_n,t)}$ is non-increasing in $n$, the expected training loss $\E \sbr{R_{\text{train}}(S_n,t)}$ is non-decreasing in $n$, and the expected generalization gap $\E \sbr{\d R(S_n,t)}$ is non-negative for all $n,t$. Then
\aeqs{
\E \sbr{\d R(S_n,t)}
\leq \mathbb{E} \sbr{\bar \D_n(t)} \leq \E \sbr{\d R(S_{n-1},t)}.
}
If we also have $\E[\delta R(S_{n},t)] / \E[\delta R(S_{n-1},t)] \to 1$ as $n \to \infty$, then,
\[
\E \sbr{\d R(S_n,t)} = \E \sbr{\bar \D_n(t)}+o\rbr{\E \sbr{\d R(S_n,t)}}.
\]
\end{lemma}

All proofs are deferred to \cref{app: methods}. This lemma shows that the expected generalization gap $\E_{S_n}\sbr{\d R(S_n,t)}$ can be  approximated by the expected averaged loss difference $\E_{S_n}\sbr{\bar \D_n(t)}$. This lemma motivates the theory developed in this paper. Our strategy will be to characterize the generalization gap using quantities derived from $\D_n(t)$. \cref{app: approx gap} discusses a few remarks on the assumptions in this lemma. It gives numerous analytically-tractable models used in the literature where these assumptions hold.

\subsection{Evolution of the averaged loss difference $(\D_n^{\bs i}$ and $\bar \D_n$)}
\label{s: evo gap}

We begin with a lemma that describes the evolution of $\D_n^{\bs i}(t)$ in terms of two quantities, a contraction factor $c_n^{\bs i}(t)$ and a perturbation factor $\e_n^{\bs i}(t)$.

\begin{lemma}
\label{lem: contraction}
For a loss function $\ell(w,z)$ that is differentiable in $w$ for all $z$,
\[
    \dv{\D^{\bs i}_n(t)}{t}= -c_n^{\bs i}(t)\D_n^{\bs i}(t)+\e_n^{\bs i}(t), \quad t  > 0, \quad \D_n^{\bs i}(0) = 0,
\]
where
\[
\aed{
c_n^{\bs i}(t) &=
\f{\nabla \ell(w,z_i) \cdot \nabla \bar \ell(w,S_n^{\bs i}) \big|^{w_n^{\bs i}(t)}_{w_n(t)}}{\D_n^{\bs i}(t)} \text{ and}\\
\e_n^{\bs i}(t) &=
\nabla \ell(w,z_i) \cdot \rbr{\nabla \bar \ell(w,S_n) - \nabla \bar \ell (w,S_n^{\bs i})} \bigg |_{w_n(t)}
}
\]
are the pointwise contraction and perturbation factors, respectively.
\end{lemma}
The numerator of the contraction factor $c_n^{\bs i}(t)$
is the discrepancy between weights $w_n(t)$ and $w_n^{\bs i}(t)$ along training trajectories of two different datasets ($S_n$ and $S_n^{\bs i}$ respectively). We can interpret the contractor factor as a ``spring constant'' that pulls these two weight trajectories closer due their shared $(n-1)$ samples. The perturbation factor $\e_n^{\bs i}(t)$ is the force that keeps the two trajectories apart because of one distinct sample, it depends upon the term $\nabla \bar \ell(w,S_n) - \nabla \bar \ell (w,S_n^{\bs i})$. This lemma can be extended to any piecewise differentiable loss if we define the gradient $\nabla \ell (w,z)$ at the non-differentiable point to belong to the sub-differential and have a bounded norm.
This covers all commonly used neural architectures and activation functions.

\cref{app:s: evo gap} shows how we can take the average over the numerator and denominator of $c_n^{\bs i}(t)$, and over $\e_n^{\bs i}(t)$ in \cref{lem: contraction} to obtain:
\aeq{
\label{eq: evo gap}
\dv{\bar\D_n(t)}{t} = -\bar c_n(t) \bar\D_n(t) + \bar\e_n(t).
}
The \textbf{averaged contraction factor}
\aeq{
\label{eq: c}
\bar c_n(t)= \f{\frac{1}{n}\sum_{i=1}^n \nabla \ell(w,z_i)\cdot \nabla \bar \ell(w,S_n^{\bs i}) \big |_{w_n(t)}^{w_n^{\bs i}(t)}}{\bar \D_n(t)},
}
and the \textbf{averaged perturbation factor}
\aeq{
\label{eq: e}
\bar \e_n(t)= \f{\tr\hat \S_n(t)}{n-1},
}
where $\hat \S_n(t)=\cov_{z\sim \text{Unif}(S_n)} \rbr{\nabla \ell(w_n(t),z)}$ is the covariance matrix of $\nabla \ell(w_n(t),z)$ for $z$ sampled uniformly from the dataset $S_n$.
The solution of this differential equation can be written in the integral form as
\aeq{
\label{eq: solu gap}
\bar \D_n(t) = \int_0^t\bar \e_n(s) \exp\rbr{\int_s^t -\bar c_n(u) \dd{u}} \dd{s}
}
with the assumption that $w_n(0)=w_n^{-i}(0)$ for all $i$.

\begin{remark}
\label{rem:classical_contraction}
Contraction theory~\citep{LohmillerSlotineContraction1998, LohmillerSlotineNonlinear2000} studies dynamical systems of the form $\dv{\xi}{t} = h(\xi, t)$. Consider a uniformly positive definite matrix $M(\xi, t) \succ 0$. If $\norm{\xi(t) - \xi'(t)}_M$ is the distance between points along two trajectories $\xi(\cdot)$ and $\xi'(\cdot)$ under the metric $M$, the two trajectories are said to contract with a contraction factor $\alpha$ if
\[
\dv{}{t} \norm{\xi(t) - \xi'(t)}_M \leq -\alpha \norm{\xi(t) - \xi'(t)}_M.
\]
\cref{s: contraction} provides a more elaborate introduction. \citet{kozachkov2023generalization} gave a bound on the generalization gap by analyzing gradient flow trained on datasets with one sample replaced under the assumption that the dynamics contracts uniformly on the weight space with a contraction factor $\a$. The pointwise contraction $c_n^{-i}$ in \cref{lem: contraction} satisfies a similar equation,
\[
\dv{}{t} \rbr{\ell(w_n^{-i},z_i) - \ell(w_n, z_i)} = -c_n^{-i}(t)\ \rbr{\ell(w_n^{-i},z_i) - \ell(w_n, z_i)}.
\]
Our development thus exploits the key idea behind contraction theory but it is different in two important ways.
First, we measure the discrepancy between the trajectories in terms of the loss instead of constructing a metric on the weight space. This is important because it allows us to address the fact that the Lipschitz coefficient of neural networks is non-uniform in the weight space, and in non-convex energy landscapes such as those in deep learning because points along trajectories that are far away in the weight space can have a similar loss. Second, our contraction factor $c_n^{-i}(t)$ is local and non-uniform across both time and weight space. In other words, this paper develops a generalization of the ideas in contraction theory and allows for a more refined analysis of the generalization gap. When a dynamical system is subject to a perturbation, contraction pulls the perturbed trajectory towards the unperturbed one, whereas perturbation drives the two trajectories apart. The distance between the two trajectories is controlled by the balance between these two quantities and this is what keeps the generalization gap bounded.
In fact, if we assume a uniform bound on perturbation and contraction factors: $\bar{\epsilon}_n(t) \leq \epsilon^*$ and $\bar{c}_n(t) \geq c^*$ for all $t$, \cref{eq: evo gap} yields
\[
\bar{\Delta}_n(t) \leq \frac{\epsilon^*}{c^*} (1-e^{-c^*t}).
\]
This recovers the bound of \citet{kozachkov2023generalization}.
\end{remark}

\begin{remark}
The quantity $\E\sbr{\bar\D_n(t)}$ in \cref{lem: gap rep} is the expected value over training datasets, of the pointwise loss difference averaged over the samples in the dataset, and it is closely related to the expected generalization gap $\E \sbr{\d R(S_n, t)}$. While we can also write down a differential equation for its evolution:
\beq{
\aed{
&\dv{}{t} \E \sbr{\bar \D_n(t)} = -\bar c_n(t) \E\sbr{\bar \D_n(t)} + \bar\e_n(t), \text{ where}\\
\bar c_n &= \f{ \E \sbr{\frac{1}{n}\sum_{i=1}^n \nabla \ell(w,z_i)\cdot \nabla \bar \ell(w,S_n^{\bs i}) \big |_{w_n(t)}^{w_n^{\bs i}(t)}}}{\E \sbr{\bar \D_n(t)}}, \quad
\bar \e_n = \f{ \E \sbr{\tr \hat \S(t)}}{n-1},
}
}
analyzing this equation is quite difficult because it involves an expectation over the sample set. This is why we focus our analytical development on $\bar \D_n(t)$ and instead provide an experimental characterization of the differences between $\bar \D_n(t)$ and the true generalization gap $\delta R(S_n, t)$ in the sequel.
\end{remark}

\subsection{Understanding the contraction factor $\bar c_n$}
The following lemma and the adjoining experiments allow an interpretation of the contraction factor $c^{-i}_n(t)$ as a generalized Rayleigh quotient.

\begin{lemma}
\label{lem: contraction approx}
Assume that the loss function $\ell(w, z)$ is twice continuously differentiable with respect to $w$ for all $z \in \mathcal{Z}$ and there exist constants $G, L_1, L_2 > 0$ such that for all $w, w_1, w_2 \in \mathcal{W}$ and $z \in \mathcal{Z}$,
\begin{enumerate}[(i),nosep]
\item $\|\nabla \ell(w, z)\|_2 \le G$;
\item $\|\nabla^2 \ell(w, z)\|_2 \le L_1$;
\item $\|\nabla^2 \ell(w_1, z) - \nabla^2 \ell(w_2, z)\|_2 \le L_2 \|w_1 - w_2\|_2$.
\end{enumerate}
Suppose there exists a constant $\iota_{\text{nd}}>0$ such that $\delta w_i(t)\neq0$ and
\(
\abs{g_i\cdot\delta w_i(t)}\geq \iota_{\text{nd}} \norm{\delta w_i(t)}_2
\)
for all indices and times under consideration. If $\Delta_n^{-i}(t)\neq0$, then the pointwise contraction factor
\beq{
c_n^{-i}(t) = \hat c^{-i}_1(t) + \hat c^{-i}_2(t) + O(\|\delta w_i(t)\|_2),
\label{eq: cn_approx}
}
where
\[
\hat{c}_1^{-i} \triangleq \frac{g^{-i} \cdot H_i \delta w_i}{g_i \cdot \delta w_i},
\quad
\hat{c}_2^{-i} \triangleq \frac{g_i \cdot H^{-i} \delta w_i}{g_i \cdot \delta w_i};
\]
all quantities being functions of time $t$.
Similarly, assuming $\overline{\Delta}_n(t) \neq 0$ and
\(
\abs{\sum_{i=1}^n g_i\cdot\delta w_i(t)}
\geq \iota_{\rm nd}\sum_{i=1}^n\norm{\delta w_i(t)}_2,
\)
the averaged contraction factor
\beq{
\bar c_n(t) = \hat c_1(t) + \hat c_2(t) + O(\max_i \|\delta w_i(t)\|_2)
\label{eq: bar_cn_approx}
}
where
\[
\hat c_1 = \sum_{i=1}^n \hat c_1^{-i} \underbrace{\rbr{\frac{g_i \cdot \delta w_i}{\sum_{j=1}^n g_j \cdot \delta w_j}}}_{\alpha_i} = \sum_{i=1}^n \hat c_1^{-i} \a_i.
\]
The expression for $\hat c_2$ is analogous. Note that $\sum_i \a_i = 1$.
\end{lemma}

\begin{figure}[!htpb]
\centering
\includegraphics[width=0.330\linewidth]{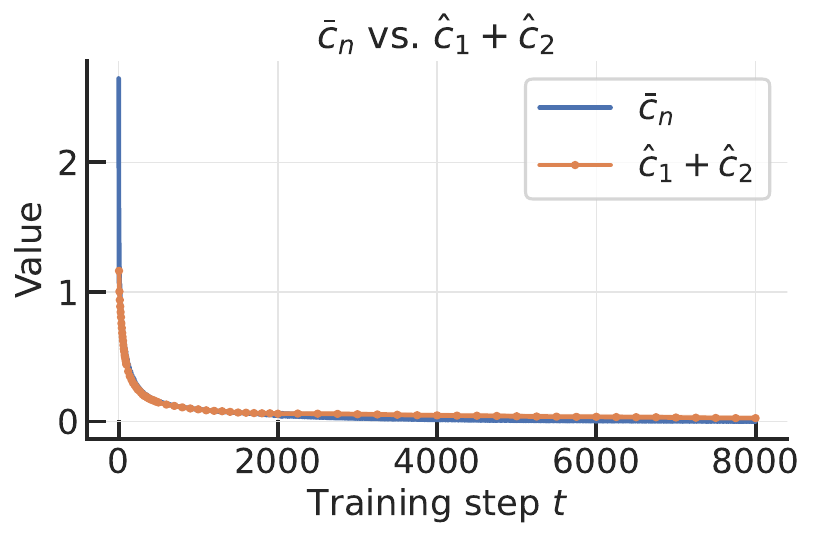}
\includegraphics[width=0.330\linewidth]{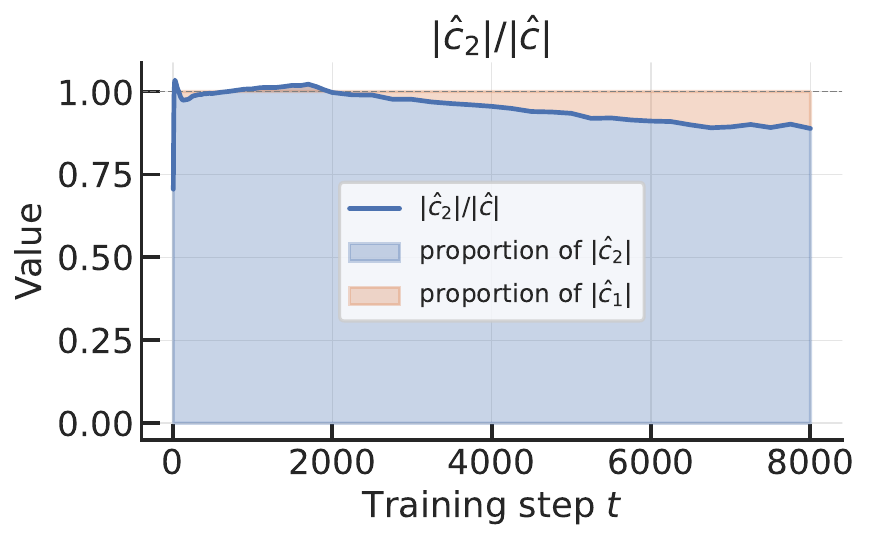}
\includegraphics[width=0.320\linewidth]{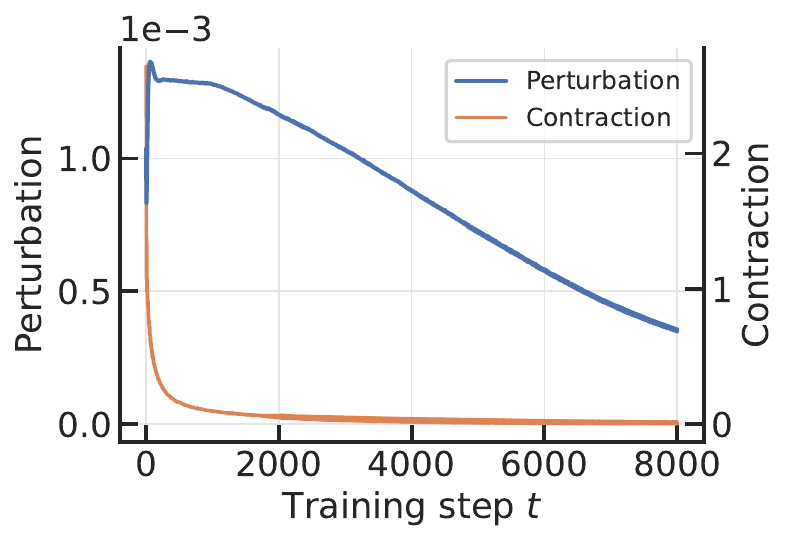}
\caption{
\textbf{Left:} True averaged contraction factor ($\bar c_n$) and its approximation by $\hat c_1 + \hat c_2$ on a full-connected network trained on MNIST for classifying two classes. Number of samples $n = 50,000$, and we conduct experiments by omitting $m = 100$ samples, averaged over 100 trajectories. The model is trained to the endpoint with train loss 0.007.
\textbf{Middle:} The contraction factor is dominated by the second component $\hat c_2$.
\textbf{Right:} The contraction and the perturbation factors decay as training progresses.
}
\label{fig: approx c}
\end{figure}

\begin{figure}[!b]
\centering
\includegraphics[width=0.325\linewidth]{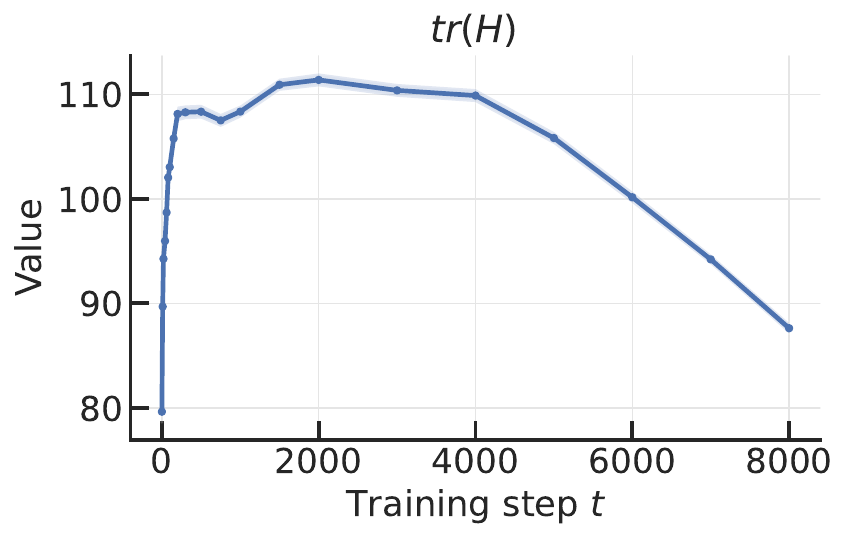}
\includegraphics[width=0.325\linewidth]{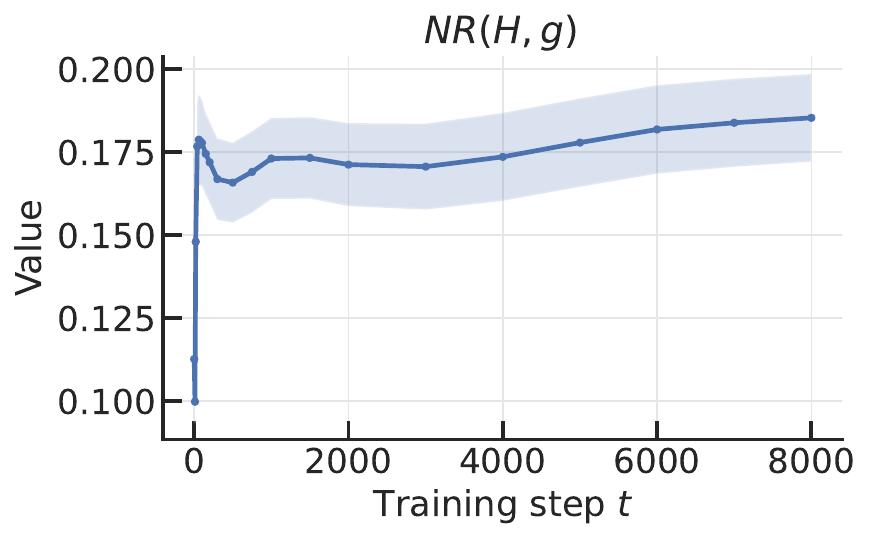}
\includegraphics[width=0.325\linewidth]{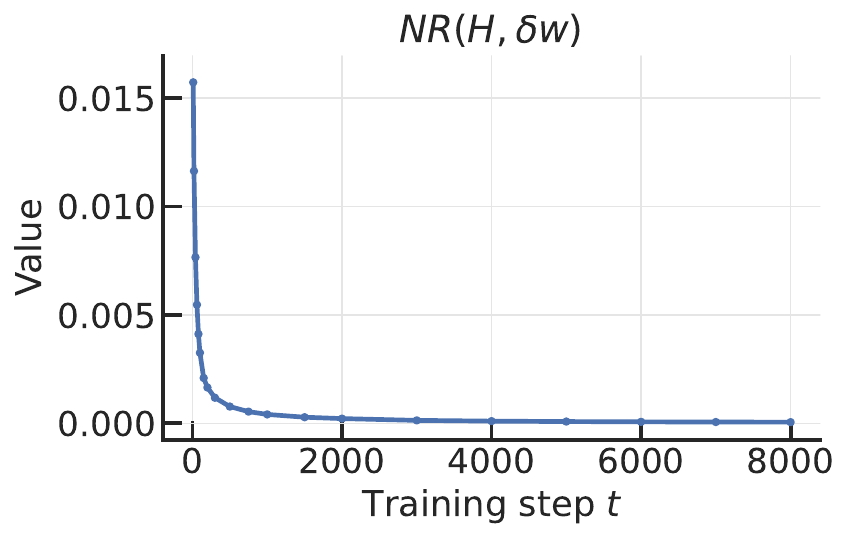}
\caption{
Statistics of the generalized Rayley quotient approximation of $\bar c_n$ for a full-connected network trained on a binary classification problem on MNIST for $n = 50,000$ samples by omitting $m = 100$ samples, averaged over 100 trajectories.
\textbf{Left:} The averaged trace of the Hessian $\tr(H) = \mathbb{E}_{(m)}\sbr{H^{-(m)}}$ decreases during training.
\textbf{Middle:} The average of the normalized Rayley quotient of Hessian and the gradient $\text{NR}(H, g) = \E_{(m)} \sbr{\f{g_{(m)}^\top H^{-(m)}g_{(m)}}{\lambda_{\max}(H^{-(m)}) \norm{g_{(m)}}^2}}$ remains large at the end of training.
\textbf{Right:} The average of the normalized Rayley quotient of Hessian and parameter difference $\text{NR}(H,\delta w) = \E_{(m)}\sbr{\f{\delta w_{(m)}^\top H^{-(m)}\delta w_{(m)}}{\lambda_{\max}(H^{-(m)}) \norm{\delta w_{(m)}}^2}}$ decays to zero at the end of training.
}
\label{fig: rayley analysis}
\end{figure}

\cref{fig: approx c} shows that $\bar c_n$ is well approximated by $\hat c_1 + \hat c_2$ and all three quantities are mostly positive throughout the training process. 
Notice that
\(
    \hat c_2(t) = \sum_{i=1}^n \a_i \rbr{\f{g_i \cdot H^{-i} \delta w_i}{g_i \cdot \delta w_i}}
\)
which is a weighted generalized Rayleigh quotient.
%
%
Both contraction and perturbation factors decay as training progresses. This is useful because when the magnitude of the perturbation factor is large during early training, the magnitude of the contraction factor which shrinks the generalization gap is also large. We calculate the decay of the contraction and perturbation factors precisely in the high-dimensional regression setting in \cref{lem: eg high dim con pert}, reflecting the similar coupled decay trend.

\cref{fig: rayley analysis} analyses the Rayleigh quotient further. It shows that even if the trace of the Hessian decreases during training, the gradient is strongly aligned with the large eigenvalues of the Hessian in the later stages of training. On the other hand, the parameter difference $\d w_i$ increasingly aligns with the small eigenvalues of the Hessian as training progresses.
This behavior arises from the driven, Hessian-damped dynamics of \(\delta w_i\), 
\[
\frac{\dd \delta w_i}{\dd t}
\approx-H^{-i}(t)\delta w_i+\frac{1}{n}
\left(\nabla\ell(w_n(t),z_i)-\nabla\bar{\ell}(w_n(t),S_n^{-i})\right).
\]
its components along large-eigenvalue directions are rapidly suppressed, whereas those along small-eigenvalue directions persist and accumulate.
The decreasing alignment of $\delta w_i$ and $H^{-i}$ is what causes the weighted generalized Rayleigh quotient $\hat c_2(t)$ to decrease during training. We can calculate this decay precisely in the high-dimensional regression setting in \cref{lem: eg high dim NR}.

\subsection{Understanding the perturbation factor $\bar \e_n$}
\label{s: evo res}
Let $r(w,z)=\dv{\ell(w,z)}{f(w,z)} \in \YY$ denote the gradient of the loss function with respect to the output of the predictor $f$. Let $r_i(t) = r(w_n(t),z_i)$ denote this gradient for sample $z_i$ and weights $w_n(t)$. We can now define the \textbf{residual} on dataset $S_n$ at time $t$ to be
\beq{
\vec{r}_n(t) = \f{1}{\sqrt n} [r_1(t), \dots , r_n(t)]^\top \in \YY^n.
\label{eq:normalized_residual}
}
The residual is the collection of loss-predictor gradients on the dataset $S_n$. It effectively describes how well the weights at time $t$ have fitted to the data and indicates the direction in the predictor space in which the training will proceed. Our goal in this section is to show that the averaged perturbation factor $\bar \e_n$ is controlled by the residual.

As an example, if we consider the squared loss $\ell(y,y') = \f 1 2 (y-y')^2$ for $y, y' \in \reals$, the residual is the normalized displacement vector from the predictor to the target, i.e., $\vec r_n(t)= \f{1}{\sqrt n} (\vec f - \vec y)$, where $\vec y = [y_1,\dots,y_n]^\top$, and $\vec f = [f(w_n(t),x_1),\dots,f(w_n(t),x_n)]^\top$. If we initialize the predictor such that $f(w_n(0),x_i)=0$ for all $i\in[n]$, then the $\ell_2$-norm of the residual $\norm{\vec{r}_n(0)}_2$ is the largest at initialization and vanishes at interpolation following the gradient flow \cref{eq: gf}. Our definition of residual above generalizes the displacement vector in the squared loss case, and can be applied to any loss function with global minimum of zero. We make some remarks on the normalization by $\sqrt n$ in \cref{app: other_numerical_experiments}.

If weights evolve according to the gradient flow in \cref{eq: gf}, then
\beq{
\label{eq: evo res}
\aed{
\dv{\vec{r}_n(t)}{t} &= -\f{1}{n} P_n(t) \vec{r}_n(t),\\
P_n(t) &= \sbr{\nabla r(w_n(t),z_i)^\top \nabla f(w_n(t),x_j)}_{i,j\in[n]}.
}}
This is a linear time-varying ordinary differential equation. In general, its solution can be written as
\beq{
\label{eq: solu res}
\vec{r}_n(t) = \Omega_n(t_0,t)\ \vec{r}_n(t_0), 
}
where $\Omega_n(t_0,t)$ is called the propagator. 
In numerical experiments, we approximate $\Omega_n(t_0,t)$ by multiplying the discrete-time update matrices for the residual along the training trajectory, i.e., using Euler discretization.

\begin{lemma}
\label{lem: res}
The trace of the gradient covariance%
\footnote{Let us emphasize that $P_n(t)$ and $F_n(t)$ are elements of $\YY^n\times \YY^n$. For regression problems, we might have $\YY \subseteq \reals$ in which case they are simply matrices in $\reals^{n \times n}$. For classification problems with $C$ categories, $\YY \subset \reals^C$, and therefore both quantities are four-dimensional tensors. But we can interpret them as elements of $\reals^{nC \times nC}$.
This amounts to vectorizing the tensor as a matrix.
}
\aeq{
\label{eq: cov form}
\tr \rbr{\hat\S_n(t)} = \vec{r}_n(t)^\top F_n(t) \vec{r}_n(t),
}
where $F_n \in \YY^n \times \YY^n$ has entries
\beq{
{F_n(t)}_{ij} = \begin{cases}
    \rbr{1-\frac{1}{n}}\nabla f(w_n(t), x_i)^\top \nabla f(w_n(t), x_i) & \text{if } i = j,\\
    - \frac{1}{n} \nabla f(w_n(t),x_i)^\top \nabla f(w_n(t),x_j) & \text{if } i \neq j.
\end{cases}
\label{eq: F}
}
\end{lemma}

\cref{eq: e,eq: solu res} and the above lemma give
\beq{
(n-1) \bar \e_n = \tr\rbr{\hat \S_n(t)} = \vec{r}_n(0)^\top \Omega_n(t)^\top F_n(t) \Omega_n(t) \vec{r}_n(0),
\label{eq:e_n_final}
}
where we denote $\Omega(0,t)$ as $\Omega(t)$ for short. In \cref{eq: cov form}, the term $F_n$ pertains to the covariance of the predictor on the training dataset $S_n$.
We can see that the residual $\vec{r}_n(t)$ controls the magnitude of gradients $\nabla \ell (w,z_i)$ for $i \in[n]$ and therefore also the covariance. For networks that train quickly, the residual norm $\norm{\vec{r}_n(t)}_2$ vanishes quickly, leading to a smaller accumulation of the perturbation term $\bar \e_n$ above, and hence a smaller generalization gap.

\subsection{An effective Gram matrix governs the generalization gap}
\label{s: eff ker}

We next derive an expression of averaged loss difference $\bar \D_n(t)$ in terms of a certain quadratic form of the initial residual and an ``effective Gram matrix'', by analyzing the evolution of $\bar \D_n(t)$ and $\vec{r}_n(t)$ during training.  The following theorem combines the solution of $\bar\D_n(t)$ in \cref{eq: solu gap} and $\vec{r}_n(t)$ in \cref{eq: solu res}, along with the decomposition of $\hat\S_n(t)$ in \cref{eq: cov form}.

\begin{theorem}
\label{thm: eff ker}
Assume that the evolution of $w_n(t)$ and $w_n^{\bs i}(t)$ follows \cref{eq: gf} and the loss function $\ell(w,z)$ is smooth in $w$ for every $z \in \ZZ$. We have
\aeq{
\label{eq: ker gap}
&\bar \D_n(t)
= \vec{r}_n(0)^\top K_n(0,t) \vec{r}_n(0),
}
where
\aeq{
\label{eq: eff ker}
K_n(0,t) = \frac{\int_{0}^t \Omega_n(s)^\top F_n(s) \Omega_n(s) \exp\rbr{-\int_s^t\bar c_n(u) \dd u} \dd s}{(n-1)}
}
is positive semi-definite. Let 
\[
K_n \triangleq \lim_{t\to\infty} K_n(0,t)
\]
when the limit exists, then we have 
\aeqs{
\bar \D_n(\infty)  \triangleq \lim_{t\to\infty}\bar\D_n(t)
= \vec{r}_n(0)^\top K_n \vec{r}_n(0).
}
We call $K_n$ the \textbf{effective Gram matrix} because it is a weighted average of Gram matrices of the form $V^\top V$, where $V=\sqrt{\f{F_n(s)}{n-1}}\Omega_n(s)$.
\end{theorem}

The quadratic form $\vec{r}_n(0)^\top K_n \vec{r}_n(0)$ in \cref{thm: eff ker} gives a data and architecture dependent measure of complexity that characterizes the generalization gap of general deep neural networks. The key conclusion of this theorem is that if the initial residual $\vec r_n(0)$ projects on the subspace of $K_n$ with small eigenvalues, then the eventual generalization gap $\bar \D_n(\infty)$ is small. This result is slightly stronger than ``post-hoc'' estimates of generalization. While the effective Gram matrix depends upon the entire training trajectory, the initial residual can be computed before training a model. The following lemma suggests conditions under which the limiting object, the effective Gram matrix, exists.

\begin{remark}[The effective Gram matrix under a fixed training kernel]
\label{rmk: fixed kernel}
Consider the squared loss and suppose $P_n(t)=P$ is fixed for all $t\geq 0$ with $P\succ 0$ symmetric.
Then $F_n(t)=F$ is also fixed and
\(
    \Omega_n(t)=\exp(-t P/ n).
\)
If the contraction factor satisfies $\bar c_n(t)\geq 0$ and
\(
    0\preceq F\preceq \beta I,
\)
then the effective Gram matrix
\[
\begin{aligned}
    K_n
    &\preceq
    \frac{1}{n-1}
    \int_0^\infty
    \exp(-\frac{s}{n}P)
    F
    \exp(-\frac{s}{n}P)
    \dd s 
    \preceq
    \frac{\beta n}{2(n-1)}P^{-1}.
\end{aligned}
\]
Consequently, since $\vec{r}(0)=-\vec{y}/\sqrt{n}$,
\[
    \bar\D_n(\infty)
    \leq
    \frac{\beta}{2(n-1)}
    \vec y^\top P^{-1}\vec y.
\]
For the ReLU kernel, assuming unit-norm inputs, we have $0 \preceq F=\frac{1}{2}I - \frac{1}{n}P \preceq \frac{1}{2}I$, hence we can take $\beta = \frac{1}{2}$.
This gives a bound on the generalization gap that has a similar form as that of \citet{AroraFineGrained2019}.
\end{remark}


\begin{lemma}[Existence of the effective Gram matrix]
\label{lem: cvg ker}
Let $\l_{\min}(t)$ be the smallest eigenvalue of $(P_n(t)+P_n(t)^\top)/2$.
Let $m(t) = \f{1}{n-1} \norm{F_n(t)}_2$ and $\omega(t) = \exp \rbr{-\f{2}{n} \int_0^t \l_{\min}(s) \dd s}$.
If
\begin{enumerate}[(i),nosep]
\item $\int_0^\infty \omega(s)\ m(s) \dd s < \infty$, and
\item the contraction factor $\bar c_n(t) \geq 0$ for all $t\geq 0$,
\end{enumerate}
then $\lim_{t\to \infty}K_n(0,t)$ exists in the $\ell_2$-norm.
\end{lemma}

Sometimes the effective Gram matrix calculated from the propagator derived from $P_n(t)$ in \cref{eq: evo res} may not converge. In such cases, we can create a perturbed version of $P_n(t)$ with a controlled $\l_{\min}(t)$ such that the conditions of \cref{lem: cvg ker} are satisfied.
This guarantees the convergence of $\lim_{t\to\infty}K_n(0,t)$ while preserving the trajectory of $\vec{r}_n(t)$ given $\vec{r}_n(0)$. We use this perturbed version in \cref{sec: eg two pt} for calculations on linear regression problems.


\subsection{An empirical characterization of the quantities in \cref{thm: eff ker}}
\label{s: approximation}

We now show that the quantities in \cref{thm: eff ker} capture the true generalization gap faithfully for different configurations of neural network training. Eigenvalues of $K_n$ represent the relative contribution to the generalization gap accumulated in the different subspaces. We will also see that if the initial residual predominantly projects onto the subspace of $K_n$ with small eigenvalues, the training process results in a small eventual generalization gap.

\begin{figure}[!htpb]
\centering
\includegraphics[width=0.495\linewidth]{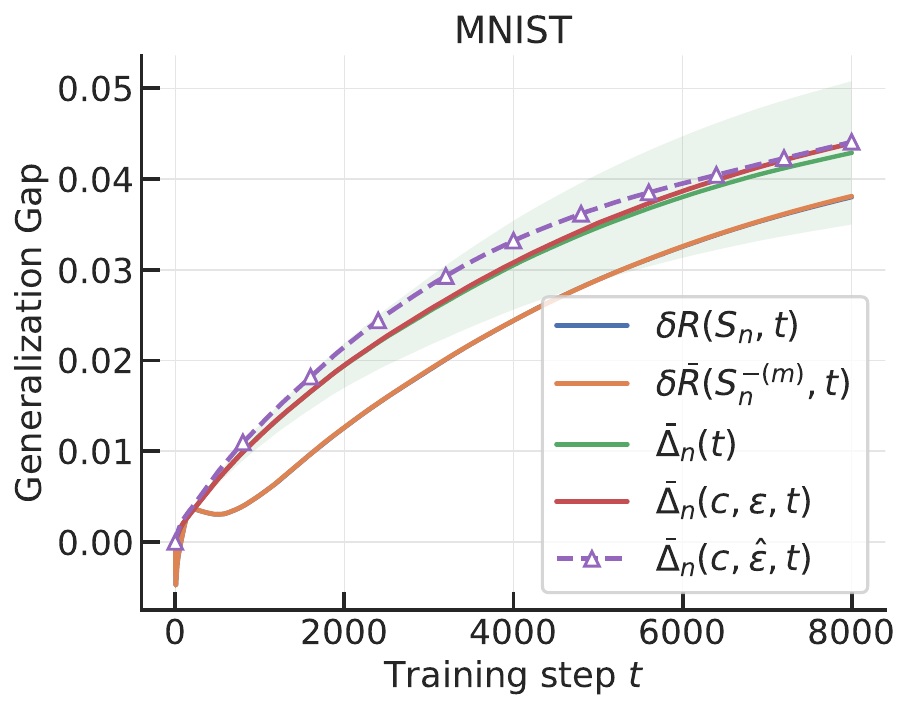}
\includegraphics[width=0.495\linewidth]{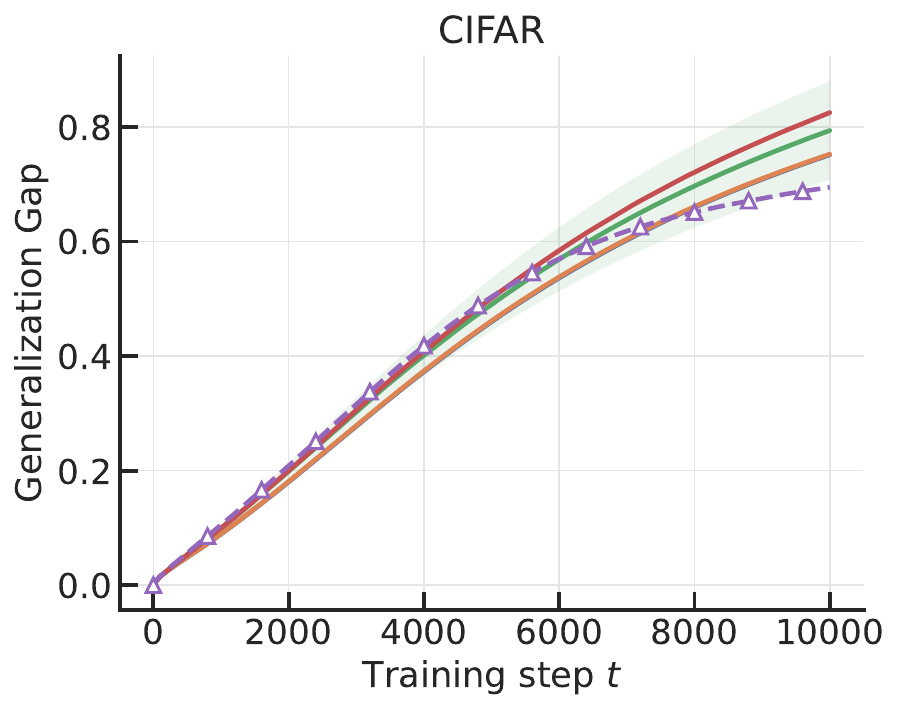}
\caption{
\textbf{Left:} A fully-connected network trained on MNIST-2, with $n=50,000$ samples and statistics computed over 100 perturbed datasets obtained by omitting $m=100$ samples. The final training loss is 0.007 after training using gradient descent.
\textbf{Right:} A fully-connected network trained on CIFAR-2 with $n=2000$ samples, $m=10$ and 100 perturbed datasets. The final training loss is 0.052 after training using gradient descent.
In both figures, the blue and orange curves are somewhat indistinguishable because they are on top of each other.
}
\label{fig: approx}
\end{figure}

We approximate the gradient flow \cref{eq: gf} by gradient descent with different learning rates.
We calculate the averaged contraction factor $\bar c_n(t)$, the averaged perturbation factor $\bar \e_n(t)$, and the matrix representation $F_n(t)$ of the gradient-covariance trace using \cref{eq: c,eq: e,eq: F} respectively.
The propagator $\Omega_n(t)$ is approximated by an Euler discretization. The integrals for $\bar \D_n(t)$ in \cref{eq: evo gap} and $K_n$ in \cref{eq: eff ker} are approximated by the trapezoidal method. For all experimental validation, we create perturbed datasets by omitting $m$ samples (see \cref{app: omit m} for the explicit expressions). \cref{app: expt setup} provides more details of these experiments.

\cref{fig: approx} shows that the true generalization gap $\d R(S_n, t)$, averaged loss difference $\bar \D_n(t)$, and the gap $\E_{(m)} \sbr{\d R(S_n^{\bs (m)},t)}$ (denoted by $\d \bar R(\cdot)$ in the plot) are all close to each other throughout training. This indicates that the generalization gap can be approximated well by $\bar \D_n(t)$.

We calculate two numerical approximations of $\bar\D_n(t)$: the quantity $\bar \D_n(c,\e,t)$ uses the true perturbation factor in \cref{eq: e} and $\bar \D_n(c, \hat \e,t)$ with an approximate perturbation factor derived from \cref{eq:e_n_final} using the propagator. First, note that $\bar \D_n(c,\e,t)$ is close to $\bar \D_n(t)$. This shows that the gradient descent approximation of \cref{eq: gf} and the trapezoidal approximation of \cref{eq: solu gap} are good. Second, the similarity of $\bar \D_n(c, \hat \e, t)$ and $\bar \D_n(c,\e, t)$ indicates that the Euler discretization of the propagator is working well. The CIFAR-2 experiment shows a similar qualitative agreement, albeit with slightly less accurate estimates of the generalization gap.
%
\cref{tab:result details} provides more detailed results where we characterize these quantities for a variety of neural architectures (fully-connected and convolutional) and datasets (MNIST and CIFAR, datasets where inputs are labeled randomly, as well as synthetically generated datasets of different kinds). Broadly across these experiments, our estimate $\bar \D_n(c, \hat \e, t)$ tracks the true generalization error $\delta R(S_n, t)$ extremely well.

\begin{table}[!htbp]
\centering
\renewcommand{\arraystretch}{1.75}
\large
\resizebox{\linewidth}{!}{%
\begin{tabular}{c c c r c r r r}
\toprule
\rowcolor[gray]{1} \textbf{Paper} & \textbf{Architecture} & \textbf{Dataset} & \textbf{\# samples} & \textbf{Training method} & \textbf{Bound on Test Data} & \textbf{Actual Value} & \textbf{Relative} \\
\midrule
\citet{AroraFineGrained2019} & FC & MNIST-2 & 10,000 & GD (second layer fixed) & 0.05 ($\ell 1$ loss) & $<$ 0.01 ($\ell 1$ loss) & $>$4 \\
\citet{DziugaiteComputing2017} & FC & MNIST-2 & 55,000 & SGD & 0.161 (error) & 0.018 (error) & 7.9 \\
\citet{WangGeneralization2022} & FC & MNIST-2 & 55,000 & SGD & 0.25 (CE loss) & & \\
Ours & FC & MNIST-2 & 50,000 & GD & 0.052 (CE loss) & 0.045 (CE loss) & 0.15 \\
Ours & FC & MNIST-10 & 1,100 & GD & 0.47 (CE loss) & 0.45 (CE loss) & 0.05 \\
Ours & LENET-5 & MNIST-10 & 1,100 & GD & 0.24 (CE loss) & 0.20 (CE loss) & 0.18 \\
\cite{NegreaInformationSGLD2019} & CNN & MNIST-10 & 55,000 & SGLD & 0.25 (CE loss) & 0.02 (error) & \\
\cite{MouSGLD2018} & CNN & MNIST-10 & 55,000 & SGLD & 1.25 (CE loss) & 0.02 (error) & \\
Ours & FC & MNIST-10 & 1,100 & SGD & 0.43 (CE loss) & 0.42 (CE loss) & 0.02\\
\midrule
Ours & FC & CIFAR-2 & 2,000 & GD & 0.597 (CE loss) & 0.580 (CE loss) & 0.013 \\
\cite{AroraFineGrained2019} & FC & CIFAR-2 & 10,000 & GD (second layer fixed) & 0.6 ($\ell 1$ loss) & 0.45 ($\ell 1$ loss) & 0.33 \\
Ours & WRN-10-2 & CIFAR-2 & 2,000 & GD & 0.554 (CE loss) & 0.546 (CE loss) & 0.014 \\
Ours & ViT-small & CIFAR-2 & 2,000 & GD & 0.529 (CE loss) & 0.546 (CE loss) & 0.032 \\
\bottomrule
\end{tabular}%
}
\caption{
\textbf{Comparison with previous results in terms of the relative accuracy of the estimate of the generalization error.}
CE loss indicates cross-entropy loss. (S)GD indicates (stochastic) gradient descent, SGLD is stochastic gradient Langevin dynamics.
``Bound" in this table refers to the numerical value of the generalization bound. ``Actual Value" is test loss or error on held-out test data.
``Relative inaccuracy" equal ``$|$Bound-Actual Value$|$ / Actual Value".
}
\label{tab:comparison}
\end{table}
\cref{tab:comparison} compares our estimate of the generalization gap with existing techniques in the literature. Different papers make different assumptions, apply to different models of neural networks, loss functions, training methods, and also use different techniques. So one must interpret this table with care, for example most methods in this table are upper bounds on the test loss while ours is an estimate. The relative inaccuracy of our estimate is quite small across a variety of settings.
This indicates that the different approximations used in our analytical development and numerical implementation are adequate.

\begin{figure}[!htpb]
\centering
\includegraphics[width=0.47\linewidth]{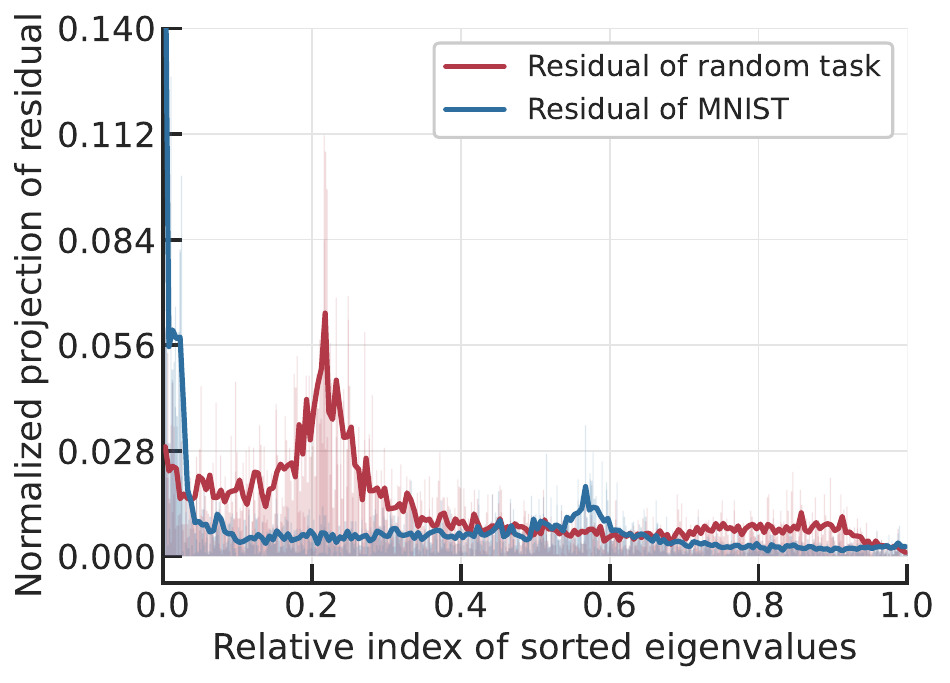}
\includegraphics[width=0.45\linewidth]{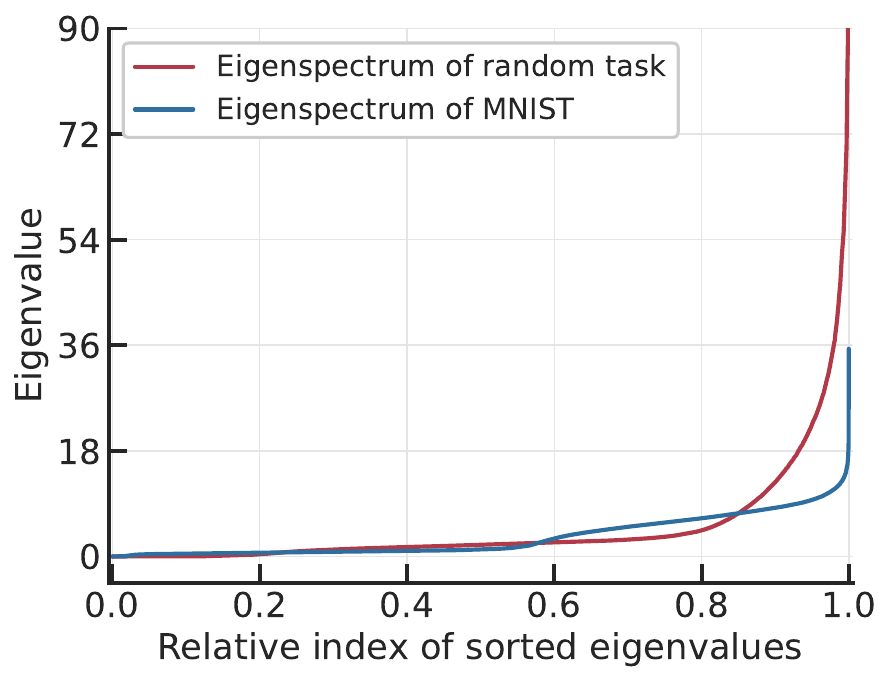}
\caption{
\textbf{Statistics of the residual $\vec{r}_n$ and effective Gram matrix $K_n$ for two different tasks.}
We use a fully-connected network trained on MNIST-2 with $n=2,000$ samples and 100 perturbed datasets created by omitting $m=10$ samples. The network has a final generalization gap of 0.118.
The ``random task'' experiment is conducted under the same setting except that MNIST-2 images are assigned classes randomly. The network has a generalization gap of 1.372.
Both tasks are trained to an endpoint with $\norm{\vec{r}_n}_2=0.094$.
The eigenspace corresponding to the smallest 5\% of eigenvalues captures 87\% of the residual magnitude for the benign task, but only 8\% for the random task.
The initial residual $\vec r_n(0)$ projects predominantly into the subspace of the effective Gram matrix $K_n$ with small eigenvalues for MNIST, whereas for the randomly labeled task it projects on subspaces with larger eigenvalues. The X-axis represents different eigenvectors sorted in increasing order of their eigenvalues, normalized by the shape of the matrix. The left figure calculates the projection of the residual onto the corresponding eigenvector $|r^\top U_k| / \norm{r}_2$. The magnitude of the eigenvalues themselves is shown on the right. The mean eigenvalue is 3.29 for MNIST and 5.08 for the random task.
}
\label{fig: align}
\end{figure}

\begin{remark}
Although the effective Gram matrix $K_n$ is defined as the infinite-time
limit of $K_n(0,t)$, exact interpolation may not be attainable while
training models in practice. Since
$\bar\D_n(t)=\vec r_n(0)^\top K_n(0,t)\ \vec r_n(0)$
holds at every finite time $t$, we instead evaluate the finite-time effective
Gram matrix $K_n(0,t_{\mathrm{end}})$ at the training endpoint
$t_{\mathrm{end}}$. Within each comparison in this subsection,
we match the normalized endpoint
residual norm $\norm{\vec r_n(t_{\mathrm{end}})}_2$ across models, ensuring
that the effective Gram matrices are evaluated at comparable stages of
training. When interpolation is attained asymptotically,
$K_n(0,t_{\mathrm{end}})$ approaches the limiting matrix $K_n$.
\label{rmk: interpolation}
\end{remark}

\cref{thm: eff ker} suggests that if the initial residual $\vec r_n(0)$ projects on the subspace with small eigenvalues of $K_n$, then the eventual generalization gap $\bar \D_n(\infty)$ is small. To study this result further, we construct two tasks: the first task (``typical'') is a MNIST classification problem with two classes (even digits vs.\@ odd digits), while the second task (``random'') consists of the same inputs but with random binary labels. A well-trained network has a small generalization gap on the typical task (here, it is 0.118). For the random task, even if the network fits the training set well, it does not obtain a good gap (1.372). \cref{fig: align} (left) indeed shows that the initial residual on the ``typical'' task lies predominantly in the subspace of $K_n$ with small eigenvalues, and on the ``random'' task the initial residual has a much higher projection on the subspace with large eigenvalues. The eigenspace corresponding to the smallest 5\% of eigenvalues captures 87\% of the residual magnitude for the benign task, but only 8\% for the random task. The mean eigenvalue of the numerically estimated effective Gram matrix is also larger for the random task (5.075) than for the typical task (3.285), contributing to the generalization gap. This shows that, much like kernel machines, if the task that we need to fit is simple, in the sense that the initial residual predominantly lies in the tail subspace of the gram matrix, then the eventual generalization gap is small.




\subsection{Other empirical results}
In \cref{app: other_numerical_experiments}, we also compute the effective Gram matrix for different datasets, architectures, and numbers of samples to show that our technique for estimating the generalization gap applies across these settings.

\subsection{Extension to stochastic gradient descent}
\label{s: sgd}

Our theoretical framework can be adapted from full-batch gradient descent to stochastic gradient descent (SGD).
Suppose that at the $t$-th iteration for the trajectory $w_n$ trained on the full dataset $S_n$, SGD samples a mini-batch with index set $B_t \subset [n]$ uniformly without replacement to update the weights. For the perturbed trajectory $w_n^{-(m)}$ trained on the reduced dataset $S_n^{-(m)}$ (where $(m)$ represents a subset of $m$ omitted samples), the effective mini-batch becomes $B_t^{-(m)} = B_t \setminus (m)$. When the omitted batch $(m)$ does not intersect with the sampled mini-batch ($B_t \cap (m) = \emptyset$), the exact same samples are used to update both $w_n$ and $w_n^{-(m)}$ at time $t$.
Consequently, the averaged contraction factor for the batched setting under SGD becomes:
\begin{equation}
\bar{c}_n(t) = \frac{\E_{(m)} \left[ \nabla \bar{\ell}(w, S_{(m)}) \cdot \nabla \bar{\ell}(w, S_{B_t^{-(m)}})|_{w_n(t)}^{w_n^{-(m)}(t)} \right]}{\bar{\Delta}_n(t)}
\end{equation}
This represents the force that pulls the two trajectories together under the stochastic gradient field $\nabla \bar{\ell}(w, S_{B_t^{-(m)}})$, instead of the full gradient field used in \cref{eq: c}. The averaged perturbation factor $\bar{\epsilon}_n(t)$ remains structurally the same as that of full-batch gradient descent (as defined in \cref{eq: e}) for all batch sizes $|B_t| > m$. This is because the perturbation is larger for SGD than for GD whenever $|(m) \cap B_t| / |B_t| > m/n$, and smaller whenever $|(m) \cap B_t| / |B_t| < m/n$, which perfectly balances out in expectation.
The evolution of the residual $\vec r_n(t)$ changes slightly under SGD because each optimization step only uses the gradients for the samples in $B_t$. Thus, the columns of the matrix $P_n(t)$ (from \cref{eq: evo res}) corresponding to the unsampled indices are zero. Because of these dynamics, the effective Gram matrix for SGD is slightly modified, yielding an appropriately modified quadratic form for the generalization gap.

We can calculate all quantities in this paper on trajectories obtained from SGD. We trained fully-connected networks on MNIST with 1100 samples using SGD. Our estimated generalization gap in terms of the cross-entropy loss is 0.429, which is very close to the true value of 0.416. Observe that this estimate (with relative error 0.02) is more accurate than the information-theoretic bounds in \cref{tab:comparison} such as those by \cite{NegreaInformationSGLD2019} and \cite{MouSGLD2018}.

\section{Calculations for analytically tractable models}
\label{s: examples}



In this section, we will focus on analytically tractable models, (i) high-dimensional linear regression, (ii) a linear regression problem with a two point-point domain that we began the introduction with, and (iii) a simplified variant of self-attention. We will obtain analytical expressions for the various quantities developed in the prequel. Our goal is to compute the contraction and perturbation factors, and the effective Gram matrix, to obtain a precise expression of the generalization gap. We defer all proofs of this section to \cref{app:s: examples}.


\subsection{High-dimensional linear regression}
\label{sec: eg high dim con pert}
Consider the case with $\XX = \reals^p$ and $\YY = \reals$. In the high-dimensional setting, we will set $p/n = \gamma$ to be the over-parameterization ratio with $p,n \to \infty$. Suppose data is generated from a linear model
\[
    y = \beta^\top x + \varepsilon, \text{ with } x \sim \mathcal{N}(0, I_p), \text{ and } \varepsilon \sim \mathcal{N}(0, \sigma^2),
\]
where the true signal $\beta$ satisfies $\norm{\beta}_2 = \rho$. 
We denote the contraction and perturbation factor as $\bar c_\infty$ and $\bar \e_\infty$ in the asymptotic limit.

The Marchenko-Pastur (MP) \citep{MPDistribution1967} spectral density function for a shape parameter $\theta$
\begin{equation*}
    f_{\text{MP}}(\lambda;\theta) = \frac{1}{2\pi\theta\lambda}\sqrt{(\lambda_+ - \lambda)(\lambda - \lambda_-)} \cdot \mathbb{I}_{[\lambda_-, \lambda_+]}, \quad \theta \in (0,1],
\end{equation*}
where the spectral bounds are $\lambda_\pm(\theta) = (1 \pm \sqrt{\theta})^2$. For the critical case where $\theta = 1$, the spectral bounds are $\lambda_- = 0$ and $\lambda_+ = 4$ and the density simplifies to
\[
     f_{\text{MP}}(\lambda;1) = \frac{1}{2\pi} \sqrt{\frac{4-\lambda}{\lambda}} \cdot \mathbb{I}_{[0, 4]}.
\]
Let the $k$-th order MP moment
\begin{equation}
\label{eq: MP moments}
    M_k(\tau, \theta) = \int_{\lambda_-}^{\lambda_+} \lambda^k e^{-\lambda \tau} f_{\text{MP}}(\lambda; \theta) \dd \lambda.
\end{equation}

\begin{lemma}
\label{lem: eg high dim con pert}
The contraction factor $\overline{c}_{\infty}(t)$ and the perturbation factor $\overline{\epsilon}_{\infty}(t)$ exhibit the following decay rates over time $t$, dictated by the overparameterization ratio $\gamma$,
\begin{enumerate}[(i)]
    \item Under-parameterized regime ($\gamma < 1$): Both factors converge to strictly positive constants. Specifically, $\lim_{t\to\infty} \overline{c}_{\infty}(t) = \frac{2(1-\gamma)^2}{2-\gamma}$ and $\lim_{t\to\infty} \overline{\epsilon}_{\infty}(t) = \sigma^2 \gamma(1-\gamma)$.
    \item Critical regime ($\gamma = 1$): Both factors exhibit polynomial decay. The contraction factor decays as $\overline{c}_{\infty}(t) = \Theta(t^{-1})$, and the perturbation factor decays as $\overline{\epsilon}_{\infty}(t) = \Theta(t^{-1/2})$.
    \item Over-parameterized regime ($\gamma > 1$): Both factors exhibit exponential decay up to polynomial prefactors. The contraction factor decays as $\overline{c}_{\infty}(t) = \Theta(t e^{-\lambda_{-} t})$, and the perturbation factor decays as $\overline{\epsilon}_{\infty}(t) = \Theta(e^{-2\lambda_{-} t})$, where $\lambda_{-}$ denotes $\lambda_{-}(1/\gamma)$.
\end{enumerate}
\end{lemma}

The proof follows by evaluating the limits $\bar{c}_\infty$ and $\bar{\epsilon}_\infty$ using the moments $M_0$ and $M_1$ of the Marchenko-Pastur (MP) distribution. MP moments decay exponentially for $0 < \theta < 1$ but polynomially when $\theta = 1$. This spectral structure induces different decay rates for the contraction and perturbation factors in each regime.
This demonstrates that the contraction and the perturbation factors have the same decay trend (constant, polynomial, or exponential) across the different parameterization regimes. The trend also reflects in the training of neural networks in \cref{fig: approx c}.

\begin{remark}
We can integrate \cref{eq: evo gap} in the underparameterized regime to notice that, since $\bar{c}_\infty$ and $\bar{\epsilon}_\infty$ converge to strictly positive constants, the asymptotic averaged loss difference
\[
    \lim_{t \to \infty} \bar \D_\infty(t) = \frac{\sigma^2 \gamma(2-\gamma)}{2(1-\gamma)}
\]
In over-parameterized regime, since $\bar c_\infty$ and $\bar\e_\infty$ both decay exponentially the asymptotic averaged loss difference remains bounded.
In contrast, in the critical regime, the averaged loss difference diverges. These limits derived from our theory coincide perfectly with the generalization gap for high-dimensional linear regression and double descent results in the literature \citep{HastieSurprises2022,belkin2019reconciling,DobribanHigh2018}.
Let us note that existing analyses of double descent focus on population risk, whereas our analysis concerns the generalization gap.
\end{remark}

To corroborate the numerical example in \cref{fig: rayley analysis}, we next show that the parameter shift between the original and leave-one-out trajectories progressively aligns with the smaller eigenvalues of the Hessian during training in the high-dimensional setting.
\begin{lemma}
\label{lem: eg high dim NR}
The Rayley quotient
\[
    \text{R}(H,\delta w) = \f{\delta w_{i}^\top H^{-i}\delta w_{i}}{\norm{\delta w_{i}}^2},
\]
exhibits the following decay rates over time $t$,
\begin{enumerate}[(i),nosep]
    \item Under-parameterized regime ($\gamma < 1$): The term $\text{R}(H,\delta w)$ decays to strictly positive constant $1-\gamma$.
    \item Critical regime ($\gamma = 1$): The term $\text{R}(H,\delta w)$ decays polynomially at a rate of $\Theta(t^{-1})$.
    \item Over-parameterized regime ($\gamma > 1$): The term $\text{R}(H,\delta w)$ decays exponentially at a rate of $\Theta(t e^{-\lambda_{-} t})$, where $\lambda_{-}$ denotes $\lambda_{-}(1/\gamma)$.
\end{enumerate}
\end{lemma}

\cref{lem: eg high dim NR} can be proved in a similar way as \cref{lem: eg high dim con pert}.
The decay of the Rayleigh quotient $\text{R}(H,\delta w)$ can be intuitively understood using the evolution of $\delta w_i$,
\[
\f{\dd \delta w_i}{\dd t} = -H^{-i} \delta w_i + \frac{x_i r_i}{n} 
\]
as derived in \cref{app: eg high dim con pert}.
This differential equation characterizes a dynamical system driven by an isotropic source term $\frac{x_{i}r_{i}}{n}$ and a selective damping force $-H^{-i}\delta w_{i}$. Components of the parameter shift $\delta w_i$ that align with the large eigenvalues of the Hessian are rapidly dissipated, while components that align with the small eigenvalues accumulate.
As training proceeds, the parameter shift projects more and more into the subspace of Hessian with small eigenvalues, which causes the Rayley quotient to decrease. This trend also reflects in the training of neural networks (\cref{fig: rayley analysis}).
In comparison, the Rayley quotient of Hessian and gradient, defined as $\text{R}(H,g_i) \triangleq \frac{g_i^\top H^{-i} g_i}{\norm{g_i}^2} = \frac{x_i^\top H^{-i} x_i}{\norm{x_i}^2}$ is unchanged during training, which is also reflected in \cref{fig: rayley analysis}.

\subsection{An analytical effective Gram matrix for linear regression}
\label{sec: eg two pt}

We next consider a very simple example that we discussed in the Introduction. Suppose the sample space is supported on two points with orthonormal inputs, i.e., $\ZZ=\{(x_1,y_1),(x_2,y_2)\}$, with $x_1^\top x_2 = 0$, each with unit norm $\norm{x_1}_2=\norm{x_2}_2=1$. Consider the dataset $S_n$ with $n$ even, where $z_i=(x_1,y_1)$ when $i\leq n/2$ and $z_i=(x_2,y_2)$ when $i>n/2$. Even if this case is extremely simple, it is illuminating because we will be able to calculate the effective Gram matrix exactly and show that for ``easy tasks'', i.e., ones where the generalization gap at the end of training is small, the initial residual predominantly projects in the subspace of the effective Gram matrix with small eigenvalues.

\begin{lemma}
\label{lem: eg two pt}
The contraction and perturbation factors for the above example are
\[
\bar c_n(t) = \bar c \triangleq \f{n-2}{2(n-1)}, \quad 
\bar{\epsilon}_n(t) = \frac{y_1^2 + y_2^2}{4(n-1)} e^{-t}.
\]
The effective Gram matrix $K_n(0,t) = U\Lambda^K(t) U^\top$ where $\L^K(t) = [\l^K_1(t),\dots,\l^K_n(t)]$, and
\aeqs{
\l^K_1(t) = \l^K_2(t) = \Theta(e^{-\bar c t}), \quad
\l^K_3(t) = \dots = \l^K_n(t) = \Theta(1).
}
The initial residual $\vec{r}_n(0)$ lies in the subspace of $K_n=\lim_{t\to\infty}K_n(t)$ with zero eigenvalue, which guarantees that $\bar \Delta_n = 0$ as $t\to\infty$.
\end{lemma}
Note that since $\ZZ$ is supported on only two points, it is intuitive that gradient flow trained on a dataset containing both types of samples generalizes and achieves zero loss for any distribution $D$ supported on $\ZZ$. This coincides with the calculation above.
If we assume a uniform distribution on $\ZZ$ instead, we have
\[
     \mathbb{E}[\bar \Delta_n(t)] = \mathbb{E}[y^\top K_n y] = \Theta(2^{-n}),
\]
because $S_n$ is supported on only one of the samples only with probability $2^{-(n-1)}$. This example therefore provides a tight prediction of the generalization gap.

\begin{remark}[Comparison with weight-norm based generalization bound]
\cite{AroraFineGrained2019} suggest that one can bound the generalization gap in terms of the norm of the eventual weights. The Gram matrix of the linear regression described above is calculated in the \cref{app: eg two pt} to be $H^\varepsilon = P_n^\varepsilon (t)$ with $\ve(t) = \bar \ve$ for some constant $\bar \ve \ll 1$. For technical reasons, this is a perturbed version of the differential equation in \cref{eq: evo res} and it guarantees positive definiteness of the drift operator without affecting the residual dynamics. By the calculations of \citet{AroraFineGrained2019}, the norm of weights can be bounded by $\sqrt{y^\top (H^\ve)^{-1} y}$, this gives a generalization bound of
\[
    \sqrt{ \f{2y^\top (H^\ve)^{-1} y}{n}} = \sqrt{\f{2(y_1^2+y_2^2)}{n}}.
\]
for 1-Lipschitz loss. As we can see, this bound is far looser than the actual generalization error, which is $\Theta(2^{-n})$ for 1-Lipschitz loss from our calculation above.
The key point to emphasize here is that by characterizing the evolution of the point-wise loss difference using the contraction factor, we can work directly in the prediction space instead of working in the weight space. This is the reason why our estimate of the generalization gap is more accurate.
\end{remark}

\begin{remark}[Comparison with information-theoretic generalization bound]
\label{rmk: example compare MI}
\citet{NeuInformationSGD2021, NegreaInformationSGLD2019,BanerjeeStability2022} derive mutual information-based generalization bounds for stochastic algorithms that depend on the sum of trace of the gradient covariance along the training trajectory $\sum_t \text{tr} \rbr{\hat\Sigma_n(t)}$. We can adapt their expression to deterministic algorithms like gradient descent using our theory. If we assume that that contraction factor $\bar c_n(t)$ is non-negative, we can bound $\bar \D_n(\infty)$ by $\int_0^\infty \bar \epsilon_n(t) \dd t$. Notice that from \cref{eq: e}, this would correspond to an estimate of the sum of the trace of the gradient covariances. If the contraction factor were identically zero, the expression for the effective Gram matrix in \cref{thm: eff ker} gives a generalization estimate of $\f{(y_1^2+y_2^2)}{4(n-1)}$. This is also a loose estimate. Evidently, the bound obtained using the sum of the trace of the gradient covariances is loose because these existing techniques do not incorporate the contraction factor.
\end{remark}

\subsection{A single layer simplified self-attention network}
\label{sec: eg attention}

We next study a nonlinear classifier with simplified self-attention operator. We show that trainable attention yields faster asymptotic decay of the averaged loss difference and perturbation than fixed attention and, under an additional positivity condition, a larger asymptotic contraction factor. This illustrates how trainable attention can improve generalization dynamics through the lens of our contraction and perturbation factors.

The data is generated as follows. Let $L$ be the number of tokens contained in one datum. For an integer
$1\le M<L$, write
\[
  \pi_{\rm sig}=\frac ML\in(0,1)
  \qquad\text{and}\qquad
  \pi_{\rm bg}=1-\pi_{\rm sig}.
\]
Let $\mathcal C:=\{1,2\}$ denote the set of classes. Let $e_1,e_2,e_3\in\R^{d_{\rm emb}}$ be orthonormal vectors and $d_{\rm emb}\ge 3$. An input $x_j$ of class $j\in\mathcal C$ contains $M$ copies of the signal token $e_j$ and $L-M$ copies of the background token $e_3$, in an arbitrary order. Its label is $y_j=(-1)^{j+1}$. Assume that the training set contains $n/2$ inputs from each class with $n\ge4$ even.
For any input $x\in\R^{L\times d_{\rm emb}}$, let the predictor be
\[
  \aed{
  f(w,x) &= v^\top h_\vartheta(x)\\
  \text{with } h_\vartheta(x) &= \frac1L\one^\top \softmax(x B_\vartheta x^\top)\ x,\\
  B_\vartheta &= \kappa \rbr{\vartheta_1e_1e_1^\top + \vartheta_2e_2e_2^\top+ \vartheta_be_3e_3^\top}
  }
\]
for a constant $\kappa\ge0$. The weights (trainable parameters) in this predictor are $w =(v,\vartheta_1,\vartheta_2,\vartheta_b)$. We are therefore considering a predictor that pools the outputs of self-attention (so-called ``mean-pooled'' operation). The quantity $B_\vartheta$ directly parameterizes the query-key map, since there are only three non-zero token-token inner-products regardless of the value of $L$ and $M$. The value map is set to identity. The case $\kappa=0$ indicates uniform attention (and a linear predictor), whereas $\kappa>0$ gives trainable attention scores.

We use the squared loss $\ell(w,(x,y)) = (f(w,x) - y)^2 / 2$ as in the previous two examples, and initialize all training trajectories at $w=0$.
We are interested in three quantities: $\bar \Delta_{n,\kappa}$, $\bar c_{n,\kappa}$, and $\bar \e_{n,\kappa}$, which are the averaged loss, contraction and perturbation respectively.
We denote $\nu_\Delta(\kappa)$, $\nu_\epsilon(\kappa)$ as the decay rate of $\bar\Delta_{n,\kappa}$ and $\bar\e_{n,\kappa}$. We denote $c(\kappa)$ as the limit infimum of $\bar c_{n,\kappa}(t)$.
\begin{align}
  \nu_\Delta(\kappa)
  & =\liminf_{\substack{n\to\infty\\ n\ {\rm even}}}
    \liminf_{t\to\infty}
    \left[-\frac1t\log
      \left|\bar \Delta_{n,\kappa}(t)\right|\right],
  \label{eq:RDelta-definition}\\
  c(\kappa)
  & =\liminf_{\substack{n\to\infty\\ n\ {\rm even}}}
    \liminf_{t\to\infty}\bar c_{n,\kappa}(t),
  \notag\\
  \nu_\epsilon(\kappa)
  & =-\lim_{t\to\infty}\frac1t
    \log\epsbar_{n,\kappa}(t).
  \notag
\end{align}

\begin{lemma}
\label[lemma]{lem:main}
Fix $\kappa>0$. The decay rate of $\bar\Delta_{n,\kappa}$ and $\bar\e_{n,\kappa}$ satisfy
\begin{equation}
  \nu_\Delta(\kappa)>\nu_\Delta(0)=\pi_{\rm sig}^2,
  \qquad
  \nu_\epsilon(\kappa)>\nu_\epsilon(0)=\pi_{\rm sig}^2.
  \label{eq:main-unconditional-results}
\end{equation}
Suppose, in addition, that the averaged loss difference $\bar\Delta_{n,\kappa}(t)$ is eventually positive for all sufficiently large sample sizes,
then
\begin{equation}
  c(\kappa)
  > c(0)
  =\pi_{\rm sig}^2.
  \label{eq:main-contraction-result}
\end{equation}
\end{lemma}

The above lemma shows that trainable attention learns to emphasize the class-specific signal tokens over the shared background tokens. The resulting representation separates the two classes more effectively, allowing the predictions to approach the correct labels more rapidly. Under the squared loss, the sample-wise gradients and their variation across samples therefore vanish faster, leading to faster perturbation decay. Meanwhile, by relating to the residual during training,we show that the predictor with $\kappa > 0 $ makes the loss difference between the full-data and leave-one-out trajectories shrink faster, resulting in stronger contraction under the additional positivity condition. Thus, trainable attention improves both perturbation and contraction through the representation it learns. This example, albeit simplistic, shows that we can calculate the contraction and perturbation factors and the generalization gap for different types of architectures.

We should note that the predictor is invariant to permutations of token positions because self-attention is permutation equivariant and the subsequent mean pooling is permutation invariant. Consequently, all inputs with the same numbers of signal and background tokens have the identical results regardless of the ordering.

Related ideas, in which attention progressively concentrates on the most relevant signal tokens, also appear in \citet{LiTheoretical2023} and \citet{OymakOntherole2023}. Under structured data models with informative and irrelevant tokens, these works show that gradient-based training can emphasize task-relevant information. Our example complements them by quantifying how this adaptive selection affects the contraction and perturbation components of generalization dynamics.

%% file: discussions.tex

\section{Related Work and Discussion}
\label{s:related_work}

Our approach adapts and develops concepts from several established lines of literature to provide a tighter, data-dependent estimate of generalization.

\paragraph{Comparing trajectories in terms of their loss instead of their weights.}
\citet{AroraFineGrained2019} also study the residual dynamics to obtain an estimate of the norm of eventual weights that are reachable by gradient descent. They derive a weight-norm-based bound using the PAC learning framework \citep{ValiantTheory1984} and Rademacher complexity \citep{bartlettRademacherGaussianComplexities2002,BartlettSpectrally2017}.
Similar ideas are adopted by \citet{AllenLearning2019} and \citet{CaoGeneralization2019} for analyzing SGD and online learning, respectively, in fully-connected neural nets.
\citet{LiuFrom2022} derive a weight-norm-bound under uniform conditions for the Lojasiewicz gradient inequality. \citet{RichardsStability2021} and \citet{AkbariHow2021} analyze the sensitivity of the weights found by the algorithm upon replacing one sample for uniformly Lipschitz losses. The key reason for loose generalization bounds in these analyses is that they are conducted in weight space, under uniform assumptions over the entire loss landscape. 
In contrast, we study the evolution of the residual in \cref{eq: evo res}, which is more closely related to the difference in loss after training on perturbed datasets---as opposed to the difference in weights. Our analysis is conducted directly in the prediction space, which is why it provides a more precise characterization of generalization.

\paragraph{Relation to stability-based, trajectory-based and information-theoretic approaches.}
Stability based generalization bounds\cite{BousquetStability2002, HardtTrain2016, XuRobustness2012} can be adapted for stochastic algorithms using information-theoretic approaches \cite{XuInformation2017, MouSGLD2018, ChuUnified2023} to provide remarkably tight generalization bounds.
\citet{DadiGeneralization2025} compare SGLD runs on different datasets to derive generalization bounds that do not grow with training time in non-convex settings.
\citet{NicklMemory2023} use the memory-perturbation equation to estimate sample sensitivity and predict test loss through approximate leave-one-out cross-validation during training.
\citet{SteinkeReasoning2020} develop a conditional mutual information framework that relates generalization to how much a learned model reveals about which examples in a super-sample were used for training.
\citet{NegreaInformationSGLD2019, NeuInformationSGD2021, BanerjeeStability2022} develop information-theoretic generalization bounds in terms of the sum of the trace of the gradient covariance along the training trajectory---this is $\sum_t \tr \hat \S_n(t)$ in our notation. This sum tells us about the size of the tube of trajectories in loss space that arises from training on slightly different datasets. The quality of the estimate of this tube completely determines the quality of the bound.
Our expression in \cref{eq: solu gap} provides a more general and tighter estimate of this tube than these prior works, because the damping factor $\exp (\int_s^t -\bar c_n(u)du)$ corrects for the size of the tube. A positive contraction factor leads to faster contraction of two trajectories that are being trained on slightly different datasets.
Our analysis applies to deterministic algorithms, unlike previous work on information-theoretic bounds \citep{XuInformation2017,MouSGLD2018,FutamiInformationBound2023}, which only holds for randomized algorithms because the proof relies on the non-expansiveness of the KL-divergence of non-singular distributions.

Beyond information-theoretic approaches, several works derive generalization bounds from quantities computed along the training trajectory: \citet{ClericoGeneralisation2025} use pathwise curvature within a deterministic PAC-Bayes framework, \citet{FuLearning2023} use the decrease in the training loss and gradient covariance, and \citet{ChenAnalyzing2023,ChenGeneralization2025} use cumulative per-example gradient norms through a data-dependent loss path kernel. \citet{JohnsonInconsistency2023} bound the expected generalization gap of stochastic training using output inconsistency and instability. 
These approaches summarize a single training path by accumulating certain quantities along the trajectory. 
Trajectory-based analysis can also motivate algorithmic choices such as early stopping: \citet{YaoEarlyStopping2007} compare empirical and population gradient-descent trajectories to derive stopping rules that balance approximation and sample-size errors.


\paragraph{Generalization bounds for kernel machines.}
Generalization loss in kernel ridge regression \citep{RakhlinJust2020, MallinarBenign2022} can be expressed in terms of quantities that resemble ours, namely, the alignment of the residuals with the Gram matrix $r(0)^\top K r(0)$ \citep{AroraFineGrained2019, JacotKernel2020}. 
\citet{WoodworthKernelRich2020} show that initialization scale controls the transition between kernel and rich regimes in a family of deep linear networks, influencing both implicit bias and generalization.
Beyond kernel methods, \citet{LiBeyondNTK2020} analyze the training dynamics of two-layer ReLU networks and show that they can achieve better generalization than kernel methods under Gaussian inputs and orthogonal target weights.
Our effective Gram matrix generalizes this idea to arbitrary deep neural networks and loss functions, going beyond two-layer neural networks and squared losses. However, unlike kernel ridge regression, where the Gram matrix is derived from a fixed kernel that directly recovers the target function, the effective Gram matrix $K_n$ in our setting varies for different datasets and training regimes and does not necessarily coincide with any fixed kernel.

\paragraph{Contraction theory.} \citet{LohmillerSlotineContraction1998,LohmillerSlotineNonlinear2000} provide a uniform contraction rate and perturbation magnitude over time and space ($\a$ and $\overline b$ in \cref{thm: slotine ctr,thm: slotine pert}).
\citet{DavydovNonEuclidean2022} extend contraction analysis to arbitrary norms and establish incremental input-to-state stability bounds that quantify how contraction and input perturbations jointly govern the distance between trajectories.
In contrast, we derive these terms $c_n^{\bs i}(t)$ and $\e_n^{\bs i}(t)$ in \cref{lem: contraction} directly from the evolution of $\bar \D_n^{-i}(t)$. This is a local quantity and describes only the contraction and perturbation of $w_n(t)$ and $w_n^{-i}(t)$. It varies over time.
The central motivation of this paper is that the non-uniformity allows for a more refined analysis of gradient flow for neural networks. Indeed, the energy landscape may not be uniformly good in the entire weight space, but as we see in~\cref{fig: approx c}, it can be benign along most of the training trajectory. Our development in this paper therefore diverges significantly from the generalization bounds derived in contraction theory \citep{kozachkov2023generalization,CharlesStability2018}, and our results generalize these ideas, see \cref{rem:classical_contraction}.

\paragraph{Cross-validation-based estimators of the generalization gap.}
These are widely used methods for estimating risk \citep{HastieElements2009}. \citet{BlumBeating1999} show that the cross-validated estimator performs no worse than the corresponding data-splitting estimator under certain conditions. \citet{KaleCross2011} show that the $k$-fold cross-validated estimator achieves at least a $k$-fold reduction in variance under certain stability conditions. \citet{KumarNear2013} relax the stability condition.
\citet{AusternAsymptotics2020} improve these results by proving central limit theorems and Berry-Esseen bounds for the cross-validation loss estimators under certain stability conditions.
\cite{PatilFailures2024} show that, in high-dimensional least squares, LOOCV consistently estimates prediction risk uniformly along the early-stopped gradient-descent trajectory.
\citet{chen2023learning} build upon PAC-Bayes bounds developed by \cite{YangDoes2022} to obtain an estimate of the effective dimensionality of deep networks by computing the relationship between the leave-one-out estimator of the test loss and the number of training samples.
Relatedly, \citet{ObermanFastRates2026} uses leave-one-out stability to derive error bounds for transductive semi-supervised learning on data-augmentation graphs.

%% file: appendix.tex

\renewcommand\thesection{S.\arabic{section}}
\renewcommand\thefigure{S.\arabic{figure}}
\renewcommand\thetable{S.\arabic{table}}
\setcounter{figure}{0}
\setcounter{section}{0}
\setcounter{table}{0}
\renewcommand{\thesubsection}{\thesection.\arabic{subsection}}

\begin{appendix}
\label{app}



\section{Contraction theory}
\label{s: contraction}
This section introduces some preliminary material on contraction theory \citep{LohmillerSlotineContraction1998,LohmillerSlotineNonlinear2000}, which provides a way to analyze solutions of slightly different dynamical systems. Contraction theory reformulates Lyapunov stability \citep{IsidoriNonlinearControl1995,Marino1995} using a quadratic Lyapunov function corresponding to a Riemannian contraction metric and its uniform positive definite matrix to characterize the necessary and sufficient conditions for exponential convergence of trajectories to each other and the stability of these trajectories to perturbations of the dynamics.
Consider a nonlinear dynamical system
\beq{
\label{eq: slotine sys}
\dv{\xi}{t} = h(\xi,t).
}
The following theorem gives guarantees of the exponential convergence of trajectories with different initializations.

\begin{theorem}[Theorem 2.1 from \cite{TsukamotoContraction2021}]
\label{thm: slotine ctr}
If there exists a uniformly positive definite matrix $M(\xi,t) \succ 0$ for all $\xi, t$, such that the following condition holds for some $\alpha > 0$,
\begin{equation}
\label{eq: slotine ctr}
\forall \xi, t:\quad \dot M+M \nabla_\xi h+\nabla_\xi h^\top M \preceq -2\a M,
\end{equation}
then all trajectories of \cref{eq: slotine sys} converge to a single trajectory under the metric induced by $M$ exponentially fast regardless of their initial conditions, i.e., for all trajectories $\xi$, $\xi'$ of \cref{eq: slotine sys},
$d(\xi(t),\xi'(t))_M \leq d(\xi(0),\xi'(0))_M e^{-\a t}$, where $d(\cdot,\cdot)_M$ denotes the distance under the metric induced by $M$. A dynamical system \cref{eq: slotine sys} satisfying \cref{eq: slotine ctr} is said to be ``contracting'', under the  ``contraction metric'' induced by $M$. The factor $\a$ is defined as the ``contraction factor''.
\end{theorem}
Using \cref{thm: slotine ctr}, we can also analyze trajectories of a perturbed dynamical system
\begin{equation}
\label{eq: slotine pert}
\dv{\xi}{t} = h(\xi,t)+b(\xi,t).
\end{equation}
Let $\xi_0(t)$, $\xi_1(t)$ be solutions of \cref{eq: slotine sys} and \cref{eq: slotine pert}, respectively. The next theorem shows that for a contracting system, the solution of the perturbed system does not differ too much from that of the original system, under certain conditions.

\begin{theorem}[Theorem 2.3 from \cite{TsukamotoContraction2021}]
\label{thm: slotine pert}
Assume that the dynamical system \cref{eq: slotine sys} is contracting under $M$ with factor $\a$. If $\overline b = \sup_{\xi,t} \norm{b(\xi,t)}$ and there exist constants $\underline m, \overline m >0$ such that $\underline m I \preceq M(\xi,t) \preceq \overline m I$ for all $\xi,t$, then we have
\aeqs{
d(\xi_1(t),\xi_0(t)) &\leq \frac{d(\xi_1(0),\xi_0(0))_M}{\sqrt{\underline m}} e^{-\a t} + \frac{\overline b}{\a} \sqrt{\f{\overline m}{\underline m}}\rbr{1-e^{-\a t}},\\
d(\xi_1(t),\xi_0(t))_M &\leq d(\xi_1(0),\xi_0(0))_M e^{-\a t} + \frac{\overline b \sqrt{\overline m}}{\a} \rbr{1-e^{-\a t}},
}
where $d(\cdot, \cdot)$, $d(\cdot, \cdot)_M$ denote the distance under Euclidean metric and metric induced by $M$, respectively.
\end{theorem}
In short, for contracting systems, for large times $t$, the bound on the distance between the solution of the original dynamics and the perturbed dynamic is determined by the perturbation of the system, the contraction factor, and the eigenvalues of the metric. In this paper, we will be interested in using these ideas to understand the difference between two trajectories evaluated on certain loss functions that are fitted using slightly different datasets.



\section{Proofs and Calculations in \cref{s: methods}}
\label{app: methods}

\subsection{Proof of \cref{lem: gap rep}}
\label{app: approx gap}

By the definition of the averaged loss difference $\bar \Delta_n(t)$,
\aeqs{
\bar \Delta_n(t) = \frac{1}{n} \rbr{\sum_{i=1}^n \ell(w_n^{-i}(t),z_i)} - \bar \ell (w_n(t),S_n)
}
Taking the expectation on both sides, we have
\aeqs{
\E \sbr{\bar \Delta_n(t)} = \E \sbr{R(S_{n-1},t)} - \E \sbr{R_{\text{train}}(S_{n},t)}
}
Using the definition of the generalization gap, this identity can also be written as
\aeqs{
\E \sbr{\bar \Delta_n(t)}
= \E \sbr{\d R(S_n,t)}
+ \E \sbr{R(S_{n-1},t)}-\E \sbr{R(S_n,t)}.
}
Therefore, if
\aeqs{
\E\sbr{R(S_{n-1},t)}-\E\sbr{R(S_n,t)}
=o\rbr{\E\sbr{\d R(S_n,t)}},
}
then
\[
\E \sbr{\d R(S_n,t)}
=\E \sbr{\bar \D_n(t)}+o\rbr{\E \sbr{\d R(S_n,t)}}.
\]

Under the two monotonicity assumptions, we have
\aeqs{
\E \sbr{R(S_{n},t)} \leq \E \sbr{R(S_{n-1},t)}, \quad \E \sbr{R_{\text{train}}(S_{n-1},t)} \leq \E \sbr{R_{\text{train}}(S_{n},t)}.
}
Therefore,
\aeqs{
\E \sbr{\d R(S_n,t)}
\leq \mathbb{E} \sbr{\bar \D_n(t)} \leq \E \sbr{\d R(S_{n-1},t)}.
}

Finally, under these monotonicity assumptions, if $\E [\d R(S_{n},t)] / \E [\d R(S_{n-1},t)] \to 1$ as $n\to \infty$, then
\aeqs{
\frac{\E [\d R(S_{n-1},t)] - \E [\d R(S_{n},t)]}{\E [\d R(S_{n},t)]} \to 0
}
as $n \to \infty$. The sandwich bound then implies
\aeqs{
0\leq
\E\sbr{\bar\D_n(t)}-\E\sbr{\d R(S_n,t)}
\leq
\E\sbr{\d R(S_{n-1},t)}-\E\sbr{\d R(S_n,t)}
=o\rbr{\E\sbr{\d R(S_n,t)}}.
}
Hence,
\[
\E \sbr{\d R(S_n,t)} = \E \sbr{\bar \D_n(t)}+o\rbr{\E \sbr{\d R(S_n,t)}}.
\]

\paragraph{Remarks on the assumptions}
The expected generalization loss $\E[R(S_n,t)]$ is non-increasing in $n$: This holds for ridgeless linear regression without label noise. When label noise is non-zero, the generalization loss decays monotonically when the number of samples is greater than the number of features. This result also holds for ridge regression when the ridge coefficient $\lambda$ decays with $n$, but not too fast, i.e., $\lambda > \frac{\sigma^2}{\sigma^2 + \norm{\theta^*}^2} \cdot \frac{1}{n}$ where, $\sigma$ is the noise variance and $\theta^*$ is the true regressor. See \cite{HastieSurprises2022} for reference.
Similar results hold for kernel regression and random feature regression-based architectures \citep{MeiGeneralization2022}, which are both widely used models in the analysis of neural networks. For consistent estimators, the generalization loss converges to the Bayes risk asymptotically, although this decrease need not be strictly monotonic. Estimators like Empirical Risk Minimization (ERM), Structural Risk Minimization (SRM) are consistent under mild assumptions on the hypothesis class, e.g., having finite capacity.

The expected training loss $\E[R_{\text{train}}(S_n,t)]$ is non-decreasing: this holds for ridgeless linear regression in general, and therefore for kernel regression and random feature-based models of neural networks. Note that this assumption can be modified slightly to be $\E[R_{\text{train}}(S_{n-1},t)] \leq \E[R_{\text{train}}(S_n,t)]+B/n$. This new condition holds for empirical risk minimization (ERM) with bounded loss $|\ell(w,z)|\leq B$ in general. The resulting left-hand side of the inequality in Lemma 3 gets an additive term of $B/n$ correspondingly. The rest of our calculations stay as they are.
The expected generalization gap $\E [\d R(S_n,t)]$ is non-negative: This holds for empirical risk minimization (ERM) in general.

As we show next, the concentration of $\bar \D_n(t)$ to $\E\sbr{\bar \D_n(t)}$ can also be guaranteed if algorithm stability is assumed.

\paragraph{Concentration of $\bar \D_n(t)$ to $\mathbb{E}[\bar \D_n(t)]$}

We first define the notion of stability for a deterministic algorithm $\mathcal{A}$ that maps from space of datasets to weight space, i.e. $\mathcal{A}: \cup_{n=0}^\infty \ZZ^n \to \WW$.

\begin{definition}
An algorithm $\mathcal{A}$ is uniformly $\ve$-stable if for all datasets $S$, $S'$ differing in at most one sample, we have
\aeqs{
\sup_z \abs{\ell(\mathcal{A}(S),z)-\ell(\mathcal{A}(S'),z)} \leq \ve
}
\end{definition}
Now we define the set of algorithms $\Gamma$ that map the dataset $S$ to points on the gradient flow trajectory trained on $S$ at certain time points.
\aeqs{
\G = \bigg\{\mathcal{A}: \mathcal{A}(S) = w(t), t\geq 0, \text{ where $w$ satisfies } \dv{w}{t}=-\nabla \bar\ell(w,S), w(0)\in \WW, S\in \cup_{n\in\mathbb{N}} \ZZ^n \bigg\}
}

\begin{lemma}
\label{lem: concentration delta}
Assume that (1) $\abs{\ell(w,z)}\leq B$ for all $w \in \WW, z\in\ZZ$, (2) $\forall \mathcal{A}\in \Gamma$, $\mathcal{A}$ is $\ve$-stable, then for all $t>0$, with probability $1-\delta$,
\aeqs{
\abs{\bar \D_n(t) - \mathbb{E}[\bar \D_n(t)]} \leq (n\ve+2B) \sqrt{\f{2\log(2/\delta)}{n}}
}
\end{lemma}

\begin{proof}
Let $\tilde{S}_n$ denote a modified dataset of $S_n$ by replacing the sample $z_j$ with a different sample $\tilde z_j$. Let $\tilde w_n(t)$, $\tilde w_n^{-i}(t)$ be the corresponding trajectories trained with $\tilde{S}_n$ and $\tilde S_n^{-i}$ (the removed-$i$th sample version of $S_n$). Note that $\tilde w_n^{-j}(t)= w_n^{-j}(t)$. Let $\bar \D(\tilde S_n,t)$ and $\bar \D(S_n,t)$ be the averaged loss difference calculated on $\tilde S_n$ and $S_n$ respectively. By assumptions (1) and (2), we have 
\aeqs{
|\bar \ell(\tilde w_n(t),\tilde S_n) - \bar \ell(w_n(t),S_n)| &\leq \f{(n-1)\ve}{n} + \f{2B}{n} \leq \ve + \f{2B}{n}\\
|\ell(\tilde w_n^{-i}(t),z_i) - \ell(w_n^{-i}(t),z_i)| &\leq \ve \quad \forall i\neq j\\
|\ell(\tilde w_n^{-j}(t),z_j) - \ell(w_n^{-j}(t),z_j)| &\leq \f{2B}{n}
}
Hence we have
\aeq{
\label{eq: diff replace}
\abs{\bar \D(\tilde S_n,t) - \bar \D(S_n,t)} \leq 2\ve + \f{4B}{n}
}

Inequality \cref{eq: diff replace} gives the replace-one-sample difference of $\bar \D_n$, hence by McDiarmid's inequality \citep{McdiarmidOn1989}, we have the following concentration inequality,
\aeqs{
\mathbb{P}_{S_n} \sbr{|\bar\D_n - \mathbb{E}[\bar \D_n]| \geq a} \leq 2\exp\rbr{-\f{2a^2}{n(2\ve + 4B/n)^2}}
}
Setting the right-hand side to $\delta$, we have with probability at least $1-\delta$,
\aeqs{
\abs{\bar \D_n - \mathbb{E}[\bar \D_n]} \leq (n\ve+2B) \cdot \sqrt{\f{2\log(2/\delta)}{n}}.
}
\end{proof}

\begin{remark}
In general, the convergence of $\bar \D_n$ to $\mathbb{E}\sbr{\bar \D_n}$ can be guaranteed by different versions of algorithm stability (e.g. hypothesis stability, pointwise hypothesis stability and uniform stability \citep{BousquetStability2002}).
\cite{CharlesStability2018} shows that
the algorithm $\mathcal{A}$ is $C(L,\mu)/(n-1)$-uniformly stable if $\ell(w,z)$ is $L$-Lipchitz in $w$ and $\bar\ell(w,S)$ is $\mu$-PL (Polyak Lojasiewicz) in $w$, where $C(L,\mu)$ is a constant depending on $L$ and $\mu$. Other versions of stability can also be guaranteed by PL and QG (quadratic growth) conditions as shown in \cite{CharlesStability2018}. The averaged loss difference $\bar \Delta_n$ is in nature similar to cross-validation loss estimator. Under certain stability conditions, \cite{AusternAsymptotics2020} proves central limit theorems and Berry-Esseen bounds for the cross validation loss estimator.
\end{remark}

\subsection{Proof of \cref{lem: contraction}}

By taking the derivative of the pointwise loss difference $\Delta_n^{-i}(t)$, we have,

\aeqs{
\f{\dd \D_n^{-i}(t)}{\dd t} &= \f{\dd \rbr{ \ell(w_n^{-i}(t),z_i) - \ell(w_n(t),z_i)}}{\dd t}\\
&= -\nabla \ell(w,z_i) \cdot \nabla \bar \ell(w,S_n^{-i}) \big |_{w_n^{-i}(t)} - \rbr{-\nabla \ell(w,z_i) \cdot \nabla \bar \ell(w,S_n) \big|_{w_n(t)}}\\
&= -\rbr{\nabla \ell(w,z_i) \cdot \nabla \bar \ell(w,S_n^{-i}) \big |_{w_n^{-i}(t)} - \nabla \ell(w,z_i) \cdot \nabla \bar \ell(w,S_n^{-i}) \big |_{w_n(t)}}\\
&+ \rbr{-\nabla \ell(w,z_i) \cdot \nabla \bar \ell(w,S_n^{-i}) \big |_{w_n(t)} + \nabla \ell(w,z_i) \cdot \nabla \bar \ell(w,S_n) \big|_{w_n(t)}}\\
&= -\rbr{\nabla \ell(w,z_i) \cdot \nabla \bar \ell(w,S_n^{-i}) \big |_{w_n^{-i}(t)} - \nabla \ell(w,z_i) \cdot \nabla \bar \ell(w,S_n^{-i}) \big |_{w_n(t)}}\\
&+ \nabla \ell(w,z_i) \rbr{\nabla \bar \ell(w,S_n)-\nabla \bar \ell(w,S_n^{-i})} \big |_{w_n(t)}
}
Hence, 
\aeqs{
    \dv{\D^{\bs i}_n(t)}{t}= -c_n^{\bs i}(t)\D_n^{\bs i}(t)+\e_n^{\bs i}(t),
}
where 
\aeqs{
c_n^{\bs i}(t)=\f{\nabla \ell(w,z_i) \cdot \nabla \bar \ell(w,S_n^{\bs i}) \big|^{w_n^{\bs i}(t)}_{w_n(t)}}{\D_n^{\bs i}(t)},
}
and 
\aeqs{
\e_n^{\bs i}(t)=\nabla \ell(w,z_i) \cdot \rbr{\nabla \bar \ell(w,S_n) - \nabla \bar \ell (w,S_n^{\bs i})} \bigg |_{w_n(t)}.
}

\subsection{Evolution of $\bar \D_n(t)$}
\label{app:s: evo gap}

The evolution of $\bar \D_n(t)$ can be derived through that of $\D_n^{-i}(t)$.
\aeqs{
\f{\dd \bar \D_n(t)}{\dd t} &= \f{1}{n}\sum_{i=1}^n \f{\dd \D_n^{-i}(t)}{\dd t}\\
&= -\f{1}{n}\sum_{i=1}^n \rbr{c_n^{-i}(t)\D_n^{-i}(t) + \e_n^{-i}(t)}\\
&= - \f{\f{1}{n}\sum_{i=1}^n c_n^{-i}(t)\D_n^{-i}(t)}{\bar \D_n(t)} \bar \D_n(t) + \f{1}{n}\sum_{i=1}^n \e_n^{-i}(t)\\
&= -\bar c_n(t) \bar\D_n(t) + \bar\e_n(t).
}
Here we have,
\aeqs{
\bar c_n(t) &= \f{\f{1}{n}\sum_{i=1}^n c_n^{-i}(t)\D_n^{-i}(t)}{\bar \D_n(t)}\\
&= \f{\frac{1}{n}\sum_{i=1}^n \nabla \ell(w,z_i)\cdot \nabla \bar \ell(w,S_n^{\bs i}) \big |_{w_n(t)}^{w_n^{\bs i}(t)}}{\bar \D_n(t)},
}
and
\aeqs{
\bar \e_n(t) &= \f{1}{n}\sum_{i=1}^n \e_n^{-i}(t)\\
&= \f{1}{n} \sum_{i=1}^n \nabla \ell_i^\top \rbr{\f{1}{n}\sum_{j=1}^n \nabla \ell_j - \f{1}{n-1}\sum_{j\neq i}\nabla \ell_j}\\
&= \f{1}{n} \sum_{i=1}^n \nabla \ell_i^\top \rbr{\f{1}{n}\nabla \ell_i - \frac{1}{n(n-1)}\sum_{j\neq i}\nabla \ell_j}\\
&= \frac{1}{n^2} \sum_{i=1}^n \nabla \ell_i^\top \nabla \ell_i - \frac{1}{n^2(n-1)} \sum_{i\neq j}\nabla \ell_i^\top \nabla \ell_j.
}
In the calculation above, we use $\nabla \ell_i$, $\nabla \bar \ell$ as an abbreviation for $\nabla \ell(w_n(t),z_i)$, $\nabla \bar \ell(w_n(t),S_n)$ respectively.
Notice that we have the following decomposition of the gradient covariance matrix $\hat\S(t)$:
\aeqs{
\hat \S (t) &= \frac{1}{n}\sum_{i=1}^n \rbr{\nabla \ell_i - \nabla \bar \ell}\rbr{\nabla \ell_i - \nabla \bar \ell}^\top\\
&= \frac{1}{n}\sum_{i=1}^n \nabla \ell_i \nabla \ell_i^\top - \frac{1}{n^2} \rbr{\sum_{i=1}^n \nabla \ell_i} \rbr{\sum_{i=1}^n \nabla \ell_i}^\top\\
&= \f{n-1}{n^2} \sum_{i=1}^n \nabla \ell_i \nabla \ell_i^\top - \f{1}{n^2} \sum_{i\neq j} \nabla \ell_i \nabla \ell_j^\top
}
Hence, we have 
\aeqs{
\bar \e_n(t) = \f{\tr \hat \S(t)}{n-1}, \quad \hat \S_n(t)=\cov_{z\sim \text{Unif}(S_n)}\nabla \ell(w_n(t),z),
}
where $\hat \S_n(t)$ represents the covariance matrix of $\nabla \ell(w_n(t),z)$ for $z$ sampled uniformly from the dataset $S_n$.

\paragraph{Evolution of $\E\sbr{\bar\D_n(t)}$:}
A modified version of $\bar c_n$ and $\bar \e_n$, 
\aeqs{
\bar c_n = \f{ \E \sbr{\frac{1}{n}\sum_{i=1}^n \nabla \ell(w,z_i)\cdot \nabla \bar \ell(w,S_n^{\bs i}) \big |_{w_n(t)}^{w_n^{\bs i}(t)}}}{\E \sbr{\bar \D_n(t)}}, \quad
\bar \e_n = \f{ \E \sbr{\tr \hat \S(t)}}{n-1},
}
gives the evolution of $\mathbb{E}\sbr{\bar \D_n(t)}$,
\aeqs{
\dv{\mathbb{E}\sbr{\bar \D_n(t)}}{t} = -\bar c_n(t) \mathbb{E}\sbr{\bar \D_n(t)} + \bar\e_n(t).
}

\subsection{Proof of \cref{lem: contraction approx}}

%
%
%

Let $h(u) = w_n + u \delta w_i$ for $u \in [0, 1]$ parameterize the line segment between $w_n$ and $w_n^{-i}$. Let $\phi(w) = \nabla \ell(w, z_i)^\top \nabla \overline{\ell}(w, S_n^{-i})$ be the inner product of the gradients. The numerator of $c_n^{-i}$ is $\phi(w_n^{-i}) - \phi(w_n)$. We prove the lemma by Taylor expansion of the numerator and the denominator.

The assumptions imply that $\nabla\phi$ is Lipschitz with a constant depending only on $G,L_1,L_2$. By Taylor's theorem and the product rule,
\begin{align*}
\phi(w_n^{-i})-\phi(w_n)
&=A_i+O(d_i^2),\\
A_i
&=g^{-i}\cdot H_i\delta w_i+g_i\cdot H^{-i}\delta w_i,
\end{align*}
where $d_i=\norm{\delta w_i}_2$. Similarly,
\begin{equation*}
\ell(w_n^{-i},z_i)-\ell(w_n,z_i)
=B_i+O(d_i^2),
\qquad
B_i=g_i\cdot\delta w_i.
\end{equation*}
Since $A_i=O(d_i)$ and $\abs{B_i}\geq\iota_{\rm nd}d_i$, division gives
\begin{equation*}
c_n^{-i}
=\frac{A_i+O(d_i^2)}{B_i+O(d_i^2)}
=\frac{A_i}{B_i}+O(d_i)
=\hat c_1^{-i}+\hat c_2^{-i}+O(d_i).
\end{equation*}

For the averaged contraction factor, let $d_{\max}=\max_i d_i$. Summing the expansions above gives
\begin{equation*}
\bar c_n
=\frac{\sum_i A_i+O(\sum_i d_i^2)}
       {\sum_i B_i+O(\sum_i d_i^2)}.
\end{equation*}
Using $\sum_i d_i^2\leq d_{\max}\sum_i d_i$ and the averaged non-degeneracy condition, we obtain
\begin{equation*}
\bar c_n
=\frac{\sum_i A_i}{\sum_i B_i}+O(d_{\max})
=\hat c_1+\hat c_2+O(d_{\max}),
\end{equation*}
which proves the result.

\subsection{Evolution of $\vec{r}_n(t)$}

We derive the equation governing the evolution of $\vec{r}_n(t)$ by calculating its time derivative.

\aeqs{
\f{\dd \vec{r}_n(t)}{\dd t} 
&= \frac{1}{\sqrt{n}}
\bmat{
\dv{r(w_n(t),z_1)}{t}\\
...\\
\dv{r(w_n(t),z_n)}{t}
}
= \frac{1}{\sqrt{n}}
\bmat{
\nabla r(w_n(t),z_1)^\top \dv{w_n(t)}{t}\\
...\\
\nabla r(w_n(t),z_n)^\top \dv{w_n(t)}{t}
}\\
&= \frac{1}{\sqrt{n}}
\bmat{
-\f{1}{n}\sum_{j=1}^n\nabla r(w_n(t),z_1)^\top \nabla f(w_n(t),x_j) \cdot r(w_n(t),z_j)\\
...\\
-\f{1}{n}\sum_{j=1}^n\nabla r(w_n(t),z_n)^\top \nabla f(w_n(t),x_j)\cdot r(w_n(t),z_j)
}
= -\f{1}{n} P_n(t) \vec{r}_n(t).
}
Here the third equality follows from the evolution of $w_n(t)$:
\aeqs{
\dv{w_n(t)}{t} &= -\nabla \ell(w_n(t),S_n)\\
&= -\frac{1}{n}\sum_{i=1}^n  \nabla f(w_n(t),x_i) \dv{\ell(f(w_n(t),x_i),y_i)}{f(w_n(t),x_i)}\\ 
&= -\frac{1}{n}\sum_{i=1}^n \nabla f(w_n(t),x_i) r(w_n(t),x_i).
}
Note that
\aeqs{
\nabla r(w,z_i), \nabla f(w,z_i) &\in \WW \times \YY\\
P_n(t) = \sbr{\nabla r(w_n(t),z_i)^\top \nabla f(w_n(t),x_j)}_{i,j\in[n]} &\in \YY^n \times \YY^n.
}

\subsection{Proof of \cref{lem: res}}

We have the following decomposition of the gradient covariance $\hat \S_n(t)$:
\aeqs{
\hat \S_n (t) &= \frac{1}{n}\sum_{i=1}^n \rbr{\nabla \ell_i - \nabla \bar \ell}\rbr{\nabla \ell_i - \nabla \bar \ell}^\top\\
&= \frac{1}{n}\sum_{i=1}^n \nabla \ell_i \nabla \ell_i^\top - \nabla \bar \ell \nabla \bar \ell^\top
}
where
\aeqs{
\nabla \ell_i &= \nabla f(w_n(t),x_i) r_i(t)\\
\nabla \bar \ell &= \frac{1}{n}\sum_{i=1}^n \nabla f(w_n(t),x_i) r_i(t).
}
Taking traces and using the normalized residual in \cref{eq:normalized_residual}, we obtain
\aeqs{
\tr\hat \S_n(t)
&= \frac{1}{n}\sum_{i=1}^n r_i(t)^2\nabla f(w_n(t),x_i)^\top\nabla f(w_n(t),x_i)\\
&\quad-\frac{1}{n^2}\sum_{i,j=1}^n r_i(t)r_j(t)\nabla f(w_n(t),x_i)^\top\nabla f(w_n(t),x_j)\\
&= \vec{r}_n(t)^\top F_n(t)\vec{r}_n(t),
}
where the diagonal and off-diagonal blocks of $F_n(t)$ are given in \cref{eq: F}.

\subsection{Proof of \cref{thm: eff ker}}

By \cref{eq: solu res} and \cref{lem: res},
\beq{
\tr\hat \S_n(t) = \vec{r}_n(0)^\top \Omega_n(t)^\top F_n(t) \Omega_n(t) \vec{r}_n(0).
}
Combining with the solution of $\bar \D_n(t)$ \cref{eq: solu gap}, we have
\aeqs{
\bar \D_n(t) = \vec{r}_n(0)^\top \rbr{ \frac{\int_{0}^t \Omega_n(s)^\top F_n(s) \Omega_n(s) \exp\rbr{-\int_s^t\bar c_n(u) \dd u} \dd s}{n-1} } \vec{r}_n(0)
}
Hence, we have 
\aeqs{
\bar \D_n(t)
= \vec{r}_n(0)^\top K_n(0,t) \vec{r}_n(0)
}
where 
\aeqs{
K_n(0,t) = \frac{\int_{0}^t \Omega_n(s)^\top F_n(s) \Omega_n(s) \exp\rbr{-\int_s^t\bar c_n(u) \dd u} \dd s}{n-1}.
}
For any $q=[q_1,\dots,q_n]\in\YY^n$, let $a_i=\nabla f(w_n(s),x_i)q_i$. Then
\aeqs{
q^\top F_n(s)q
&=\sum_{i=1}^n\norm{a_i}_2^2-\frac{1}{n}\norm{\sum_{i=1}^n a_i}_2^2\geq 0,
}
where the inequality follows from Cauchy--Schwarz. Thus $F_n(s)$ is PSD for every $s$, and $K_n(0,t)$ is PSD as an integral of PSD matrices with non-negative scalar weights.

\subsection{Proof of~\cref{lem: cvg ker}}

For any $x\in\YY^n$, let $y(t)=\Omega_n(t)x$. Using $\dd\Omega_n(t)/\dd t=-P_n(t)\Omega_n(t)/n$, we have
\aeqs{
\dv{\norm{y(t)}_2^2}{t}
&=-\frac{1}{n}y(t)^\top\rbr{P_n(t)+P_n(t)^\top}y(t)
\leq-\frac{2\l_{\min}(t)}{n}\norm{y(t)}_2^2.
}
Since $\Omega_n(0)=I$, we have $y(0)=x$. Grönwall's inequality gives
\[
\norm{y(t)}_2^2\leq\omega(t)\norm{x}_2^2.
\]
Since this holds for every $x$, taking the supremum over $\norm{x}_2=1$ gives $\norm{\Omega_n(t)}_2^2\leq\omega(t)$. Let
\[
A(t)=\frac{\Omega_n(t)^\top F_n(t)\Omega_n(t)}{n-1},
\qquad
\tilde c(s,t)=\exp\rbr{-\int_s^t\bar c_n(u)\dd u}.
\]
Then $\norm{A(t)}_2\leq\omega(t)m(t)$. Moreover, since $\bar c_n(t)\geq0$, for every fixed $s$, $\tilde c(s,t)$ is non-increasing in $t$ and converges to a limit $\tilde c_\infty(s)\in[0,1]$.

We can write
\[
K_n(0,t)=\int_0^\infty 1_{[0,t]}(s)A(s)\tilde c(s,t)\dd s.
\]
The integrand converges pointwise to $A(s)\tilde c_\infty(s)$ and its matrix 2-norm is bounded by the integrable function $\omega(s)m(s)$. Therefore, by the dominated convergence theorem and condition (i),
\[
\lim_{t\to\infty}\norm{K_n(0,t)-\int_0^\infty A(s)\tilde c_\infty(s)\dd s}_2=0,
\]
which proves the result.

\subsection{Contraction and perturbation factors for the omitting-$m$-samples experiments}
\label{app: omit m}

In the omitting $m$-samples setting, by similar calculations as in \cref{app: methods}, the batch-wise contraction and perturbation factors are
\aeqs{
c_n^{\bs (m)}(t)=\f{\nabla \bar\ell(w,S_{(m)}) \cdot \nabla \bar \ell(w,S_n^{\bs (m)}) \big|^{w_n^{\bs (m)}(t)}_{w_n(t)}}{\D_n^{\bs (m)}(t)},
}
\aeqs{
\e_n^{\bs (m)}(t)=\nabla \bar\ell(w,S_{(m)}) \cdot \rbr{\nabla \bar \ell(w,S_n) - \nabla \bar \ell (w,S_n^{\bs (m)})} \bigg |_{w_n(t)}.
}
The averaged contraction and perturbation factors are
\aeqs{
\bar c_n(t) = \f{\E_{(m)}\sbr{\nabla \bar\ell(w,S_{(m)}) \cdot \nabla \bar \ell(w,S_n^{\bs (m)}) \big|^{w_n^{\bs (m)}(t)}_{w_n(t)}}}{\bar \D_n(t)},
}

\aeqs{
\bar \e_n(t) = \f{\tr \hat \S(t)}{n-1}, \quad \hat \S_n(t)=\cov_{z\sim \text{Unif}(S_n)}\nabla \ell(w_n(t),z),
}
where $\hat \S_n(t)$ represents the covariance matrix of $\nabla \ell(w_n(t),z)$ for $z$ sampled uniformly from the dataset $S_n$. Note that the averaged perturbation factor $\bar \e_n(t)$ in the omit-$m$ setting is independent of $m$.

\section{Proofs and Calculations in \cref{s: examples}}
\label{app:s: examples}

\subsection{Proof of \cref{lem: eg high dim con pert}}
\label{app: eg high dim con pert}
In this proof, we adopt the following notaion. The leave-one-out Hessian is denoted as $H^{-i} = \sum_{j\neq i} x_j x_j^\top / (n-1)$, the residual is denoted as  $r_i = w_n^\top x_i - y_i$, and the parameter shift is denoted as $\delta w_i = w_n^{-i} - w_n$.

\textbf{The evolution of the parameter shift.}
The evolution of parameter shift $\delta w_i$ follows
\[
\aed{
    \f{\dd \delta w_i}{\dd t} &= \f{\dd w_n^{-i}}{\dd t} - \f{\dd w_n}{\dd t}\\
    &= -H^{-i} \delta w_i - \frac{1}{n} \left( \nabla \bar \ell(w_n,S_n^{-i}) - x_i r_i \right).
}
\]

\textbf{The exact formulation of contraction and perturbation.}
We then give the exact formula for contraction and perturbation. The numerator of the contraction can be expressed in the following way,
\[
\aed{
\text{Num}_n &= \f{1}{n} \sum_{i=1}^n \nabla \ell(w,z_i) \cdot \nabla \bar \ell(w,S_n^{-i}) \big |^{w_n^{-i}(t)}_{w_n(t)}\\
&= \frac{1}{n} \sum_{i=1}^n \left( r_i\left( x_i^\top H^{-i} \delta w_i \right)
- \left( \frac{n}{n-1} \dot{r}_i + \frac{\|x_i\|^2}{n-1} r_i \right) \left( x_i^\top \delta w_i \right)
+ \left( x_i^\top \delta w_i \right) \left( x_i^\top H^{-i} \delta w_i \right) \right).
}
\]
The denominator of the contraction factor can be expressed as
\[
\bar{\Delta}_n = \frac{1}{n} \sum_{i=1}^n r_i(x_i^\top \delta w_i) + \frac{1}{2n} \sum_{i=1}^n (x_i^\top \delta w_i)^2.
\]
The perturbation factor can be expressed as
\[
\bar{\epsilon}_n = \frac{\tr(\hat{\Sigma}_n)}{n-1} = \frac{1}{n-1} \left( \frac{1}{n} \sum_{i=1}^n r_i^2 \|x_i\|^2 - \left\| \frac{1}{n} \sum_{j=1}^n r_j x_j \right\|^2 \right).
\]

\textbf{The asymptotics of contraction and perturbation for large $n$.}
We now analyze the large $n$ limit of the contraction and perturbation factors.
We first define several terms that are essential components of the asymptotic formulation. We define $Q(t,s)$ to be the asymptotic residual product,

\beq{
\label{eq: q}
Q(t,s) \triangleq \lim_{n \to \infty} \frac{1}{n} \sum_{i=1}^n r_i(t) r_i(s).
}
We also use the following term in the derivation,
\beq{
\label{eq: dq}
\partial_t Q(t,s) = \lim_{n \to \infty} \frac{1}{n} \sum_{i=1}^n \dot{r}_i(t) r_i(s).
}
We define the zeroth-order and first-order spectral moment as
\begin{equation}
\label{eq: spectral moment}
D(\tau) \triangleq \lim_{n \to \infty} \frac{1}{n} \text{Tr}(e^{-H\tau}),
\qquad
N(\tau) \triangleq \lim_{n \to \infty} \frac{1}{n} \text{Tr}(H e^{-H\tau}),
\end{equation}
respectively. 
%
%

By observing that for a matrix $M$ with bounded operator norm w.r.t. $n$, 
\[
\frac{x_i^\top M \nabla \bar \ell(w_n,S_n^{-i})}{n} = \frac{\sum_{j\neq i}x_i^\top M x_jr_j}{n^2} = O(1/n),
\]
while 
\[
\frac{x_i^\top M x_ir_i}{n} = O(1),
\]
we can conclude that the modified evolution 
\[
\f{\dd \delta w_i}{\dd t} = -H^{-i} \delta w_i + \frac{x_i r_i}{n} 
\]
gives the same contraction factor in the limit of $n\to \infty$. This shows that the gradient average term in the drift of the evolution of parameter shift is negligible.

We should also notice that in the perturbation factor,
\[
\frac{1}{n} \sum_{i=1}^n r_i^2 \|x_i\|^2 = O(n), 
\quad
\left\| \frac{1}{n} \sum_{j=1}^n r_j x_j \right\|^2 = \frac{1}{n} r^\top \left( \frac{1}{n} X X^\top \right) r = O(1), 
\]
which implies the negligibility of the second term. Hence the modified expression 
\[
\bar{\epsilon}_n = \frac{\tr(\hat{\Sigma}_n)}{n-1} = \frac{1}{n(n-1)} \rbr{\sum_{i=1}^n r_i^2 \|x_i\|^2}.
\]
gives the same perturbation factor in the limit $n\to \infty$.

Substituting the previously established definitions from \cref{eq: q,eq: dq,eq: spectral moment}, we can express the limits of the contraction factor's numerator and denominator, along with the perturbation factor, as follows,
\[
\aed{
\text{Num}_\infty(t) &= \int_0^t N(t-s) Q(t,s) \dd s - \int_0^t D(t-s) \partial_t Q(t,s) \dd s \nonumber \\
&\quad - \gamma \int_0^t D(t-s) Q(t,s) \dd s + \int_0^t \int_0^t D(t-s_1) N(t-s_2) Q(s_1, s_2) \dd s_1 \dd s_2,\\
\bar \Delta_\infty(t) &= \int_0^t D(t-s) Q(t,s) \dd s + \frac{1}{2} \int_0^t \int_0^t D(t-s_1) D(t-s_2) Q(s_1, s_2) \dd s_1 \dd s_2,\\
\bar{\epsilon}_\infty(t) &= \gamma Q(t,t).
}
\]

\paragraph{The decay trend of the macroscopic limit of contraction and perturbation for large $t$.}

To analyze the asymptotic behavior of the contraction and perturbation factors, we derive analytical expressions for asymptotic residual product $Q$ and spectral moments $D,N$ (\cref{eq: q,eq: dq,eq: spectral moment}) using MP moments $M_0$ and $M_1$ (\cref{eq: MP moments}).

By invoking the Marchenko-Pastur theorem \cite{MPDistribution1967}, we derive analytical expressions for $D$, $N$ and $Q$ based on the MP moments $M_0$ and $M_1$.
\[
D(\tau) = 
\begin{cases} 
    \gamma M_0(\tau, \gamma) & \text{if } \gamma < 1 \\ 
    (\gamma - 1) + M_0(\tau, 1/\gamma) & \text{if } \gamma > 1 \\
    M_0(\tau, 1) & \text{if } \gamma = 1 
\end{cases},
\qquad
N(\tau) = 
\begin{cases} 
    \gamma M_1(\tau, \gamma) & \text{if } \gamma < 1 \\ 
    M_1(\tau, 1/\gamma) & \text{if } \gamma > 1 \\
    M_1(\tau, 1) & \text{if } \gamma = 1 
\end{cases},
\]

\begin{equation*}
Q(t, s) =
\begin{cases} 
      \sigma^2 \left[ (1-\gamma) + \gamma M_0(t+s, \gamma) \right] + \rho^2 \gamma M_1(t+s, \gamma) & \text{if } \gamma < 1 \\
      \sigma^2 M_0(t+s, 1/\gamma) + \rho^2 M_1(t+s, 1/\gamma) & \text{if } \gamma > 1 \\
      \sigma^2 M_0(t+s, 1) + \rho^2 M_1(t+s, 1) & \text{if } \gamma = 1
\end{cases}.
\end{equation*}

We now proceed to analyze the asymptotic behavior of the contraction and perturbation factors as $t \to \infty$. We have the following results on the integrals of MP moments for $0<\theta <1$.
\[
\int_0^\infty M_k(\tau, \theta) d\tau = \int_{\lambda_-}^{\lambda_+} \lambda^k \rbr{\int_0^\infty e^{-\lambda \tau} \dd \tau} f_{\text{MP}}(\lambda; \theta) d\lambda = \int_{\lambda_-}^{\lambda_+} \lambda^{k-1} f_{\text{MP}}(\lambda; \theta) d\lambda
\]
which implies 
\[
\int_0^\infty M_1(\tau, \theta) d\tau = 1,
\quad
\int_0^\infty M_0(\tau, \theta) d\tau =  \f{1}{1-\theta}.
\]
We also have for both $k=0,1$, these exists some constant $M$ such that,
\[
|M_k(\tau, \theta)| \le M e^{-\lambda_-\tau}.
\]

Now we are ready to estimate the decay pattern of the contraction and perturbation factors in different regimes.

\begin{enumerate}[nosep]
\item \textbf{Under-parameterized regime, $\gamma <1$.}
The residual product can be decomposed as 
\[
    Q(t,s) = Q_\infty + R(t,s)
\]
where $Q_\infty = \sigma^2(1-\gamma)$, and $|R(t,s)| \le M e^{-\lambda_-(t+s)}$ for some constant $M > 0$. Furthermore, because the residual reaches a steady state, its derivative vanishes, i.e. $\lim_{t, s \to \infty} \partial_t Q(t,s) = 0$.

We will now show that the integral in the contraction involving $R(t,s)$ vanishes.
For a single integral convolved with an operator $\mathcal{O}$ bounded by $C e^{-\lambda_- \tau}$,
\begin{align*}
    \left| \int_0^t \mathcal{O}(t-s) R(t,s) ds \right| &\le \int_0^t \left(C e^{-\lambda_-(t-s)}\right) \left(M e^{-\lambda_-(t+s)}\right) ds \\
    &= \int_0^t C M e^{-2\lambda_-t} ds = C M t e^{-2\lambda_-t}.
\end{align*}
For a double integral,
\[
\aed{
    &\left| \iint \mathcal{O}(t-s_1) \mathcal{O}(t-s_2) R(s_1, s_2) ds_1 ds_2 \right| \\
    &\le \iint \left(C e^{-\lambda_-(t-s_1)}\right) \left(C e^{-\lambda_-(t-s_2)}\right) \left(M e^{-\lambda_-(s_1+s_2)}\right) ds_1 ds_2 \leq C^2 M t^2 e^{-2\lambda_-t}.
}
\]
Hence, 
\[
\aed{
\lim_{t\to\infty}\text{Num}_\infty(t) 
&= Q_\infty \int_0^\infty \mathcal{N}(\tau) d\tau -\gamma Q_\infty \int_0^\infty \mathcal{D}(\tau) d\tau + Q_\infty \left( \int_0^\infty \mathcal{D}(\tau) d\tau \right) \left( \int_0^\infty \mathcal{N}(\tau) d\tau \right)\\
&= Q_\infty \gamma,\\
\lim_{t\to\infty}\bar \Delta_{\infty}(t)
&= Q_\infty \int_0^\infty \mathcal{D}(\tau) d\tau + \frac{1}{2} Q_\infty \left( \int_0^\infty \mathcal{D}(\tau) d\tau \right)^2
= Q_\infty \frac{2\gamma - \gamma^2}{2(1-\gamma)^2}.
}
\]
As a result, we have 
\[
\lim_{t\to\infty} \bar c_\infty(t) = \f{2(1-\gamma)^2}{2-\gamma}
\]
Similarly, for the perturbation factor, we derive the following limit,
\[
\lim_{t\to\infty}\bar \e_{\infty}(t) = \lim_{t\to\infty} \gamma Q(t,t) = \sigma^2\gamma(1-\gamma).
\]

\item \textbf{Over-parameterized regime, $\gamma > 1$.}
We first analyze the numerator. We should note that $D(t) \to D_\infty$, which is lower bounded, $N(t)$ follows exponential decay with respect to $t$, and $Q(t,s)$ follows exponential decay on $t+s$. Here is the asymptotic order of all the four terms in $\text{Num}_\infty(t)$.
\begin{equation*}
    \int_0^t N(t-s) Q(t,s) ds \le C \int_0^t e^{-\lambda_-(t-s)} e^{-\lambda_-(t+s)} ds = C \int_0^t e^{-2\lambda_- t} ds = \Theta(t e^{-2\lambda_- t})
\end{equation*}
\begin{equation*}
    \int_0^t D(t-s) |\partial_t Q(t,s)| ds \le C \int_0^t e^{-\lambda_-(t+s)} ds = \Theta(e^{-\lambda_- t})
\end{equation*}
\begin{equation*}
    \gamma \int_0^t D(t-s) Q(t,s) ds \le C \int_0^t e^{-\lambda_-(t+s)} ds = \Theta(e^{-\lambda_- t}) 
\end{equation*}
\begin{align*}
    \iint D(t-s_1) N(t-s_2) Q(s_1, s_2) ds_1 ds_2 &\le C \iint e^{-\lambda_-(t-s_2)} e^{-\lambda_-(s_1+s_2)} ds_1 ds_2 \\
    &= C t e^{-\lambda_- t} \left( \frac{1 - e^{-\lambda_- t}}{\lambda_-} \right)= \Theta(t e^{-\lambda_- t})
\end{align*}
Hence, we have
\[
\text{Num}_\infty(t) = \Theta(te^{-\lambda_{-}t}).
\]
For the denominator $\bar \Delta_n(t)$, the first term vanishes, and the second term converges to 
\[
\frac{D_\infty^2}{2} \int_0^t \int_0^t Q(s_1, s_2) ds_1 ds_2,
\]
which is constant. 
Hence, we have 
\[
\bar c_\infty = \Theta(te^{-\lambda_{-}t}).
\]
For the perturbation factor, by the exponential decay of $M_0$ and $M_1$, we have 
\[
\bar \e_\infty(t) = \gamma Q(t,t) = \Theta(e^{-2\lambda_- t}).
\]

\item \textbf{Critical regime, $\gamma = 1$.}
For the Marchenko-Pastur distribution $f_{\text{MP}}(\lambda, \theta)$, when $\theta = 1$, we have,
\begin{equation*}
    f_{\text{MP}}(\lambda, 1) = \frac{1}{2\pi} \sqrt{\frac{4-\lambda}{\lambda}} \sim \frac{1}{\pi\sqrt{\lambda}} \quad \text{for $\lambda$ close to 0}.
\end{equation*}
By applying Laplace's method, the asymptotic behavior of the spectral moments $D, N$ is dominated by the density near the origin,
\begin{align*}
    D(\tau) &= \int_0^4 e^{-\lambda \tau} \dd f_{\text{MP}}(\lambda,1) \sim \int_0^\epsilon \frac{e^{-\lambda \tau}}{\pi\sqrt{\lambda}} \dd \lambda = \Theta(\tau^{-1/2}),\\
    N(\tau) &= \int_0^4 \lambda e^{-\lambda \tau} \dd f_{\text{MP}}(\lambda,1) \sim \int_0^\epsilon \frac{\lambda e^{-\lambda \tau}}{\pi\sqrt{\lambda}} \dd \lambda = \Theta(\tau^{-3/2}),
\end{align*}
where the last equality is derived from a change of variable $u = \lambda \tau$.
Similarly, we have $Q(\tau) = \Theta(\tau^{-1/2})$.

We will first analyze the decay of the denominator.
The denominator $\bar{\Delta}_\infty(t)$ is dominated by the double integral term. Let $s_1 = t u_1$ and $s_2 = t u_2$ to scale the integration variables out of the time domain, we have
\begin{align*}
    &\iint_0^t D(t-s_1) D(t-s_2) Q(s_1, s_2) \dd s_1 \dd s_2 \\
    \sim &\iint_0^t (t-s_1)^{-1/2} (t-s_2)^{-1/2} (s_1+s_2)^{-1/2} \dd s_1 \dd s_2 \\
    = &t^2 \iint_0^1 (t - tu_1)^{-1/2} (t - tu_2)^{-1/2} (tu_1 + tu_2)^{-1/2} \dd u_1 \dd u_2\\
    = &t^2 \cdot t^{-3/2} \iint_0^1 (1-u_1)^{-1/2} (1-u_2)^{-1/2} (u_1+u_2)^{-1/2} \dd u_1 \dd u_2
\end{align*}
By the change of variable $s_1 = t u_1$ and $s_2 = t u_2$, we get the first equality by scaling the integration variables out of the time domain. The second inequality can be achieved by factoring $t$ out of the integrand.
Because the remaining integral over $u_1, u_2$ converges to a strictly positive constant, the denominator diverges as,
\begin{equation*}
    \bar{\Delta}_\infty(t) = \Theta(t^{1/2})
\end{equation*}

Now we analyze the decay of the numerator.
We apply the identical scaling procedure to the dominant double integral of the numerator $\text{Num}_\infty(t)$. Crucially, the first order spectral moment $\mathcal{N}(\tau)$ decays much faster ($\Theta(\tau^{-3/2})$) due to the suppression of the zero-eigenvalue singularity:
\begin{align*}
    &\iint_0^t D(t-s_1) N(t-s_2) Q(s_1, s_2) \dd s_1 \dd s_2\\
    \sim& \iint_0^t (t-s_1)^{-1/2} (t-s_2)^{-3/2} (s_1+s_2)^{-1/2} \dd s_1 \dd s_2 \\
    =& t^2 \iint_0^1 (t - tu_1)^{-1/2} (t - tu_2)^{-3/2} (tu_1 + tu_2)^{-1/2} \dd u_1 \dd u_2\\
    =& t^2 \cdot t^{-5/2} \iint_0^1 (1-u_1)^{-1/2} (1-u_2)^{-3/2} (u_1+u_2)^{-1/2} \dd u_1 \dd u_2
\end{align*}
Hence, we have
\begin{equation*}
    \text{Num}_\infty(t) = \Theta (t^{-1/2}).
\end{equation*}
The calculations above yield the asymptotic decay order of the contraction factor
\begin{equation*}
    \bar{c}_\infty(t) = \frac{\Theta(t^{-1/2})}{\Theta(t^{1/2})} = \Theta(t^{-1}).
\end{equation*}

For the perturbation factor, we can similarly show that 
\begin{equation*}
    Q(t,t) = C_1 t^{-1/2} + C_2t^{-3/2} = \Theta(t^{-1/2}).
\end{equation*}

\end{enumerate}

\subsection{Proof of \cref{lem: eg two pt}}
\label{app: eg two pt}

The gradients for the averaged loss $\bar \ell(w,S_n)$ and $\bar \ell(w,S_n^{-i})$ are
\aeqs{
\nabla \bar \ell(w,S_n) &= \frac{1}{2}\rbr{w^\top x_1-y_1}x_1 + \frac{1}{2}\rbr{w^\top x_2-y_2}x_2\\
\nabla \bar \ell(w,S_n^{-1}) &= \frac{n-2}{2(n-1)}\rbr{w^\top x_1-y_1}x_1 + \frac{n}{2(n-1)}\rbr{w^\top x_2-y_2}x_2\\
\nabla \bar \ell(w,S_n^{-2}) &= \frac{n}{2(n-1)}\rbr{w^\top x_1-y_1}x_1 + \frac{n-2}{2(n-1)}\rbr{w^\top x_2-y_2}x_2.
}
The averaged contraction factor can be calculated from \cref{eq: c},
\aeqs{
\bar c_n(t) &= \f{\f{1}{n}\sum_{i=1}^n\nabla \ell(w,z_i)\cdot \nabla \bar\ell(w,S_n^{-i}) \big|^{w_n^{-i}(t)}_{w_n(t)}}{\bar \D_n(t)}\\
&= \f{\f{1}{2}\rbr{\nabla \ell(w,z_1)\cdot \nabla \bar\ell(w,S_n^{-1})\big|^{w_n^{-1}(t)}_{w_n(t)} + \nabla \ell(w,z_2)\cdot \nabla \bar\ell(w,S_n^{-2})\big|^{w_n^{-2}(t)}_{w_n(t)}}}{\f{1}{2}\rbr{\f{1}{2}(w^\top x_1-y_1)^2\big|^{w_n^{-1}(t)}_{w_n(t)} + \f{1}{2}(w^\top x_2-y_2)^2\big|^{w_n^{-2}(t)}_{w_n(t)}}}\\
&= \f{\f{1}{2}\rbr{\f{n-2}{2(n-1)}(w^\top x_1-y_1)^2\big|^{w_n^{-1}(t)}_{w_n(t)} + \f{n-2}{2(n-1)}(w^\top x_2-y_2)^2\big|^{w_n^{-2}(t)}_{w_n(t)}}}{\f{1}{2}\rbr{\f{1}{2}(w^\top x_1-y_1)^2\big|^{w_n^{-1}(t)}_{w_n(t)} + \f{1}{2}(w^\top x_2-y_2)^2\big|^{w_n^{-2}(t)}_{w_n(t)}}}
= \f{n-2}{2(n-1)}=: \bar c.
}

From \cref{eq: evo res} and \cref{eq: F} we can derive that
\aeqs{
F_n(t)=I_n-\frac{P_n(t)}{n},\qquad
P_n(t) = \diag\rbr{\vec{1}\vec{1}^\top, \vec{1}\vec{1}^\top},
}
with $\vec{1}=[1,\dots,1]\in \mathbb{R}^{n/2}$.
Note that $P_n(t)$ is not full rank when $n>2$. By \cref{lem: cvg ker}, the convergence of $\lim_{t\to\infty}K_n(0,t)$ is largely controlled by the smallest eigenvalue of $P_n(t)$, which cannot be too small. Hence, to ensure convergence, we define a modified version of $P_n(t)$ with small perturbation $\ve(t)$ on its singular subspace, i.e., $P_n^\ve(t)=U\L^\ve U^\top$, with $\L^\ve = \diag \rbr{n/2,n/2,n\ve(t)/2,\dots,n\ve(t)/2}$, $U=[u_1,\dots u_n]$, where
\aeqs{
u_1= \sqrt{\f{2}{n}}\ [1,\dots,1,0,\dots,0], \quad
u_2=\sqrt{\f{2}{n}}\  [0,\dots,0,1,\dots,1].
}
In this case, when $\ve(t) \equiv 0$, we have $P_n^\ve(t) \equiv P_n(t)$.
The dynamics
\(
\dd \vec{r}_n(t) / \dd t = - P^\ve_n(t) \vec{r}_n(t) / n
\)
give the same trajectory of $\vec{r}_n(t)$ as \cref{eq: evo res}, since $\vec{r}_n(0)=n^{-1/2}[y_1,\dots,y_1,y_2,\dots,y_2]\in \text{span}(u_1,u_2)$. The corresponding perturbed covariance matrix is $F_n^\ve(t)=I_n-P_n^\ve(t)/n$.

We set $\ve(t)= \bar \ve \rbr{1_{[0,1]}(t)+1_{[1,\infty]}(t)/t^2}$ with $\bar \ve \ll 1$. The propagator $\Omega_n^\ve(t)$ of the evolution $\dd \vec{r}_n(t) / \dd t = -P_n^\ve(t) \vec{r}_n(t) / n$ is
\aeqs{
\Omega_n^\ve(t) = \exp\rbr{-\f{\int_0^tP_n^\ve(s)\dd s}{n}} = U\exp\rbr{-\f{\int_0^t \L^\ve(s) \dd s}{n}}U^\top.
}
Hence, the effective Gram matrix $K_n(0,t)$ can be calculated as
\aeqs{
K_n(0,t) &= \f{\int_0^t U \exp \rbr{- \f{2 \int_0^s \L^\ve(u) \dd u}{n}} \rbr{I-\f{\L^\ve(s)}{n}} U^\top \exp(-(t-s)\bar c) \dd s}{n-1}\\
&= U \L^K(t) U^\top
}
where $\L^K(t) = [\l^K_1(t),\dots,\l^K_n(t)]$, and we have 
\aeqs{
\l^K_1(t) = \l^K_2(t) = \l(t) &\triangleq \f{(1-\bar c)^{-1}/2}{n-1} \rbr{e^{-\bar ct} - e^{-t}},\\
\l^K_3(t) = \dots = \l^K_n(t) = \l'(t) &\triangleq \f{\int_0^t \exp\rbr{-\int_0^s\ve(u)\dd u} \cdot \exp\rbr{-(t-s)\bar c} \cdot \rbr{1-\f{\ve(s)}{2}} \dd s}{n-1}
}
For $\ve(t) = \bar \ve \rbr{1_{[0,1]}(t)+1_{[1,\infty]}(t)/t^2}$, $\exp\rbr{-\int_0^s \ve(u)\dd u} \in [\exp(-2\bar\ve),1]$, $1-\ve(s)/2 \in [1-\bar \ve/2,1]$, hence, for $\bar\ve$ small enough, $\f{1-\exp\rbr{-\bar c t}}{2\bar c (n-1)} \leq \l'(t)\leq \f{1-\exp\rbr{-\bar c t}}{\bar c (n-1)}$. Hence, we have $\l(t) = \Theta(\exp(-\bar ct))$, $\l'(t) = \Theta(1)$.
The generalization gap approaches 0 since $\lambda(t) \to 0$ as $t\to \infty$.

Here we also calculate the solution of $w_n(t)$ and $w_n^{-i}(t)$, although not required in the derivation of the effective gram matrix.
\aeqs{
w_n(t) &= \int_0^t \exp \rbr{-(t-s)\rbr{\f{1}{2}x_1x_1^\top + \f{1}{2}x_2x_2^\top}} \rbr{\f{1}{2}y_1x_1 + \f{1}{2}y_2x_2} ds\\
&= \rbr{1+\exp(-t/2)} \cdot (y_1x_1) + \rbr{1+\exp(-t/2)} \cdot (y_2x_2)
}
Similarly, we have 
\aeqs{
w^{-i}_n(t) &= \rbr{1-\exp(-\f{n-2}{2(n-1)}t)} \cdot (y_kx_k) + \rbr{1-\exp(-\f{n}{2(n-1)}t)} \cdot (y_lx_l)
}
where $k=\lceil 2i/n \rceil$, $l\in \{1,2\}/\{k\}$.
We can see that the solution $w_n(t)$ lies in $\text{span}(x_1,x_2)$. When trained on the dataset $S_n^{-i}$, the progress in the direction $x_{\lceil 2i/n \rceil}$ is slightly less than the other direction, which introduces the non-zero averaged loss difference $\bar \D_n(t)$ during training.
We should also note that the calculation of the contraction and perturbation factors depends heavily on the sample-wise loss gradient $\nabla \ell(w,z_i)$ being supported on $\{x_1, x_2\}$. The clustering of per-sample gradients also happens in the training of neural networks, as shown in \cite{FortEmergent2019}.

\subsection{Proof of \cref{lem:main}}

\begin{proof}[Proof of \cref{lem:main}]
In this proof, we first use class symmetry to
reduce the full-data flow to three scalar variables and show that trainable
attention makes both endpoint kernel modes strictly larger than
$\pi_{\rm sig}^2$, which directly improves the perturbation rate.  We then view
removing one sample as an $O(n^{-1})$ perturbation of the class weights;
uniform endpoint stability transfers the full-data spectrum to the leave-one-out (LOO)
residual dynamics and yields the loss-gap bound.  On an eventually positive
gap, the exact contraction identity can be related to residual quantities, and the improved residual decay rate together with the LOO modal asymptotics yields stronger contraction.



\textbf{Balanced trajectory and endpoint.}
We first reduce the balanced trajectory trained on the full dataset to three scalar coordinates and
analyze its limit
as $t\to\infty$.
We now introduce the transformed residual,
\begin{equation*}
\rho_i =y_if(w,x_i)-1, \qquad
  \ell_i =\frac12\rho_i^2.
\end{equation*}
Permutation invariance, class balance, and the zero initialization imply
$\vartheta_1=\vartheta_2$ and $v\in\Span\{e_1-e_2\}$.  Define the balanced coordinates
\begin{equation*}
 \zeta_{\rm sig} =\kappa\vartheta_1=\kappa\vartheta_2, \quad
  \zeta_{\rm bg} =-\kappa\vartheta_b, \quad
  u =\frac12\langle v,e_1-e_2\rangle.
\end{equation*}
Then $v=u(e_1-e_2)$ and $u(0)=0$.
The total masses for a signal query and a background query are
\begin{equation}
a =\frac{\pi_{\rm sig} e^{\zeta_{\rm sig}}}{\pi_{\rm sig} e^{\zeta_{\rm sig}}+\pi_{\rm bg}},\quad
b =\frac{\pi_{\rm sig} e^{\zeta_{\rm bg}}}{\pi_{\rm sig} e^{\zeta_{\rm bg}}+\pi_{\rm bg}}, \quad
m =\pi_{\rm sig} a+\pi_{\rm bg} b.
\label{eq:attention-masses}
\end{equation}
Hence the pooled class-$j$ feature is
$h_j=me_j+(1-m)e_3$, and both classes have the common transformed residual
\begin{equation*}
  \rho=um-1.
\end{equation*}
With $A=a(1-a)$ and $B=b(1-b)$, direct differentiation gives
\begin{equation}
  \dot u=-\frac12m\rho,
  \qquad
  \dot\zeta_{\rm sig}=-\frac{\kappa^2\pi_{\rm sig}}{2}Au\rho,
  \qquad
  \dot\zeta_{\rm bg}=-\kappa^2\pi_{\rm bg} Bu\rho.
  \label{eq:balanced-ode}
\end{equation}
For $j\in\mathcal C$, let $x_j$ denote any class-$j$ input and let
$k_j=\nabla_w(y_jf(w,x_j))$.  The balanced predictor Gram
matrix has the form
\begin{equation*}
  G(t)=
  \begin{pmatrix}
    k_1^\top k_1&k_1^\top k_2\\
    k_2^\top k_1&k_2^\top k_2
  \end{pmatrix}
  =\begin{pmatrix}d&\chi\\\chi&d\end{pmatrix},
\end{equation*}
where
\begin{align*}
  d&=m^2+(1-m)^2
    +\kappa^2u^2(\pi_{\rm sig}^2A^2+\pi_{\rm bg}^2B^2),
  \\
  \chi&=-(1-m)^2+\kappa^2u^2\pi_{\rm bg}^2B^2.
\end{align*}
Its common and difference eigenvalues are therefore
\begin{align}
  q_{\rm com}
  &=d+\chi
   =m^2+\kappa^2u^2(\pi_{\rm sig}^2A^2+2\pi_{\rm bg}^2B^2),
  \label{eq:qcom}\\
  q_{\rm diff}
  &=d-\chi
   =m^2+2(1-m)^2+\kappa^2u^2\pi_{\rm sig}^2A^2.
  \label{eq:qdiff}
\end{align}
The residual vector $\bm\rho=(\rho_1, \rho_2)$ obeys
$\dot{\bm\rho}=-G\bm\rho/2$.  Since $\rho_1=\rho_2$, the
balanced trajectory lies in the common direction,
\begin{equation}
  \dot\rho=-\frac12q_{\rm com}\rho,
  \qquad
  \rho(0)=-1.
  \label{eq:full-residual-ode}
\end{equation}

Equation~\eqref{eq:full-residual-ode} keeps $\rho<0$.  It follows from
\eqref{eq:balanced-ode} that $u,\zeta_{\rm sig},\zeta_{\rm bg}\ge0$.  Since the functions in
\eqref{eq:attention-masses} are increasing and equal $\pi_{\rm sig}$ at zero,
$a,b\ge\pi_{\rm sig}$ and hence $m\ge\pi_{\rm sig}$.  Consequently
\begin{equation}
  q_{\rm com}\ge m^2\ge\pi_{\rm sig}^2,
  \qquad
  |\rho(t)|\le e^{-\pi_{\rm sig}^2t/2}.
  \label{eq:full-residual-bound}
\end{equation}
Moreover, $\rho=um-1<0$ gives $0\le u<1/\pi_{\rm sig}$.  Since
$A,B\le1/4$, the three equations in~\eqref{eq:balanced-ode} now imply
\begin{equation*}
  |\dot u|+|\dot\zeta_{\rm sig}|+|\dot\zeta_{\rm bg}|\le C_{\pi_{\rm sig},\kappa}|\rho|.
\end{equation*}
Thus the balanced path has finite length and converges to a finite endpoint
$(u_*,\zeta_{{\rm sig},*},\zeta_{{\rm bg},*})$.  Equation~\eqref{eq:full-residual-bound} gives
$\rho_* =0$ and therefore $u_*m_*=1$.  For every $t>0$, one has $u(t)>0$;
because $\kappa>0$, this makes $\zeta_{\rm sig}(t),\zeta_{\rm bg}(t)>0$.  Hence
\begin{equation*}
  a_*>\pi_{\rm sig},
  \qquad
  b_*>\pi_{\rm sig},
  \qquad
  m_*>\pi_{\rm sig}.
\end{equation*}
Write $q_{{\rm com},*}=\lim_{t\to\infty}q_{\rm com}(t)$ and
$q_{{\rm diff},*}=\lim_{t\to\infty}q_{\rm diff}(t)$.
Evaluating~\eqref{eq:qcom}--\eqref{eq:qdiff} at this endpoint yields
\begin{equation*}
  q_{{\rm com},*}\ge m_*^2>\pi_{\rm sig}^2,
  \qquad
  q_{{\rm diff},*}\ge m_*^2>\pi_{\rm sig}^2.
\end{equation*}
Denote the corresponding full parameter endpoint by $w_*$ and set
$G_*=\lim_{t\to\infty}G(t)$.
The endpoint is used only through these inequalities; no explicit root
formula is needed.

The parameter tail is bounded by the tail integral of~\eqref{eq:full-residual-bound},
so the endpoint convergence is exponential.  In particular,
$q_{\rm com}-q_{{\rm com},*}$ is integrable and
\begin{equation}
  \rho(t)
  =-\exp\!\left[-\frac12\int_0^tq_{\rm com}(s)\,\mathrm ds\right]
  =-A_\kappa e^{-\gamma t}(1+o(1)),
  \qquad
  \gamma=\frac{q_{{\rm com},*}}2,
  \quad A_\kappa>0.
  \label{eq:full-residual-asymptotic}
\end{equation}

\paragraph{Perturbation.}
With the balanced endpoint in hand, we compute the perturbation explicitly
and determine its asymptotic decay rate.
Let $w_{n,\kappa}(t)$ denote the zero-initialized full-data trajectory.
On the balanced trajectory, write
$g_j(t)=\nabla_w\ell_j(w_{n,\kappa}(t))=\rho(t)k_j$ for
$j\in\mathcal C$.  Each sample-gradient value occurs $n/2$ times.  Hence
\begin{equation*}
  \bar g(t)=\frac12(g_1(t)+g_2(t))=\frac\rho2(k_1+k_2),
  \qquad
  g_1-\bar g=\frac\rho2(k_1-k_2),
  \qquad
  g_2-\bar g=-\frac\rho2(k_1-k_2).
\end{equation*}
Since $\|k_1-k_2\|^2=2(d-\chi)=2q_{\rm diff}$,
\begin{equation}
  \epsbar_{n,\kappa}(t)
  =\frac{\rho(t)^2}{2(n-1)}q_{\rm diff}(t).
  \label{eq:perturbation-exact}
\end{equation}
Equations~\eqref{eq:full-residual-asymptotic} and
\eqref{eq:perturbation-exact} give
\begin{equation}
  \nu_\epsilon(\kappa)=q_{{\rm com},*}>\pi_{\rm sig}^2.
  \label{eq:trainable-perturbation-rate}
\end{equation}

\paragraph{Large-sample leave-one-out stability.}
We next transfer the balanced endpoint analysis to the LOO flow by proving
stability under the $O(n^{-1})$ class imbalance.
For the two transformed class predictions, define
\begin{equation*}
  \Phi(w)
  =\begin{pmatrix}y_1f(w,x_1)\\y_2f(w,x_2)\end{pmatrix},
  \quad
  \bm\varrho(w)=\Phi(w)-\one_2,
  \quad
  J(w)=D\Phi(w),
  \quad
  \mathcal G(w)=J(w)J(w)^\top.
\end{equation*}
For $|\delta|<1$, set
\begin{equation*}
  \Pi_\delta=\diag\!\left(\frac{1-\delta}{2},
                           \frac{1+\delta}{2}\right).
\end{equation*}
The balanced flow is $\delta=0$, while deletion of one class-$1$ sample is
$\delta=\delta_n=1/(n-1)$.  The following quantities with class imbalance $\delta$ satisfy
\begin{equation}
  \dot w^{(\delta)}
  =-J(w^{(\delta)})^\top \Pi_\delta\bm\varrho_\delta,
  \qquad
  \dot{\bm\varrho}_\delta
  =-\mathcal G(w^{(\delta)})\Pi_\delta\bm\varrho_\delta,
  \qquad
  \bm\varrho_\delta=\bm\varrho(w^{(\delta)}),
  \qquad
  w^{(\delta)}(0)=0.
  \label{eq:weighted-flow}
\end{equation}
By this notation, the balanced trajectory is $w^{(0)}=w_{n,\kappa}$.  For this representative deletion, write
\[
  w_{n,\kappa}^{(-1)}=w^{(\delta_n)}.
\]

Along the balanced path, $\mathcal G=G$ and
\eqref{eq:qcom}--\eqref{eq:qdiff} give
$\mathcal G\succeq\pi_{\rm sig}^2I_2$.  The closure of that path is compact, so
there is a compact tube around it on which
$\mathcal G\succeq(\pi_{\rm sig}^2/2)I_2$ and on which $J$ and the Hessians of
the components of $\bm\varrho$ are bounded.  As long as a perturbed path
remains in the tube, its weighted residual energy
\begin{equation*}
  E_\delta=\frac12\bm\varrho_\delta^\top
  \Pi_\delta\bm\varrho_\delta
\end{equation*}
satisfies, uniformly for sufficiently small $|\delta|$,
\begin{equation}
  \dot E_\delta
  =-(\Pi_\delta\bm\varrho_\delta)^\top
    \mathcal G(w^{(\delta)})(\Pi_\delta\bm\varrho_\delta)
  \le-\lambda_EE_\delta
  \label{eq:weighted-energy}
\end{equation}
for some $\lambda_E>0$.
Here $E_0(t)=\bar\ell(w_{n,\kappa}(t),S_n)$, while
$E_{\delta_n}(t)=\bar\ell(w_{n,\kappa}^{(-1)}(t),S_n^{-1})$.

For sufficiently small $|s|$, let $w^{(s)}$ be the zero-initialized
solution of~\eqref{eq:weighted-flow} with weight $\Pi_s$, and set
$J_s=J(w^{(s)})$ and $\xi_s=\partial_s w^{(s)}$.  On the common time
interval for which these paths remain in the tube,
\eqref{eq:weighted-energy} gives
$\|\bm\varrho_s(t)\|\le Ce^{-\lambda_{\rm tube}t}$ for some $C,\lambda_{\rm tube}>0$, uniformly in $s$.
Differentiation in
$s$ gives
\[
  \dot\xi_s
  =-J_s^\top \Pi_sJ_s\xi_s
   -\sum_{j=1}^2(\Pi_s\bm\varrho_s)_j
      \nabla^2\varrho_j(w^{(s)})\xi_s
   -J_s^\top \Pi_s'\bm\varrho_s,
  \qquad
  \Pi_s'=\tfrac12\diag(-1,1).
\]
Here $\varrho_j$ denotes the $j$-th component of $\bm\varrho$.  The first
term is dissipative and the remaining terms are bounded by
$Ce^{-\lambda_{\rm tube}t}(\|\xi_s\|+1)$; hence
\begin{equation*}
  \frac{\mathrm d}{\mathrm dt}\|\xi_s(t)\|^2
  \le Ce^{-\lambda_{\rm tube}t}(\|\xi_s(t)\|^2+1),
  \qquad \xi_s(0)=0.
\end{equation*}
Gronwall's inequality bounds $\xi_s$ uniformly in $s$ and $t$, so
\begin{equation}
  \sup_{t\ge0}\|w^{(\delta)}(t)-w^{(0)}(t)\|
  \le |\delta|\sup_{|s|\le|\delta|,\,t\ge0}\|\xi_s(t)\|
  \le C|\delta|.
  \label{eq:uniform-stability}
\end{equation}
For sufficiently small $|\delta|$, this closes the tube estimate by a
first-exit argument.  Moreover,
$\|\dot w^{(\delta)}(t)\|\le Ce^{-\lambda_{\rm tube}t}$, so the path converges to a finite
interpolating endpoint and its parameter tail is exponentially small.
For $\delta=\delta_n$, write
\[
  w_{n,*}^{(-1)}
  =\lim_{t\to\infty}w_{n,\kappa}^{(-1)}(t).
\]
Taking $t\to\infty$ in~\eqref{eq:uniform-stability} and using the smoothness
of $\mathcal G$ gives, for every sufficiently large even $n$,
\begin{equation}
  \|w_{n,*}^{(-1)}-w_*\|=O(\delta_n),
  \qquad
  \|\mathcal G_{n,*}^{(-1)}-G_*\|=O(\delta_n),
  \qquad
  \|\mathcal G_n^{(-1)}(t)-\mathcal G_{n,*}^{(-1)}\|=O(e^{-\lambda_{{\rm end},n}t}),
  \label{eq:endpoint-stability}
\end{equation}
where $\mathcal G_n^{(-1)}(t)=\mathcal G(w_{n,\kappa}^{(-1)}(t))$
and $\mathcal G_{n,*}^{(-1)}=\mathcal G(w_{n,*}^{(-1)})$.
Here $\lambda_{{\rm end},n}>0$ may depend on the fixed sample size $n$.

\paragraph{The rate of $\bar \Delta_n$.}
This stability controls the LOO residual decay and yields the asymptotic
lower bound for the loss-gap rate.
For the representative class-$1$ deletion, write
\[
  \Delta_{n,\kappa}^{(-1)}(t)
  =\ell_1(w_{n,\kappa}^{(-1)}(t))
    -\ell_1(w_{n,\kappa}(t)).
\]
By the definition of the averaged loss difference in~\cref{eq: delta},
token permutation invariance and the class-swap isometry make every pointwise
loss difference identical after relabeling, and hence
$\bar\Delta_{n,\kappa}=\Delta_{n,\kappa}^{(-1)}$.
Let $\Pi_n=\Pi_{\delta_n}$.  By~\eqref{eq:endpoint-stability},
\begin{align}
  \lambda_{n,-}
  &=\lambda_{\min}\!\left(
    \Pi_n^{1/2}\mathcal G_{n,*}^{(-1)}\Pi_n^{1/2}\right),
  \notag
  \\
  2\lambda_{n,-}
  &\longrightarrow\lambda_{\min}(G_*)
   =\min\{q_{{\rm com},*},q_{{\rm diff},*}\}.
  \label{eq:spectral-limit}
\end{align}
The weighted residual equation and~\eqref{eq:full-residual-asymptotic}
therefore imply, for every small $\eta>0$ and all large $t$,
\[
  \|\bm\varrho_{\delta_n}(t)\|
  \le C_{n,\eta}e^{-(\lambda_{n,-}-\eta)t},
  \qquad
  |\rho(t)|\le C_\eta e^{-(\gamma-\eta)t}.
\]
Writing $\rho_{1,n}^{(-1)}$ for the first component of
$\bm\varrho_{\delta_n}$ gives
\begin{equation}
  \bar\Delta_{n,\kappa}
  =\Delta_{n,\kappa}^{(-1)}
  =\frac12\left[(\rho_{1,n}^{(-1)})^2-\rho^2\right],
  \qquad
  |\bar\Delta_{n,\kappa}(t)|
  \le C_{n,\eta}e^{-2(\min\{\lambda_{n,-},\gamma\}-\eta)t}.
  \label{eq:gap-residual-form}
\end{equation}
Taking successively the $t$-lower limit, $\eta\downarrow0$, and the even-$n$
lower limit, and using $2\gamma=q_{{\rm com},*}$, yields
\begin{equation}
  \nu_\Delta(\kappa)
  \ge\min\{q_{{\rm com},*},q_{{\rm diff},*}\}
  >\pi_{\rm sig}^2.
  \label{eq:trainable-gap-bound}
\end{equation}

\paragraph{Contraction on a positive $\bar\Delta_n$.}
Assuming the loss gap is eventually positive, we use the LOO residual
asymptotics to lower-bound the original contraction ratio.
Fix a sufficiently large even $n$ and work on the eventual positive tail
assumed in~\cref{lem:main}.  For the representative deletion, set
\[
  \phi_{n,\kappa}^{(-1)}(w)
  =\left\langle
    \nabla\ell_1(w),
    \nabla\bar\ell(w,S_n^{-1})
  \right\rangle,
  \qquad
  N_{n,\kappa}^{(-1)}(t)
  =\phi_{n,\kappa}^{(-1)}(w_{n,\kappa}^{(-1)}(t))
    -\phi_{n,\kappa}^{(-1)}(w_{n,\kappa}(t)),
\]
and
\[
  c_{n,\kappa}^{(-1)}(t)
  =\frac{N_{n,\kappa}^{(-1)}(t)}
          {\Delta_{n,\kappa}^{(-1)}(t)}
  =\bar c_{n,\kappa}(t),
\]
where the last equality follows from~\cref{eq: c}, token permutation
invariance, and the class-swap isometry.  Set
\[
  Q_n=\frac{\rho^2}{(\rho_{1,n}^{(-1)})^2},
  \qquad
  h_n=\frac{q_{\rm com}-\delta_nq_{\rm diff}}2,
\]
where the quantities on the right are evaluated along the full balanced
trajectory.  Then $0\le Q_n<1$.  The LOO and full evaluations of
$\phi_{n,\kappa}^{(-1)}$ are, respectively,
$-\rho_{1,n}^{(-1)}\dot\rho_{1,n}^{(-1)}$ and $h_n\rho^2$; hence
\eqref{eq:gap-residual-form} gives the exact bridge
\begin{equation}
  \frac{\bar c_{n,\kappa}}2
  =\frac{c_{n,\kappa}^{(-1)}}2
  =-\frac{\dot\rho_{1,n}^{(-1)}}{\rho_{1,n}^{(-1)}}
   +\frac{Q_n}{1-Q_n}
    \left(-\frac{\dot\rho_{1,n}^{(-1)}}
    {\rho_{1,n}^{(-1)}}-h_n\right).
  \label{eq:exact-contraction-bridge}
\end{equation}

Endpoint stability gives $\mathcal G_{n,*}^{(-1)}=G_*+O(\delta_n)\succ0$
and turns the weighted residual equation into
\[
  \frac{\mathrm d}{\mathrm dt}
  \bigl(\Pi_n^{1/2}\bm\varrho_{\delta_n}\bigr)
  =-\left[\Pi_n^{1/2}\mathcal G_{n,*}^{(-1)}\Pi_n^{1/2}
    +O(e^{-\lambda_{{\rm end},n}t})\right]
  \Pi_n^{1/2}\bm\varrho_{\delta_n}.
\]
Thus $\lambda_{n,-}$ is the smallest normal rate.  Moreover,
$\mathcal G_{n,*}^{(-1)}\succ0$ makes the endpoint residual Jacobian rank two;
the analytic implicit-function theorem gives a codimension-two interpolation
manifold, and the holomorphic stable-manifold theorem gives a two-dimensional
analytic stable fiber through the LOO endpoint
\cite[Thm.~7.1]{IlyashenkoYakovenko2008}.  Exponential endpoint convergence
places the LOO orbit in this fiber, where $\Pi_n^{1/2}\bm\varrho$ is a local
analytic coordinate.  Writing
$\lambda_{n,+}=\lambda_{\max}(\Pi_n^{1/2}\mathcal G_{n,*}^{(-1)}\Pi_n^{1/2})$,
the convergent Poincar\'e--Dulac normal form on this fiber
\cite[Thm.~5.5]{IlyashenkoYakovenko2008} has only the possible resonance
$\lambda_{n,+}=k\lambda_{n,-}$ with $k\in\mathbb Z_{\ge2}$.  Its solutions are
exponential-polynomials,
so substitution into the residual's convergent Taylor series, followed by
termwise differentiation, gives the simultaneous $C^1$ asymptotics
\begin{equation*}
  \rho_{1,n}^{(-1)}(t)=A_nt^{p_n^{\rm poly}}e^{-\mu_nt}(1+o(1)),
  \qquad
  -\frac{\dot\rho_{1,n}^{(-1)}(t)}{\rho_{1,n}^{(-1)}(t)}
  =\mu_n-\frac{p_n^{\rm poly}}{t}+o(1)\longrightarrow\mu_n\ge\lambda_{n,-},
\end{equation*}
for some $A_n\ne0$ and $p_n^{\rm poly}\in\mathbb Z_{\ge0}$.  The positive gap ensures
that the held-out residual is not identically zero, and $\mu_n$ is a positive
integer combination of the two normal rates.

Together with~\eqref{eq:full-residual-asymptotic}, this gives
\[
  Q_n(t)=\frac{A_\kappa^2}{A_n^2}t^{-2p_n^{\rm poly}}
  e^{-2(\gamma-\mu_n)t}(1+o(1)).
\]
Since $Q_n<1$, necessarily $\mu_n\le\gamma$.  If $\mu_n<\gamma$, then
$Q_n\to0$; if $\mu_n=\gamma$, then
\[
  -\frac{\dot\rho_{1,n}^{(-1)}}{\rho_{1,n}^{(-1)}}-h_n
  \longrightarrow\frac{\delta_nq_{{\rm diff},*}}2>0.
\]
Thus~\eqref{eq:exact-contraction-bridge} yields, in either case,
\[
  \liminf_{t\to\infty}\bar c_{n,\kappa}(t)
  \ge2\mu_n\ge2\lambda_{n,-}.
\]
Taking the outer lower limit and using~\eqref{eq:spectral-limit} gives
\begin{equation}
  c(\kappa)
  \ge\min\{q_{{\rm com},*},q_{{\rm diff},*}\}
  >\pi_{\rm sig}^2.
  \label{eq:trainable-contraction-bound}
\end{equation}

\paragraph{Fixed-attention baseline.}
Finally, we solve the fixed-attention dynamics explicitly to obtain the
baseline rates and complete the comparison.
When $\kappa=0$, attention remains uniform and
\begin{equation*}
  d_0=\pi_{\rm sig}^2+\pi_{\rm bg}^2,
  \qquad
  \chi_0=-\pi_{\rm bg}^2,
  \qquad
  q_{{\rm com},0}=d_0+\chi_0=\pi_{\rm sig}^2.
\end{equation*}
Equation~\eqref{eq:perturbation-exact} immediately gives
$\nu_\epsilon(0)=\pi_{\rm sig}^2$.  For the LOO flow, direct diagonalization of the
constant two-class residual operator gives
\begin{equation*}
  2\lambda_{n,-}^{(0)}
  =d_0-\sqrt{
    \delta_n^2d_0^2+(1-\delta_n^2)\pi_{\rm bg}^4}
  <\pi_{\rm sig}^2,
\end{equation*}
and $2\lambda_{n,-}^{(0)}\to d_0-\pi_{\rm bg}^2=\pi_{\rm sig}^2$.

The initial LOO residual $(-1,-1)^\top$ is not an eigenvector: equality of
the two row sums would require
$(p_--p_+)(d_0-\chi_0)=0$, where
$p_-=(1-\delta_n)/2$ and $p_+=(1+\delta_n)/2$, but both factors are
nonzero.  Since $\chi_0\ne0$, the two eigenvalues are distinct, so the
initial vector has a nonzero slow-mode coefficient.  The same nonzero
off-diagonal entry ensures that the slow eigenvector is visible in the
held-out coordinate.  Because
$2\lambda_{n,-}^{(0)}<\pi_{\rm sig}^2$, this visible LOO mode dominates the
full-data residual.  Thus, for some $B_n\ne0$,
\[
  \rho_{1,n}^{(-1)}(t)
  =B_ne^{-\lambda_{n,-}^{(0)}t}(1+o(1)),
  \qquad
  \frac{\rho(t)^2}{(\rho_{1,n}^{(-1)}(t))^2}\longrightarrow0,
  \qquad
  -\frac{\dot\rho_{1,n}^{(-1)}(t)}{\rho_{1,n}^{(-1)}(t)}
  \longrightarrow\lambda_{n,-}^{(0)}.
\]
Hence the gap is eventually positive with rate $2\lambda_{n,-}^{(0)}$.
The exact bridge~\eqref{eq:exact-contraction-bridge}, which is algebraic and
also applies when $\kappa=0$, gives the same limit for the contraction:
\begin{equation*}
  \liminf_{t\to\infty}
  \left[-\frac1t\log|\bar\Delta_{n,0}(t)|\right]
  =\lim_{t\to\infty}\bar c_{n,0}(t)
  =2\lambda_{n,-}^{(0)}.
\end{equation*}
Letting $n\to\infty$ proves
\begin{equation*}
  \nu_\Delta(0)
  =c(0)
  =\nu_\epsilon(0)
  =\pi_{\rm sig}^2.
\end{equation*}
Combining this equality with
\eqref{eq:trainable-perturbation-rate},
\eqref{eq:trainable-gap-bound}, and
\eqref{eq:trainable-contraction-bound} proves
\eqref{eq:main-unconditional-results} and
\eqref{eq:main-contraction-result}.
\end{proof}

\section{Experimental Details}
\label{app: expt}

\subsection{Setup}
\label{app: expt setup}

\paragraph{Dataset}
We use the MNIST and CIFAR-10 datasets for the experiments in \cref{s: methods,app: other_numerical_experiments}. For each sample size considered in our experiments, we draw a balanced subset from all ten classes and group them into two super-classes to construct a binary classification problem. (MNIST: even / odd; CIFAR10: airplane, bird, deer, frog, ship / automobile, cat, dog, horse, truck).
%


\paragraph{Architectures}
We use WRN-10-2(one residual block per group, channel widths 32–64–128, RMSNorm, global average pooling; 301744 parameters), ViT-small(4×4 patches, 96-dimensional embeddings, four Transformer blocks, four attention heads, MLP ratio 2, and pre-LayerNorm; 310562 parameters), and FC (a two-layer fully connected network, 128 hidden neurons). We use FC for both MNIST and CIFAR experiments, WRN-10-2 and ViT-small for only MNIST experiment.

\paragraph{Training}
We use full-batch gradient descent in all experiments except those in \cref{s: sgd}. Within each comparison, we train all models to the same endpoint residual norm $\norm{\vec r_n}_2$ to ensure a fair comparison. As defined in \cref{eq:normalized_residual}, the residual vector is normalized by $\sqrt{n}$, making its norm comparable across models trained with different numbers of samples.

\paragraph{Synthetic datasets}

Datasets labeled Gaussian-$\alpha$ are created from Gaussian data with different covariance matrices, labeled by a teacher network with random weights.

\begin{itemize}[nosep]
    \item Create a covariance matrix $A$ with $i$-th eigenvalue being $\exp(-\alpha i)$ normalized by the sum of eigenvalues. Construct $A=Q \diag(L) Q^\top$.
    \item Sample the input from the multivariate Gaussian distribution $N(0,A)$.
    \item Project the input onto the top 10 covariance eigendirections $Q_{1:10}$. 
    \item Assign binary labels to the original inputs using a randomly initialized FC-200-30-2 teacher network.
\end{itemize}

The random task in \cref{fig: align} is constructed by assigning random binary labels in $\{0,1\}$ to the original MNIST inputs while maintaining equal class sizes.

\paragraph{Constructing omit-$m$-samples perturbed datasets}
The theory in this paper was developed assuming that the modified dataset $S_n^{\bs i}$ with $(n-1)$ samples is created by omitting the $i$-th sample.
For numerical stability and efficiency of the approximation, in the experiments, we create datasets by omitting a batch of $m$ samples. Let $(m)$ denote a subset of $[n]$ with size $m$. Let $S_{(m)}=\{z_i=(x_i,y_i)_{i\in (m)}\}$. We will conduct experiments using modified datasets $S_n^{\bs (m)}=S_n \setminus S_{(m)}$ obtained by removing $S_{(m)}$ from $S_n$. We therefore consider the weight trajectory $w_n^{\bs (m)}(t)$ of the differential equation $\dot{w}=-\nabla \bar \ell(w,S^{\bs (m)}_n)$. The averaged loss difference is modified in the usual fashion $\bar \D_n(t)=\E_{(m)}\sbr{\D_n^{\bs (m)}}$ with the batch-wise loss difference
\[
\D_n^{\bs (m)}(t)=\ell(w_n^{\bs (m)}(t),S_{(m)})-\ell(w_n(t),S_{(m)}).
\]
Note that $\E_{(m)}$ denotes the expectation taken over the uniform distribution on all possible choices of $(m)$ in $[n]$. The formulae for the averaged contraction and perturbation factors $\bar c_n(t)$ and $\bar \e_n(t)$ in this setting are shown in \cref{app: omit m}.

We should note that for the experiment, we cannot sample all omit-$m$ trajectories. Hence, to reduce approximation variance, the \(m\) omitted samples are drawn evenly from the ten classes, with each trajectory omitting a different subset. To reduce approximation bias, we set \(m\) to less than \(1\%\) of the dataset.
The values of \(m\) and \(n\), together with the number of trajectories, are reported in each figure caption.

\paragraph{Omit-$m$-samples Compact Notation} 
To facilitate the analysis of the omitting-$m$-samples setting, we extend our compact notation from \cref{s: prelim} to denote the relevant batched quantities. We define the batched parameter shift as $\delta w_{(m)} \triangleq w_n^{-(m)} - w_n$. For the first-order derivatives, we denote the average gradient evaluated on the omitted batch $S_{(m)}$ as $g_{(m)} \triangleq \nabla \bar{\ell}(w_n, S_{(m)})$, and the average gradient over the reduced dataset as $g^{-(m)} \triangleq \nabla \bar{\ell}(w_n, S_n^{-(m)})$. Similarly, for the second-order derivatives, we distinguish the single-batch Hessian $H_{(m)} \triangleq \nabla^2 \bar{\ell}(w_n, S_{(m)})$ from the leave-$m$-out dataset Hessian $H^{-(m)} \triangleq \nabla^2 \bar{\ell}(w_n, S_n^{-(m)})$.


\subsection{The approximation of generalization gap}
\label{app: approximation gap}
\cref{tab:result details} records the generalization gaps, averaged loss difference, and its approximations for the experiments done in this paper. In almost all cases, these quantities are close, indicating that the averaged loss difference approximates the generalization gap well. The small training losses show that the models are trained near interpolation. The column $\bar\sigma(K_n)$ reports the mean eigenvalue of the effective Gram matrix.The column $\norm{\vec{r}_n(0)}_2^2$ shows that the squared norm of the normalized initial residual remains nearly constant as the sample size varies (e.g., the MNIST-5 rows). This supports the $1/\sqrt n$ normalization in \cref{eq:normalized_residual} when comparing experiments with different sample sizes.

\begin{table}[!htbp]
\centering
\renewcommand{\arraystretch}{1.5}
\large
\resizebox{1\linewidth}{!}{%
\begin{tabular}{c c r r r r r r r r r r r}
\toprule
\rowcolor[gray]{1}
\textbf{Architecture}
& \textbf{Dataset}
& \textbf{\# samples}
& \textbf{$\bar{\Delta}_n(c,\hat{\epsilon},t)$}
& \textbf{$\bar{\Delta}_n(c,\epsilon,t)$}
& \textbf{$\bar{\Delta}_n(t)$}
& \textbf{$\delta R(S_n,t)$}
& \textbf{$\delta\bar{R}(S_n^{-(m)},t)$}
& \textbf{Generalization error}
& \textbf{$R_{\mathrm{train}}(S_n,t)$}
& \textbf{$\bar{\sigma}(K_n)$}
& \textbf{$\norm{\vec{r}_n(t)}_2$}
& \textbf{$\norm{\vec{r}_n(0)}_2$}
\\
\midrule

FC-784-128-2 & MNIST-2 & 50000
& 0.044 & 0.044 & 0.043 & 0.038 & 0.038
& 0.014 & 0.008 & \text{--} & 0.049 & 0.705
\\

FC-784-128-2 & MNIST-2 & 2000
& 0.130 & 0.132 & 0.130 & 0.117 & 0.117
& 0.054 & 0.030 & 3.285 & 0.095 & 0.702
\\

FC-784-128-2 & MNIST-2 (random label) & 2000
& 0.501 & 1.532 & 1.441 & 1.372 & 1.379
& 0.503 & 0.046 & 5.075 & 0.095 & 0.713
\\

FC-784-128-2 & MNIST-2 & 50
& 0.405 & 0.315 & 0.312 & 0.265 & 0.283
& 0.184 & 0.135 & 1.746 & 0.215 & 0.694
\\

FC-784-128-2 & MNIST-2 & 100
& 0.274 & 0.243 & 0.241 & 0.262 & 0.271
& 0.176 & 0.119 & 1.445 & 0.215 & 0.699
\\

FC-784-128-2 & MNIST-2 & 500
& 0.221 & 0.192 & 0.191 & 0.190 & 0.194
& 0.114 & 0.099 & 1.453 & 0.215 & 0.704
\\

FC-784-128-2 & MNIST-2 & 1000
& 0.153 & 0.136 & 0.135 & 0.147 & 0.148
& 0.096 & 0.097 & 1.131 & 0.215 & 0.704
\\

FC-784-128-2 & MNIST-2 & 2000
& 0.091 & 0.088 & 0.087 & 0.081 & 0.081
& 0.067 & 0.094 & 0.744 & 0.215 & 0.702
\\

FC-784-128-2 & MNIST-2 & 10000
& 0.030 & 0.028 & 0.028 & 0.024 & 0.025
& 0.041 & 0.090 & 0.295 & 0.215 & 0.705
\\

FC-784-128-2 & MNIST-2 & 2000
& 0.126 & 0.126 & 0.125 & 0.112 & 0.112
& 0.054 & 0.036 & 2.612 & 0.110 & 0.702
\\

\midrule

FC-3072-100-2 & CIFAR-2 & 2000
& 0.695 & 0.825 & 0.794 & 0.751 & 0.752
& 0.296 & 0.052 & 6.433 & 0.110 & 0.717
\\

FC-3072-100-2 & CIFAR-2 & 2000
& 0.075 & 0.070 & 0.070 & 0.063 & 0.064
& 0.300 & 0.516 & 0.184 & 0.583 & 0.717
\\

WRN-10-2 & CIFAR-2 & 2000
& 0.070 & 0.040 & 0.040 & 0.033 & 0.033
& 0.279 & 0.514 & 0.114 & 0.583 & 0.702
\\

ViT-small & CIFAR-2 & 2000
& 0.024 & 0.017 & 0.041 & 0.036 & 0.036
& 0.278 & 0.512 & 0.034 & 0.583 & 0.722
\\

\midrule

FC-200-120-2 & Gaussian-1 & 2000
& 0.042 & 0.041 & 0.041 & 0.064 & 0.063
& 0.122 & 0.246 & 0.160 & 0.380 & 0.719
\\

FC-200-120-2 & Gaussian-0.5 & 2000
& 0.023 & 0.022 & 0.022 & 0.045 & 0.045
& 0.111 & 0.245 & 0.099 & 0.380 & 0.720
\\

FC-200-120-2 & Gaussian-0.1 & 2000
& 0.034 & 0.032 & 0.032 & 0.053 & 0.054
& 0.121 & 0.245 & 0.134 & 0.380 & 0.725
\\

FC-200-120-2 & Gaussian-0.05 & 2000
& 0.056 & 0.054 & 0.054 & 0.070 & 0.070
& 0.126 & 0.246 & 0.200 & 0.380 & 0.727
\\

FC-200-120-2 & Gaussian-0.01 & 2000
& 0.106 & 0.101 & 0.101 & 0.116 & 0.117
& 0.162 & 0.255 & 0.297 & 0.380 & 0.726
\\

\bottomrule
\end{tabular}%
}
\caption{
\textbf{Statistics of effective Gram matrix approximation for a variety
of different architectures and datasets}.
See \cref{s: approximation} for the definitions of the generalization
gaps $\delta R(S_n,t)$ and $\delta\bar{R}(S_n^{-(m)},t)$, the averaged
loss difference $\bar{\Delta}_n(t)$, and its approximations
$\bar{\Delta}_n(c,\hat{\epsilon},t)$ and
$\bar{\Delta}_n(c,\epsilon,t)$.
``Generalization error'' refers to the averaged zero-one loss on the
test dataset, reported as a value between zero and one.
$R_{\mathrm{train}}(S_n,t)$ denotes the training loss on $S_n$.
For the last three columns, $\bar{\sigma}(K_n)$ denotes the mean
eigenvalue of the effective Gram matrix,
$\norm{\vec{r}_n(t)}_2$ and $\norm{\vec{r}_n(0)}_2$ denotes the norm of the normalized residual at the training endpoint and initialization respectively.
All quantities except $\norm{\vec{r}_n(0)}_2$ are evaluated at the
specified training endpoint.
}
\label{tab:result details}
\end{table}

\subsection{Other numerical experiments}
\label{app: other_numerical_experiments}

\paragraph{Comparing the statistics for different numbers of samples}
The effective Gram matrix $K_n \in \YY^n\times \YY^n$ lies in a different space when neural networks are trained with different numbers of samples $n$. Therefore, to compare quantities like $\sigma(K_n)$, $M(\vec{r}_n, E(K_n))$ and $M(K_n)$ for different $n$ and the same $\YY$, we use a ``relative index'' as described in \cref{tab: notation}.
We rescale the original index vector to have indices from zero to one.
We should emphasize that by normalizing the residual by $\sqrt{n}$ in \cref{eq:normalized_residual}, the $\ell_2$-norm of the initial residuals $\norm{\vec{r}_n(0)}_2$ is similar when $\YY$ is the  same, even if $n$ is different.
If $K_n=E(K_n)\diag(\sigma(K_n))E(K_n)^\top$, then
\[
\vec{r}_n(0)^\top K_n\vec{r}_n(0)
=\norm{\vec{r}_n(0)}_2^2\sum_i \sigma(K_n)_i P(\vec{r}_n(0),E(K_n))_i^2.
\]
We therefore report both the residual alignment and $\bar\sigma(K_n)=n^{-1}\sum_i\sigma(K_n)_i$, a dimension-normalized measure of the overall spectral scale of $K_n$.
\cref{tab:result details} details the numerical values of these quantities for different datasets and architectures.
We use the quantities defined in \cref{tab: notation} to compare residual–Gram-matrix alignment across tasks.

\begin{table}[!t]
    \centering
    \caption{Quantities that we use to characterize the initial residual and effective Gram matrix.}
    \resizebox{\linewidth}{!}
    {
    \renewcommand{\arraystretch}{1.5}
    \begin{tabular}{l l}
        \toprule
        \textbf{Notation} & \textbf{Definition}\\
        \midrule
        $E(K), \sigma(K)$ & The eigenspace and eigenspectrum of a symmetric matrix $K$ with eigenvalue decomposition\\
        & where $K=E(K) \diag(\sigma(K)) E(K)^\top$, $\sigma(K)$ is the vector of eigenspectrum in ascending order.\\
         $\bar \s (K)$ & The mean of the eigenspectrum of a symmetric matrix $K$, $\bar\s(K)=\sum_i \s(K)_i / n$.\\
        $U_k, U_{1:k}$ & The $k$-th column of $U$, and the first $k$ columns of $U$.\\
        $P(r,U)$ & Normalized projection of a vector $r$ onto the space $U$ (with orthonormal columns).\\
        & the $k$-th element $P(r,U)_k = \abs{r^\top U_k} / \norm{r}_2$.\\
        $M(r,U)$ & ``Explained magnitude'' of a vector $r$ in the space $U_{1:k}$ (with orthonormal columns) \\
        & the $k$-th element $M(r,U)_k = \norm{r^\top U_{1:k}}^2_2 / \norm{r}_2^2$.\\
        $R(\text{Idx})$ & ``Relative index" of the index vector $\text{Idx}=[1,2,\dots,l]$, where $R(\text{Idx})=[1/n, 2/n,\dots, 1]$.\\
        \bottomrule
    \end{tabular}
    }
    \label{tab: notation}
\end{table}

\begin{figure}
    \centering
    \includegraphics[width=0.4\linewidth]{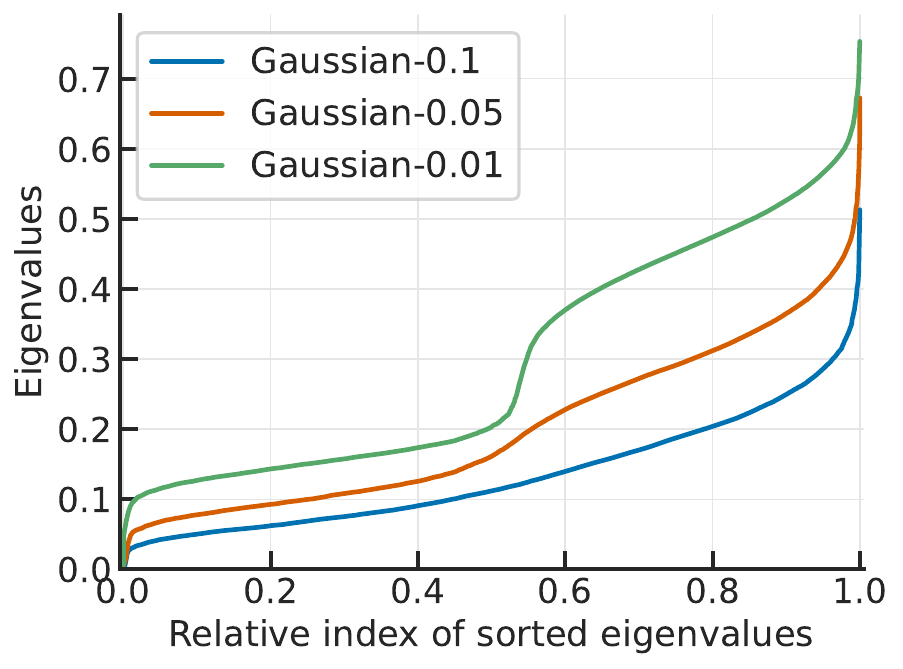}
    \includegraphics[width=0.4\linewidth]{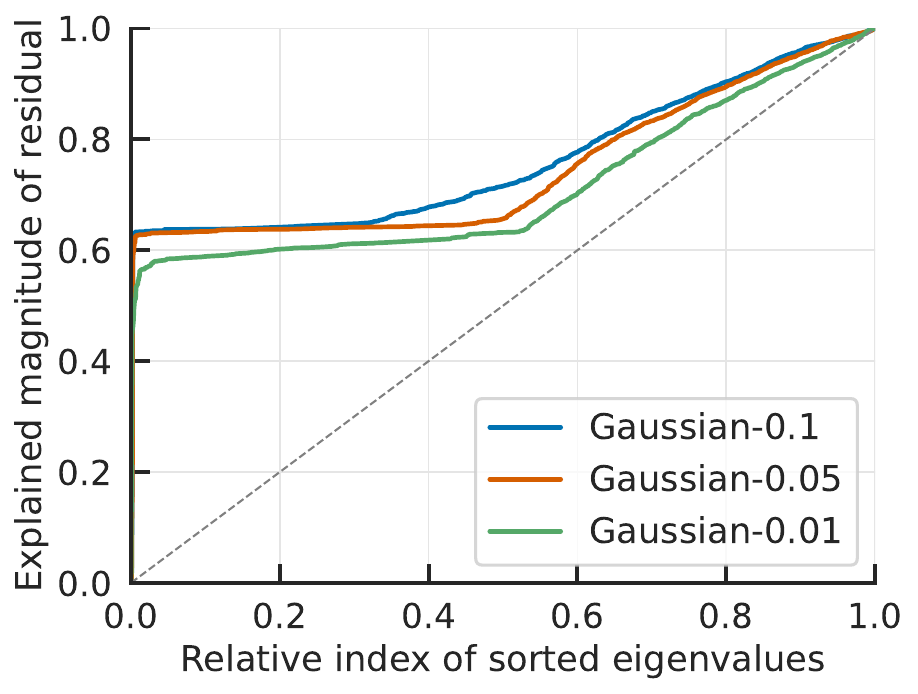}
    \caption{All models are trained to an endpoint with $\norm{\vec{r}_n}_2=0.380$, using $n=2000$, $m=10$, and 100 trajectories.
    The true generalization gaps of Gaussian-$\alpha$ ($\alpha$ from large to small) are 0.053, 0.070, and 0.116, respectively.
    \textbf{Left:} Eigenspectra of the effective Gram matrix $\sigma(K_n)$ for Gaussian-$\alpha$ have larger magnitude when $\alpha$ is smaller ($\bar\s(K_n)$ is 0.134, 0.200, and 0.297, respectively, for $\alpha$ from large to small).
    \textbf{Right:} Explained magnitude of the initial residual trends towards the top-left when we increase $\alpha$ for a larger signal-to-noise ratio.}
    \label{fig: diff tsk}
\end{figure}

\paragraph{Effective Gram matrix for different datasets}
\cref{fig: diff tsk} compares the normalized projection of residual and eigenspectra of the effective Gram matrix on synthetic datasets with different difficulty. 
We construct the following datasets. 
Gaussian-$\alpha$ with smaller $\alpha$ puts less weight on the top eigenvalues as showed in \cite{YangDoes2022}.
From the classical analysis of linear regression, we know that the task becomes more difficult as $\alpha$ decreases, since the labels capture less signal.
We see that for difficult tasks,
the residual projects more onto the subspace corresponding to larger eigenvalues, and the effective Gram matrix $K_n$ has larger magnitude, which jointly lead to a larger predicted generalization gap by our theory. And indeed, the true generalization gap corroborates this trend.

\begin{figure}
\centering
\includegraphics[width=0.4\linewidth]{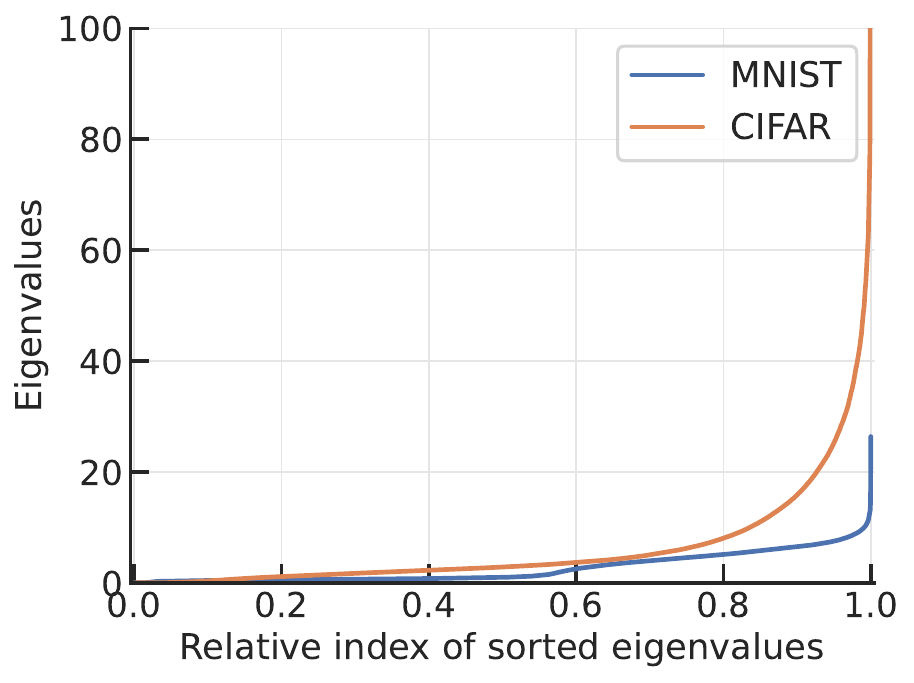}
\includegraphics[width=0.4\linewidth]{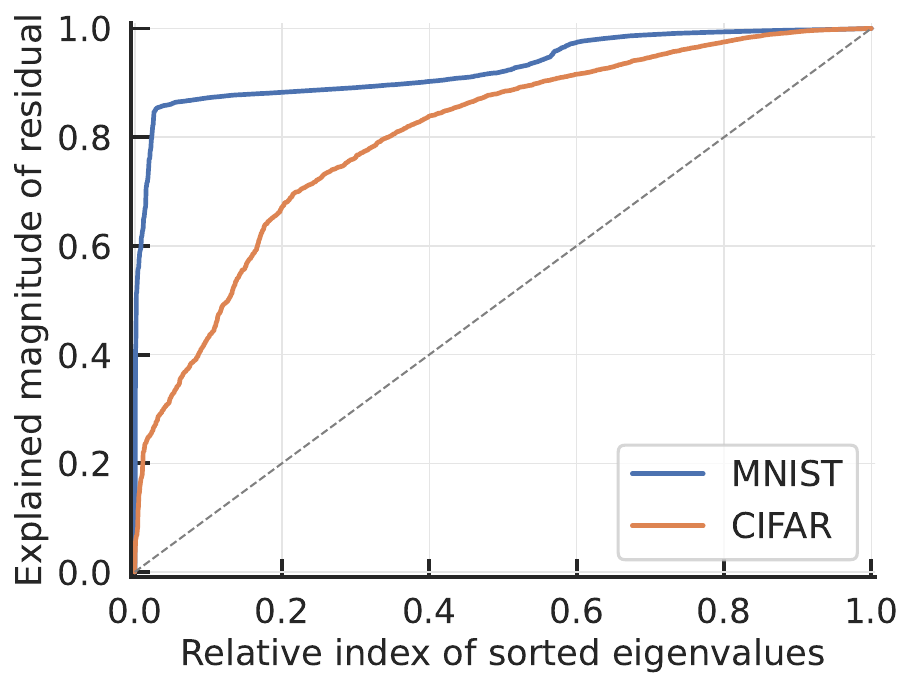}
\caption{
Residuals $\vec{r}_n(0)$ and effective Gram matrix $K_n$ for a fully connected network trained on MNIST-2 and CIFAR-2 with $n=2000$, $m=10$, and 100 trajectories. Both models are trained to the same endpoint with $\norm{\vec{r}_n}_2=0.110$. The generalization gaps for MNIST-2 and CIFAR-2 are 0.112 and 0.751, respectively.
\textbf{Left:} Eigenvalues of the effective Gram matrix $\sigma(K_n)$ have quite different magnitudes ($\bar\sigma(K_n)$ is 2.612 for MNIST-2 and 6.433 for CIFAR-2).
\textbf{Right:} Explained magnitude of the initial residual $M(\vec{r}_n(0),E(K_n))$ for CIFAR-2 has a larger overlap with the principal subspace of the effective Gram matrix than for MNIST-2. This is consistent with the larger generalization gap observed on CIFAR-2.
}
\label{fig: mnist cifar}
\end{figure}

\cref{fig: mnist cifar} compares the two-class tasks derived from MNIST and CIFAR-10. The initial residual for MNIST-2 projects more strongly onto eigenspaces of the effective Gram matrix with small eigenvalues, and the eigenvalues of $K_n$ for CIFAR-2 are larger overall. This shows that both favorable task--Gram-matrix alignment and a small spectral scale of $K_n$ are important for a ``benign training process'' and a small eventual generalization gap.

\begin{figure}
\centering
\includegraphics[width=0.4\linewidth]{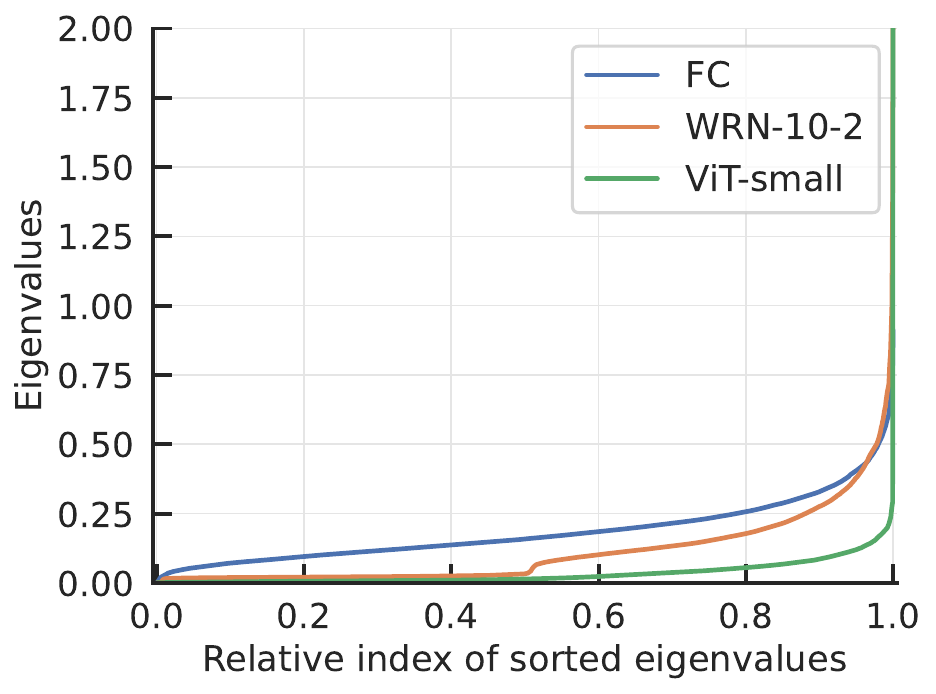}
\includegraphics[width=0.4\linewidth]{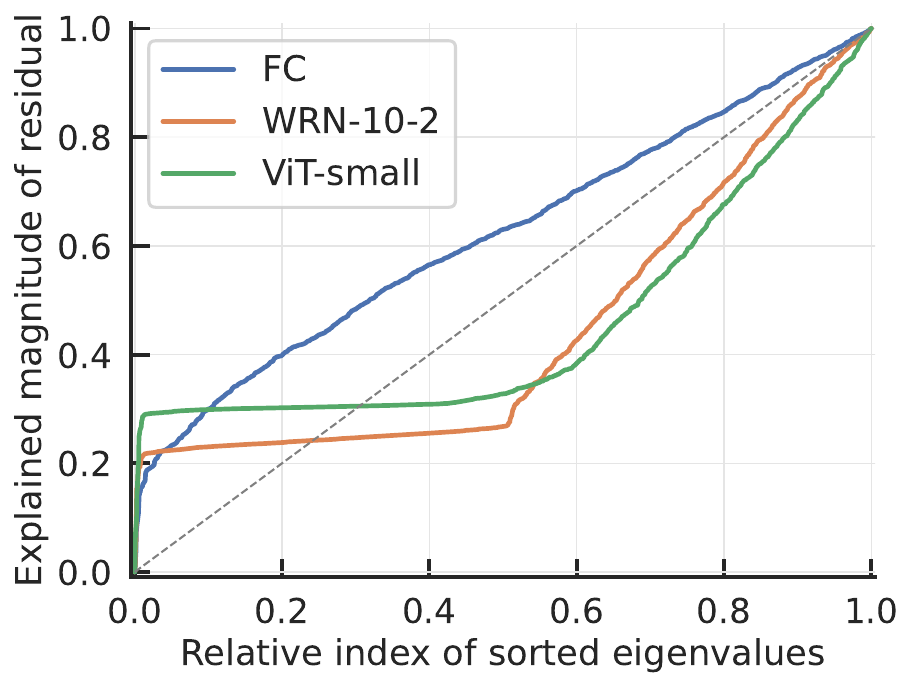}
\caption{Residual $\vec{r}_n(0)$ and effective Gram matrix $K_n$ for CIFAR-2 trained with FC (blue), WRN-10-2 (orange), and ViT-small (green), using $n=2000$, $m=10$, and 100 trajectories. All models are trained to the same endpoint with $\norm{\vec{r}_n}_2=0.583$. The generalization gaps are 0.064, 0.033, and 0.036, respectively.
\textbf{Left:} Eigenvalues $\sigma(K_n)$; the mean eigenvalues $\bar\s(K_n)$ are 0.184, 0.114, and 0.034 for FC, WRN-10-2, and ViT-small, respectively.
\textbf{Right:} Explained magnitude $M(\vec{r}_n(0),E(K_n))$; the initial residuals for WRN-10-2 and ViT-small have larger overlap with the principal subspace of the effective Gram matrix than the residual for FC.}
\label{fig: diff model}
\end{figure}

\paragraph{Effective Gram matrix for different architectures}
\cref{fig: diff model} compares the residual alignment and eigenvalues of the effective Gram matrix for CIFAR-2 trained with FC, WRN-10-2, and ViT-small. FC has the largest mean eigenvalue, whereas WRN-10-2 and ViT-small have more favorable overall spectral scales. The larger spectral scale of the FC effective Gram matrix leads to greater accumulation of the generalization gap, despite its more favorable residual alignment.

\begin{figure}
\centering
\includegraphics[width=0.4\linewidth]{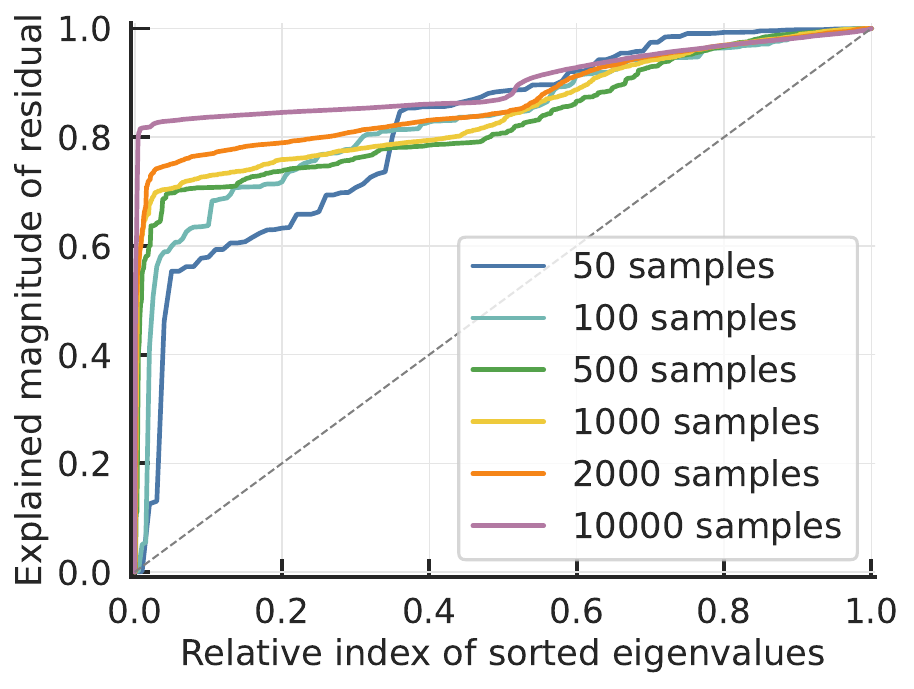}
\includegraphics[width=0.4\linewidth]{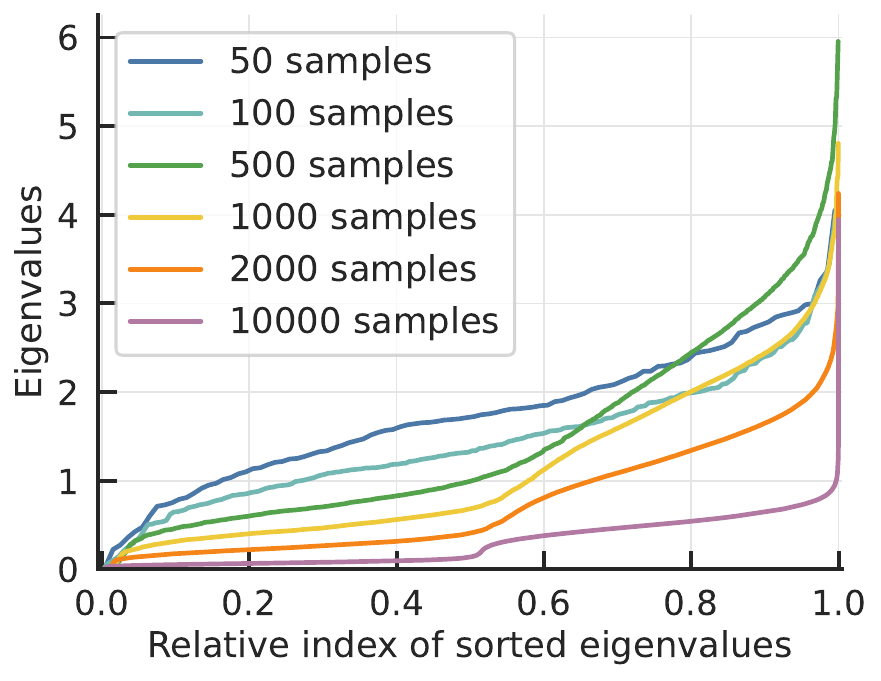}
\caption{
Residuals $\vec{r}_n(0)$ and effective Gram matrix $K_n$ for FC trained on MNIST-2 with $n=50,100,500,1000,2000,$ and $10000$. The models are trained to endpoints with the same residual norm $\norm{\vec{r}_n}_2=0.215$. The generalization gaps are 0.280, 0.270, 0.190, 0.147, 0.083, and 0.023 for $n$ from small to large.
\textbf{Left:} Explained magnitude of the initial residual $M(\vec{r}_n(0),E(K_n))$ has a similar shape for all $n$, but the overlap with the principal subspace of the effective Gram matrix is larger for smaller $n$, which is corroborated by the numerical estimates of the generalization gap in our experiments.
\textbf{Right:} The mean eigenvalue $\bar\sigma(K_n)$ generally decreases as $n$ increases (1.746, 1.445, 1.453, 1.131, 0.743, and 0.295 for $n$ from small to large).
}
\label{fig: diff samples}
\end{figure}

\paragraph{Effective Gram matrix for different number of samples}
\cref{fig: diff samples} compares the training of datasets with different sizes.
As $n$ increases, the initial residual of MNIST-2 projects more strongly onto the tail subspaces of the effective Gram matrix $K_n$, while the overall spectral scale of $K_n$ generally decreases. Both trends are consistent with the smaller generalization gap observed with more training samples.

\end{appendix}